\documentclass[10pt,twocolumn,letterpaper]{article}
\usepackage[pagenumbers]{wacv}
\usepackage{graphicx}
\usepackage{amsmath}
\usepackage{amssymb}
\usepackage{booktabs}
\usepackage{multirow}
\usepackage{enumitem}
\usepackage{microtype}

\usepackage[pagebackref,breaklinks,colorlinks]{hyperref}

\usepackage[capitalize]{cleveref}
\crefname{section}{Sec.}{Secs.}
\Crefname{section}{Section}{Sections}
\Crefname{table}{Table}{Tables}
\crefname{table}{Tab.}{Tabs.}

\usepackage[table]{xcolor}
\definecolor{Gray}{gray}{0.9}
\newcommand{\mysubparagraph}[1]{\vspace{.2cm} \noindent \textbf{#1:}}

\def\wacvPaperID{206} 
\def\confName{WACV}
\def\confYear{2025}

\begin{document}

\title{Dynamic Attention-Guided Diffusion for Image Super-Resolution}

\author{Brian B. Moser$^{1,2,3}$ \hspace{0.6em} Stanislav Frolov$^{1,2,3}$ \hspace{0.6em} Federico Raue$^{1}$ \hspace{0.6em} Sebastian Palacio$^{1}$ \hspace{0.6em} Andreas Dengel$^{1,2}$\\
$^1$German Research Center for Artificial Intelligence, Germany\\
$^2$RPTU Kaiserslautern-Landau, Germany \\
$^3$Equal Contribution\\
{\tt\small first.last@dfki.de}
}
\maketitle

\begin{abstract}
   Diffusion models in image Super-Resolution (SR) treat all image regions uniformly, which risks compromising the overall image quality by potentially introducing artifacts during denoising of less-complex regions.
    To address this, we propose ``You Only Diffuse Areas'' (YODA), a dynamic attention-guided diffusion process for image SR.
    YODA selectively focuses on spatial regions defined by attention maps derived from the low-resolution images and the current denoising time step. 
    This time-dependent targeting enables a more efficient conversion to high-resolution outputs by focusing on areas that benefit the most from the iterative refinement process, i.e., detail-rich objects.
    We empirically validate YODA by extending leading diffusion-based methods SR3, DiffBIR, and SRDiff. 
    Our experiments demonstrate new state-of-the-art performances in face and general SR tasks across PSNR, SSIM, and LPIPS metrics.
    As a side effect, we find that YODA reduces color shift issues and stabilizes training with small batches.
\end{abstract}

\section{Introduction}
\label{sec:intro}
The goal of image Super-Resolution (SR) is to enhance Low-Resolution (LR) into High-Resolution (HR) images \cite{moser2024diffusion}. 
Improvements to this field significantly impact many applications, like medical imaging, remote sensing, and consumer electronics \cite{10.1007/978-3-031-44210-0_19, el2023single, wang2022review}.
Despite its long history, image SR remains a fascinating yet challenging domain due to its inherently ill-posed nature: any LR image can lead to several valid HR images, and vice versa \cite{anwar2020densely, sun2020learned}.
Thanks to deep learning, SR has made significant progress \cite{dong2015image}.
Initial regression-based methods, such as early convolutional neural networks, work great at low magnification ratios \cite{lim2017enhanced,chen2021learning,wang2018esrgan}.
However, they fail to produce high-frequency details at high magnification ratios ($\geq$ 4) and generate over-smoothed results \cite{10041995}.
Such scale ratios require models capable of hallucinating realistic details that fit the overall image.

Recently, generative diffusion models have emerged with better human-rated quality compared to regression-based methods, but they also introduced new challenges \cite{ho2020denoising, saharia2022image, whang2022deblurring, chung2022mr}.
Their indiscriminate processing of image regions leads to computational redundancies and suboptimal enhancements.
Some recent methods address the first issue and reduce computational demands by working in latent space like LDMs \cite{rombach2022high}, by exploiting the relationship between LR and HR latents like PartDiff \cite{zhao2023partdiff}, or by starting with a better-initialized forward diffusion instead of pure noise like in CCDF \cite{chung2022come}.
Yet, strategies to adapt model capacity based on spatial importance remain underexplored.

This paper takes the first step toward addressing the second issue and challenges the common approach of SR diffusion models by asking:
Do we need to update the entire image at every time step?
We hypothesize that not all image regions require the same level of detail enhancement. 
For instance, a face in the foreground may need more refinement than a simple, monochromatic background.
Recognizing this variability in the need for detail enhancement underscores a critical inefficiency in traditional diffusion methods. 
Treating all image regions uniformly risks compromising the overall image quality by introducing artifacts and shifts to low-complex regions. 
Unlike methods that target computational efficiency, we aim to boost image quality by minimizing distortions across different low-complex regions.

In response, we introduce a diffusion mechanism focusing on detail-rich areas using time-dependent and attention-guided masking.
Our method, coined "You Only Diffuse Areas" (YODA), starts by obtaining an attention map that highlights regions that need more refinement.
After identification, YODA systematically replaces highlighted regions with SR predictions during the denoising process.
In particular, regions with high attention values (detail-rich \& salient) are refined more often.
Our approach is analogous to inpainting methods like RePaint \cite{lugmayr2022repaint}, where only a pre-defined region is updated to generate complementing content. 
In YODA's case, however, the selected regions are time-dependent.
To that end, we design a dynamic approach that creates expanding masks, starting from detail-rich regions and converging toward the overall image.

A key advantage of YODA is its compatibility with existing diffusion models, allowing for a plug\&play application.
We integrate YODA with three models: SR3 \cite{saharia2022image} and DiffBIR \cite{lin2024diffbir} for face SR and SRDiff \cite{li2022srdiff} for general SR. 
Interestingly, YODA achieves notable image quality improvements and also improves the training process. 
When training with smaller batch sizes, SR3 suffers from color shifts \cite{wang2023exploiting, choi2022perception} while YODA produces faithful color distributions.
In summary, our work:
\setlist{nolistsep}
\begin{itemize}[noitemsep]
\item introduces YODA, an attention-guided diffusion approach that emphasizes image areas through masked refinement. 
Thus, it refines detail-rich areas more often, which leads to higher image quality.
\item demonstrates that attention-guided diffusion results in better training conditions, accurate color reproduction, and competitive perceptual quality results.
\item empirically shows that YODA outperforms leading diffusion models in face and general SR tasks.
\item reveals that YODA improves the training performance when using smaller batch sizes, which is crucial in limited hardware scenarios.
\end{itemize}

\section{Background}
Our method uses attention maps for attention-guided diffusion.
We leverage the self-supervised DINO framework \cite{caron2021emerging} to extract attention maps.
Thus, this section introduces the main components: DDPMs \cite{ho2020denoising} and the DINO \cite{caron2021emerging}.
We refer to the supplementary materials for a discussion of related methods, such as other diffusion approaches and spatial-selection SR methods.

\subsection{DDPMs}
Denoising Diffusion Probabilistic Models (DDPMs) employ two distinct Markov chains \cite{ho2020denoising}: the first models the forward diffusion process $q$ transitioning from an input $\mathbf{x}$ to a pre-defined prior distribution with intermediate states $\mathbf{z}_t$, $0 < t \leq T$, while the second models the backward diffusion process $p$, reverting from the prior distribution back to the intended target distribution $p \left( \mathbf{z}_{0} \mid \mathbf{z}_T, \mathbf{x} \right)$.
In image SR, we designate $\mathbf{x}$ as the LR image and the target $\mathbf{z}_{0}$ as the desired HR image.
The prior distribution is usually Gaussian noise.

\mysubparagraph{Forward Diffusion}
In forward diffusion, an HR image $\mathbf{z}_{0}$ is incrementally modified by adding Gaussian noise over a series of time steps. 
This process can be mathematically represented as:
\begin{equation}
    q(\mathbf{z}_{t} \mid \mathbf{z}_{t-1}) = \mathcal{N}(\mathbf{z}_{t} \mid \sqrt{1-\alpha_t}\, \mathbf{z}_{t-1}, \alpha_t \mathbf{I} )
\end{equation}
The hyperparameters $0 < \alpha_{1:T} < 1$ represent the noise variance injected at each time step.
It is possible to sample from any point in the noise sequence without needing to generate all previous steps through the following simplification \cite{sohl2015deep}:
\begin{equation}
    q(\mathbf{z}_t \mid \mathbf{z}_0) = \mathcal{N}(\mathbf{z}_t \mid \sqrt{\gamma_t}\, \mathbf{z}_0, (1-\gamma_t) \mathbf{I}),
\end{equation}
where $\gamma_t = \prod_{i=1}^t (1-\alpha_i)$ . 
The intermediate step $\mathbf{z}_t$ is 
\begin{equation}
    \label{eq_sr3yt}
    \mathbf{z}_t = \sqrt{\gamma_t} \cdot \mathbf{z}_0  + \sqrt{1-\gamma_t} \cdot \varepsilon_t, \quad \varepsilon_t \sim \mathcal{N} \left( \mathbf{0}, \mathbf{I} \right)
\end{equation}

\mysubparagraph{Backward Diffusion}
The backward diffusion process is where the model learns to denoise, effectively reversing the forward diffusion to recover the HR image. 
In image SR, the reverse process is conditioned on the LR image to guide the generation of the HR image:
\begin{equation}
    \label{eq:rev_ddpms_cond}
    p_\theta \left( \mathbf{z}_{t-1} \mid \mathbf{z}_t, \mathbf{x} \right) = \mathcal{N} \left( \mathbf{z}_{t-1} \mid \mu_{\theta}(\mathbf{z}_{t}, \mathbf{x}, \gamma_t), \Sigma_\theta(\mathbf{z}_{t}, \mathbf{x}, \gamma_t) \right)
\end{equation}
The mean $\mu_{\theta}$ depends on a parameterized denoising function $f_\theta$, which can either predict the added noise $\varepsilon_t$ or the underlying HR image $\mathbf{z}_0$. Following the standard approach of Ho et al. \cite{ho2020denoising}, we focus on predicting the noise.
Hence, the mean is:
\begin{equation}
    \mu_{\theta}(\mathbf{x}, \mathbf{z}_{t}, \gamma_t) = \frac{1}{\sqrt{\alpha_t}} \left( \mathbf{z}_{t} - \frac{1-\alpha_t}{\sqrt{1 - \gamma_t}} f_\theta \left( \mathbf{x}, \mathbf{z}_{t}, \gamma_t \right)\right)
\end{equation}
Following Saharia et al.\ \cite{saharia2022image}, setting the variance of $p_\theta(\mathbf{z}_{t-1}|\mathbf{z}_t, \mathbf{x})$ to $(1 - \alpha_t)$ yields the subsequent refining step with $\varepsilon_t \sim \mathcal{N}(\textbf{0},\,\textbf{I})$:
\begin{equation}
    \mathbf{z}_{t-1} \leftarrow \mu_{\theta}(\mathbf{x}, \mathbf{z}_{t}, \gamma_t) + \sqrt{1 - \alpha_t} \varepsilon_t
\end{equation}
\noindent

\mysubparagraph{Optimization}
The optimization goal for DDPMs is to train the parameterized model to accurately predict the noise added during the diffusion process. 
The loss function used to measure the accuracy of the noise prediction is:

\begin{equation}
    \mathcal{L} \left( \theta \right) = \mathop{\mathbb{E}}_{(\mathbf{x}, \mathbf{z}_0)} \mathop{\mathbb{E}}_{t} \bigg\lVert \varepsilon_t - f_\theta \left( \mathbf{x}, \mathbf{z}_{t}, \gamma_t \right) \bigg\rVert_1
\label{eq:loss}
\end{equation}

\subsection{DINO}
DINO is a self-supervised learning approach for feature extractors on unlabeled data \cite{caron2021emerging}.
It employs a teacher and a student network, where the student learns to imitate the features learned by the teacher. 
The student gets only local views of the image (i.e., $96 \times 96$), whereas the teacher receives global views (i.e., $224 \times 224$).
This setup encourages the student to learn ``local-to-global'' correspondences.
The features learned through self-supervision are directly accessible in the self-attention modules.
These self-attention maps provide information on the scene layout and object boundaries. 
We leverage the generality, availability, and robustness of these attention maps as a measure of an image's saliency to guide the diffusion process for significantly improved image quality.
In another context, a similar approach has been applied to image compression, demonstrating its ability to capture essential image content in the attention maps \cite{baldassarre2023variable}.

\begin{figure*}[t!]
    \begin{center}
        \includegraphics[width=\textwidth]{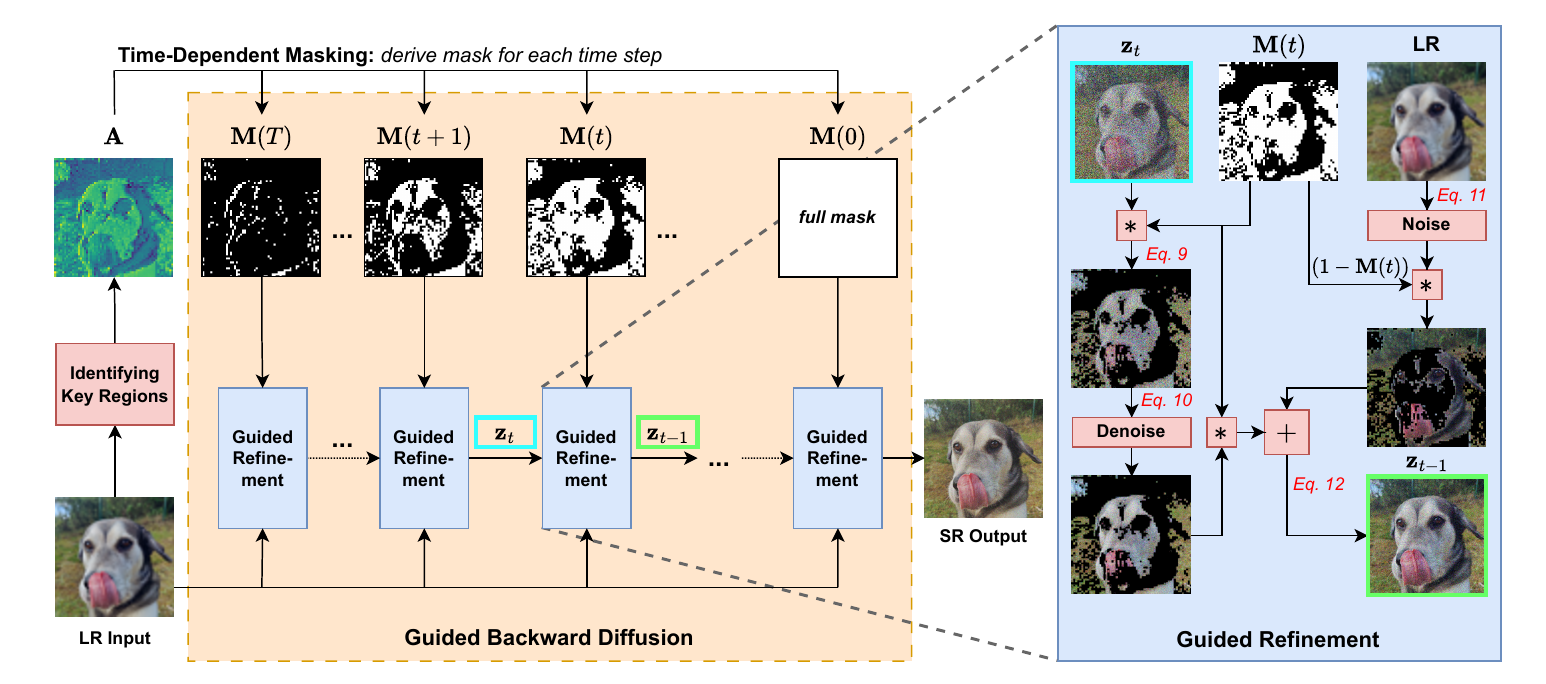}
        \caption{ \label{fig:freya}
        Overview of YODA. 
        First, extract an attention map $\mathbf{A}$ from the LR input.
        Next, use the values of $\mathbf{A}$ to produce a time-dependent masking $\mathbf{M}(t)$.
        For $t:T \to 0$, the area of selected pixels expands from detail-rich regions to the whole image.
        Our diffusion process uses these masks for dynamic and attention-guided refinement, emphasizing regions differently.
        More specifically, it starts with masked areas that need refinement (derived from $\mathbf{z}_{t}$ and $\mathbf{M}(t)$) and LR regions, which retain the noise level needed for the next time step. Finally, the SR and LR areas are combined to form a whole image with no masked-out regions for the next iteration.
        }
    \end{center}
\end{figure*}

\section{Methodology} 
Our proposed method, coined ``You Only Diffuse Areas'' (YODA), has three major phases:
\setlist{nolistsep}
\begin{itemize}[noitemsep]
    \item \textbf{Identifying Key Regions:} 
    Estimate the weighting of pixel positions in a LR image $\mathbf{x} \in \mathbb{R}^{H\times W \times C}$ with an attention map $\mathbf{A} \in \mathbb{R}^{H\times W}$.
    \item \textbf{Time-Dependent Masking:} 
    Use $\mathbf{A}$ to define a time-dependent, binary mask generator function $\mathbf{M}: \mathbb{N}_{0} \to \{0, 1\}^{H\times W}$. The generated masks identify salient areas at time step $t \in \mathbb{N}_{0}$ of the diffusion process.
    \item \textbf{Guided Backward Diffusion:} 
    Concentrate the diffusion process on the regions identified by the time-dependent masking $\mathbf{M}(t)$ and generate a partially enhanced image by combining the prediction with complementing LR areas.
\end{itemize}

\subsection{Identifying Key Regions}
YODA starts by prioritizing areas in the input.
This is achieved by generating an attention map $\mathbf{A}$ with $0 \leq \mathbf{A}_{i,j} \leq 1$ from the LR image $\mathbf{x}$. 
The greater the value of $\mathbf{A}_{i,j}$, the more refinements it receives.
Note that extracting $\mathbf{A}$ is computationally efficient as it has to be generated only once for each image.
For generating $\mathbf{A}$, we evaluated several approaches, including innate methods (i.e., not-learnable) and learnable methods, i.e., ResNet \cite{he2016deep} and Transformer architectures \cite{dosovitskiy2020image}.
For the latter, we leverage the DINO framework for its robustness in self-supervised learning, extracting refined attention maps directly from LR images without necessitating extra annotated data \cite{caron2021emerging}.
This choice is motivated by DINO's demonstrated efficacy in highlighting essential features within images using pre-existing models, e.g., for image compression \cite{baldassarre2023variable}.

Note that it is challenging to define important regions in SR because there is no clear definition.
However, we observe that foreground objects are typically critical to human perception, while background areas seem less significant.
Therefore, we choose to consider self-supervised methods to extract attention maps, which serve as an unbiased proxy for identifying objects and weightings in an image.
This choice is based on the observation that self-supervised methods like DINO naturally focus on regions that generally capture human attention \cite{caron2021emerging}.
Our experiments with YODA confirm that emphasizing these areas improves performance, as later results will demonstrate.
An additional overview of how DINO and the attention maps are used is shown in the supplementary materials.
Next, we describe the process of creating time-dependently masks for the backward diffusion by utilizing the attention map $\mathbf{A}$.

\subsection{Time-Dependent Masking}
\label{sec:time_dependent_masking}
Given the LR input image $\mathbf{x}$ and the attention map $\mathbf{A}$, we introduce a novel strategy to dynamically focus the diffusion process on salient areas. 
Even though $\mathbf{A}$ is fixed, we will use it to refocus the diffusion model during the backward diffusion process dynamically.
Thus, we can leverage it to influence the number of refinement steps for each position. 
Therefore, for two positions $(i,j)$ and $(i', j')$ with $\mathbf{A}_{i,j} > \mathbf{A}_{i',j'}$, YODA applies more refinement steps to the location $(i,j)$ than to $(i', j')$.
Since $0 \leq \mathbf{A}_{i,j} \leq 1$, the number of diffusion steps employed to a specific position $(i,j)$ is determined as a proportion of the maximum time steps, $T$.
For instance, $\mathbf{A}_{i,j} = 0.7$ means $(i,j)$ is refined during 70\% of all diffusion steps.
In addition, we introduce a lower bound hyperparameter $0 < l < 1$, ensuring that every region undergoes a minimum amount of refinements. 
In other words, the hyperparameter $l$ reliably guarantees that every spatial position is refined at least $l \cdot T$ times.
As the backward diffusion process progresses from time step \(T\) to 0, we can define the time-dependent masking process for any time step \(T \geq t \geq 0\) approaching \(t = 0\) as:
\begin{equation}
    \label{eq:td_mask}
    \mathbf{M}(t)_{i,j} = \begin{cases}
                                1\text{, if } T \cdot (\mathbf{A}_{i,j} + l) \geq t \\
                                0\text{, otherwise}
                            \end{cases}
\end{equation}
\autoref{eq:td_mask} ensures that the diffusion process gets applied a variable number of times for different regions, allowing the salient areas to diffuse over a longer time span.
It is important to highlight that once a spatial position is marked for refinement, it continues to undergo refinement across all subsequent steps: $\mathbf{M}(t)_{i,j} \geq \mathbf{M}(t-k)_{i,j} \, \forall k > 0$.
\autoref{fig:freya} shows an example of our time-dependent masking.
For each time step $t$, we can determine with $\mathbf{M}(t)_{i,j} = 1$ whether a given spatial position $(i,j)$ should be refined or not.

\subsection{Guided Backward Diffusion}
YODA's guided diffusion process iteratively refines the image from a noisy state $\mathbf{z}_{T}$ to a HR state $\mathbf{z}_{0}$. 
This phase involves selectively refining areas based on the current time step's mask, $\mathbf{M}(t)$, and blending these refined areas with the unrefined, remaining LR regions, $(1-\mathbf{M}(t))$. 
YODA ensures a seamless transition between refined and unrefined areas, improving image quality with a focus on key regions.

More specifically, at each time step $t$, the areas that will be refined when transitioning from $t$ to $(t-1)$are determined based on the current iteration $\mathbf{z}_{t}$ and the current mask $\mathbf{M}(t)$:
\begin{equation}
    \label{eq:mask1}
    \mathbf{\widetilde{z}}_{t} \leftarrow \mathbf{M}(t) \odot \mathbf{z}_{t}
\end{equation}
Next, we divide the current image$\mathbf{z}_{t}$ into two components that will later be combined as $\mathbf{z}_{t-1}$ for the next time step: $\mathbf{z}^{SR}_{t-1}$, which is the refined image prediction, and $\mathbf{z}^{LR}_{t-1}$, the complementary LR image. 
The state $\mathbf{z}^{LR}_{t-1}$ represents unchanged LR areas by using $\mathbf{x}$ as the mean. 
Both components acquire the same noise level $\Sigma_\theta(\mathbf{\widetilde{z}}_{t}, \mathbf{x}, \gamma_t)$, and can be described by:
\begin{align}   
    \label{eq:10}
    \mathbf{z}^{SR}_{t-1} &\sim \mathcal{N} \left( \mu_{\theta}(\mathbf{\widetilde{z}}_{t}, \mathbf{x}, \gamma_t), \Sigma_\theta(\mathbf{\widetilde{z}}_{t}, \mathbf{x}, \gamma_t) \right) \\
     \label{eq:11}
    \mathbf{z}^{LR}_{t-1} &\sim \mathcal{N} \left( \mathbf{x}, \Sigma_\theta(\mathbf{\widetilde{z}}_{t}, \mathbf{x}, \gamma_t) \right)
\end{align}
Finally, YODA combines the complementing and non-overlapping image regions into a full image\footnote{\autoref{eq:lr_sampling} is similar to RePaint \cite{lugmayr2022repaint}, a diffusion-based inpainting method
While RePaint uses a constant mask for all time steps, YODA has time-dependent masks to dynamically control the updated image regions.}:
\begin{equation}
    \label{eq:lr_sampling}
    \mathbf{z}_{t-1} \leftarrow  \mathbf{M}(t) \odot \mathbf{z}^{SR}_{t-1} + (1-\mathbf{M}(t)) \odot \mathbf{z}^{LR}_{t-1}
\end{equation}
Consequently, the areas refined by $\mathbf{z}^{SR}_t$ expand as $t \to 0$, whereas the areas described by $\mathbf{z}^{LR}_t$ shrink in size.
The new state, $\mathbf{z}_{t-1}$, now contains both SR and LR areas and, importantly, does not have any masked-out regions.
As a result, $\mathbf{z}_{t-1}$ can be used in the next iteration step. 
This guided refinement is depicted on the right part of \autoref{fig:freya}.

Note that we use $\mathbf{\widetilde{z}}_{t}$ instead of $\mathbf{z}_t$ in \autoref{eq:10} and \autoref{eq:11}. 
In our initial experiments, masking before the noise prediction (i.e., $\mathbf{\widetilde{z}}_{t}$) produced a marginal improvement in comparison to the full intermediate state $\mathbf{z}_t$ (around 0.1-0.2 dB in PSNR).
We theorize that it is connected to the optimization target explained next.
With $\mathbf{\widetilde{z}}_{t}$, we force the model inherently to focus locally, which, due to our selective loss function, would otherwise have to be learned.

\subsection{Optimization}
To confine the backward diffusion process to specific image regions as determined by the current time step $0 \leq t \leq T$ and the corresponding mask $\mathbf{M}(t)$, we adapt the training objective from \autoref{eq:loss} as follows to focus on regions within the mask $\mathbf{M}(t)$.
Thus, YODA optimizes only areas described by $\mathbf{M}(t)$:
\begin{equation}
    \mathcal{L} \left( \theta \right) = \mathop{\mathbb{E}}_{(\mathbf{x}, \mathbf{y})} \mathop{\mathbb{E}}_{t} \bigg\lVert \mathbf{M}(t) \odot \left[ \varepsilon_t - f_\theta \left( \mathbf{x}, \mathbf{z}_{t}, \gamma_t \right) \right] \bigg\rVert_1
\label{eq:meth_loss}
\end{equation}

\section{Experiments}
We start by analyzing different methods for obtaining attention maps for YODA.
Then, we evaluate YODA's performance in tandem with SR3 \cite{saharia2022image} and DiffBIR \cite{lin2024diffbir} for face, as well as SRDiff \cite{li2022srdiff} for general SR.
We chose SR3, DiffBIR, and SRDiff because they are the most prominent representative diffusion models for image SR in the respective tasks, where YODA can be integrated straightforwardly.
However, YODA can be theoretically applied to any existing method.
We present quantitative and qualitative results for both tasks, demonstrating YODA's high-quality results compared to the baselines using standard metrics such as PSNR, SSIM, and LPIPS \cite{10041995}.
All experiments were run on a single NVIDIA A100-80GB GPU.
In the supplementary materials, we discuss the complexity of YODA and explore its potential synergies with other diffusion models.
Also, we used a lower bound hyperparameter (see \autoref{sec:time_dependent_masking}) of $l=0.2$ in all experiments and were inspired by the rate-distortion trade-off presented by Ho et al. \cite{ho2020denoising} that reaches the semantic compression stage at roughly $t=T-0.2 \cdot T$.

\begin{figure*}[t]
    \centering
    \begin{minipage}{0.65\textwidth}
        \centering
        \includegraphics[width=\textwidth]{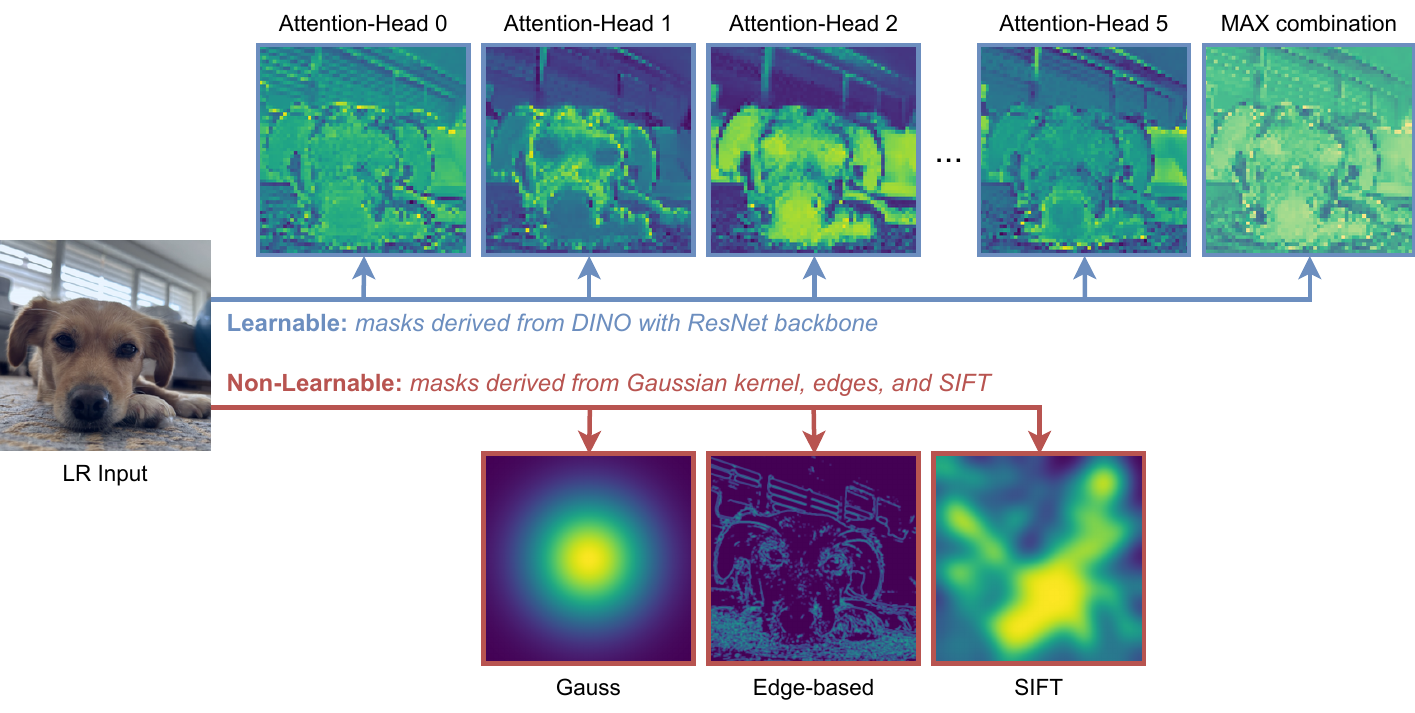} 
    \end{minipage}\hfill
    \begin{minipage}{0.35\textwidth}
        \centering
        \resizebox{.85\columnwidth}{!}{%
        \begin{tabular}{l l c c c}
            \toprule
             &\textbf{Attention Maps} & \textbf{PSNR} $\uparrow$ & \textbf{SSIM} $\uparrow$ & \textbf{LPIPS} $\downarrow$ \\
            \midrule
               \multirow{4}{*}{\rotatebox[origin=c]{90}{no masks}} &PULSE & 16.88 & 0.440 & n.a. \\
               &FSRGAN & 23.01 & 0.620 & n.a.  \\
               &SR3 Reported & 23.04 & 0.650 & n.a.  \\ 
               &SR3 Reproduced &22.35 & 0.646 & 0.082 \\
            \midrule
               \multirow{3}{*}{\rotatebox[origin=c]{90}{innate}} & Gaussian &   22.13 & 0.602 & 0.260  \\
               & Edge-based & 22.93 & 0.648 & 0.151  \\
               & SIFT & 22.84 & 0.678 & 0.095 \\
            \midrule
               \multirow{8}{*}{\rotatebox[origin=c]{90}{DINO with ViT-S/8}} & Attention-Head 0 & 22.91 & 0.650 & 0.105 \\
                &  Attention-Head 1 & 22.43 & 0.616 & 0.130 \\
                &  Attention-Head 2 & 22.55 & 0.633 & 0.111 \\
                &  Attention-Head 3 & 22.73 & 0.641 & 0.110 \\
                &  Attention-Head 4 & 22.85 & 0.645 & 0.097\\
                &  Attention-Head 5 & 22.86 & 0.648 & 0.101 \\
                &  Attention-AVG   & 23.25 & 0.663 & 0.122 \\
                &  Attention-MAX   & 23.46 & \underline{0.683} & 0.103 \\
            \midrule
                \multirow{8}{*}{\rotatebox[origin=c]{90}{DINO with ResNet-50}} &  Attention-Head 0  & 22.82 & 0.649 & 0.115 \\
                &  Attention-Head 1  & 22.54 & 0.627 & 0.117 \\
                &  Attention-Head 2  & 22.84 & 0.650 & 0.107 \\
                &  Attention-Head 3  & 22.78 & 0.645 & 0.105\\
                &  Attention-Head 4  & 22.38 & 0.620 & 0.127 \\
                &  Attention-Head 5  & 22.50 & 0.630& 0.119\\
               &   Attention-AVG    & \underline{23.55} & 0.682 & \underline{0.093}\\
                 & \textbf{Attention-MAX} & \textbf{23.84} & \textbf{0.695} & \textbf{0.072} \\
               
            \bottomrule
    \end{tabular}
    }
    \end{minipage}
    \caption{
        \textbf{(Left)} 
        Comparison of various methods to extract attention maps used for our method (blue = low attention; yellow = high attention).
        Top row denotes maps derived from ResNet-50 using DINO.
        It shows various attention head outputs and the max aggregation of all attention maps (MAX). 
        Bottom row denotes non-learnable methods, namely Gaussian, Edge-based, and using SIFT's points of interest.
        \textbf{(Right)}
         Comparison of different attention maps with SR3+YODA for $16 \rightarrow 128$ on CelebA-HQ. 
    Aggregating the attention maps extracted with DINO and ResNet-50 backbone under the MAX strategy performs best.
    The attention maps are then used for dynamic binary masking.
    }
    \label{fig:ablAttHeads}
\end{figure*}

\subsection{Choosing Good Attention Maps}
YODA relies on attention maps. 
Thus, we thoroughly evaluated different choices.
We considered the pre-trained attention heads from the last layer of DINO with the respective backbone model, i.e., ResNet \cite{he2016deep} and ViT \cite{dosovitskiy2020image}.
For ResNet-50, we used a dedicated method to extract the attention maps from its weights \cite{gur2021visualization}.
A qualitative comparison of attention maps generated with DINO and ResNet-50 is shown in \autoref{fig:ablAttHeads} (left), demonstrating that YODA highlights perceptually essential areas (more visual results are in the appendix). 
Additionally, we test non-learnable methods to extract attention maps, also shown in \autoref{fig:ablAttHeads} (left):
\setlist{nolistsep}
\begin{itemize}[noitemsep]
\item \textbf{Gaussian:} Placing a simple 2D Gaussian pattern at the center of the image provides a straightforward approach, which relies on the assumption that the essential parts of an image are centered.
\item \textbf{Edge-based:} Using the Canny edge detector, the attention maps are defined by the edges of the image, where close edges are connected and blurred.
\item \textbf{Scale-Invariant Feature Transform (SIFT):} Through Gaussian differences, SIFT \cite{lowe2004distinctive} provides an attention map characterized by scale invariance. 
It produces an attention map by applying 2D Gaussian patterns around the points of interest.
\end{itemize}

\mysubparagraph{DINO masks perform best}
\autoref{fig:ablAttHeads} (right) presents our study on several baselines and masking variants.
The straightforward Gaussian approach performs worst as it does not adapt to image features.
The edge-based segmentation and SIFT methods improve the performance over the reproduced baseline using a small batch size.
However, they underperform relative to the reported SR3 results, which used a larger batch size.
In comparison, using DINO's attention maps for YODA shows significant improvements. 
We tested individual attention heads (0 to 5) independently, along with combination strategies that include averaging (AVG) and selecting the maximum value (MAX).
The MAX combination achieved the best results compared to individual heads or the AVG combination.

\begin{figure*}[!t] 
\captionsetup[subfigure]{position=b}
\centering
   \begin{subfigure}[c]{0.49\linewidth}
       \centering
       \includegraphics[width=.83\linewidth]{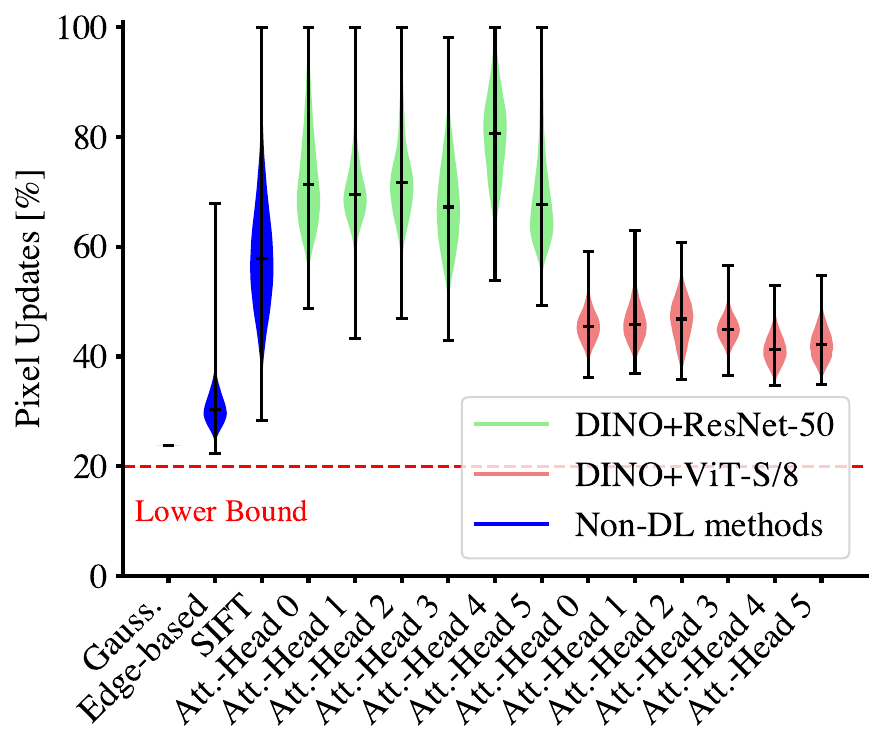}
   \end{subfigure}
   \begin{subfigure}[c]{0.49\linewidth}
       \centering
       \includegraphics[width=.71\linewidth]{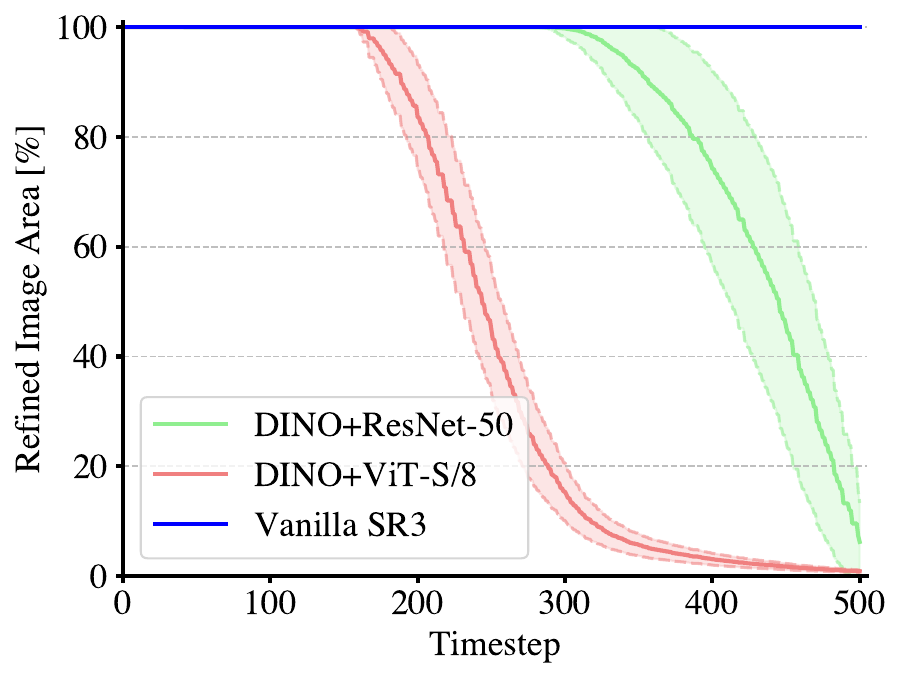}
   \end{subfigure}
   
   \caption{ \textbf{(Left)} Ratio comparison between diffused pixels using our time-dependent masking approach and the total number of pixel updates in standard diffusion. On average, DINO with a ResNet-50 backbone leads to more pixel updates than the VIT-S/8 backbone. The lower bound, defined by $l$, is a threshold to eliminate areas that would never undergo diffusion. \textbf{(Right)} Refined image area in percentage across time steps for the MAX combination.
  Note that the sampling process goes from $T=500$ to $T=0$.
  ResNet-50 initiates the refinement process much earlier, advances more rapidly toward refining the entire image, and has a higher standard deviation.
   }
   \label{fig:iterations}
\end{figure*}

\mysubparagraph{ResNet produces more sensitive masks with more pixel updates}
\autoref{fig:iterations} (left) investigates the ratio of diffused pixels using our time-dependent masking. 
The upper bound is 100\%, where diffusion is applied across all locations in every time step (standard diffusion). 
Any result under 100\% shows that not all pixels are diffused during all time steps.
As can be seen, DINO with ResNet-50 produces higher attention values, resulting in more total pixel updates.
Also, the high variance indicates a high adaptability. 
It can employ 100\% for some samples, a characteristic not observed with ViT-S/8.
Nevertheless, the ViT-S/8's improved performance compared to non-learnable methods and its low ratio make it attractive for future work on optimized inference speed based on sparser diffusion, e.g., LazyDiffusion \cite{nitzan2024lazy}.

\mysubparagraph{ResNet progresses faster towards the whole image}
\autoref{fig:iterations} (right) shows the ratio of diffused pixels depending on time steps with the MAX aggregation for ResNet-50 and ViT-S/8. 
As the backward diffusion progresses from $T$ to $0$, ResNet-50 initiates and incorporates the refinement of the whole image areas more quickly than ViT-S/8. 
For the first 200 time steps, the attention map derived by ViT-S/8 addresses less than 20\% of the image area, whereas ResNet-50 has already developed to 100\%.
Therefore, we assume that the ResNet-50 backbone's superior performance is attributed to its faster progression toward refining the whole image.
Intermediate diffusion results and error maps can be found in the supplementary materials.

\mysubparagraph{Summary} As DINO with ResNet-50 and MAX aggregation performs best, we used it for all remaining main experiments.

\subsection{Face Super-Resolution}
We use FFHQ \cite{karras2019style} for training and CelebA-HQ for testing \cite{karras2017progressive}.
All SR3 models were trained for 1M iterations as in Saharia et al. \cite{saharia2022image}.
We evaluated three scenarios with bicubic degradation: $16 \rightarrow 128$, $64 \rightarrow 256$, and $64 \rightarrow 512$.
Due to hardware limitations and missing quantitative results in the original publication of SR3, our experiments required a reduction from the originally used batch size of 256: we used a batch size of 4 for the $64 \rightarrow 512$, and 8 for the $64\rightarrow 256$ scenario to fit on a single A100-80GB GPU.
For blind face SR (unknown degradation between LR and HR, $64\rightarrow 256$), we follow Lin et al. and test YODA with DiffBIR \cite{lin2024diffbir}, which uses more complex degradation models (e.g., blurring), like introduced by Real-ESRGAN and others \cite{wang2018esrgan, xiao2019deep}.

\begin{table}[t]
    \centering
    \resizebox{\columnwidth}{!}{%
    \begin{tabular}{c l l c c c c  }
    \toprule
    \textbf{Scaling} & \textbf{Type} & \textbf{Model} & \textbf{PSNR} $\uparrow$ & \textbf{SSIM} $\uparrow$ & \textbf{LPIPS} $\downarrow$ &  \textbf{FID} $\downarrow$ \\
    \midrule
       4× & Regression & RRDB \cite{wang2018esrgan} & 27.77 & 0.870 & 0.151 & 67.46\\
      & Diffusion & SR3 \cite{saharia2022image}         & 17.98 & 0.607 & 0.138 & 80.72 \\
       & Diffusion & \textbf{SR3 + YODA}  & \textbf{26.33} & \textbf{0.838} & \textbf{0.090} & \textbf{59.99}
       \\
    \midrule
    8× & Regression & RRDB \cite{wang2018esrgan} & 26.91 & 0.780 & 0.220 & 62.85 \\
    & Diffusion & PartDiff ($K$=25) \cite{zhao2023partdiff}& n.a. & n.a. & 0.222 & n.a.\\
       & Diffusion & PartDiff ($K$=50) \cite{zhao2023partdiff}& n.a. & n.a. & 0.217 & n.a.\\
       & Diffusion & SR3 \cite{saharia2022image}         & 17.44 & 0.631 & 0.147 &  66.20\\
       & Diffusion & \textbf{SR3 + YODA}  & \textbf{25.04} & \textbf{0.800} & \textbf{0.126}& \textbf{53.95}
       \\
    \midrule
    8× & Diffusion & DiffBIR \cite{lin2024diffbir} & 24.49 & 0.717 & 0.247 & 115.22 \\
    (blind) &  Diffusion & \textbf{DiffBIR + YODA} & \textbf{24.56} & \textbf{0.718} & \textbf{0.245} & \textbf{111.93}\\
    \bottomrule
    \end{tabular}
    }
    \caption{\label{tab:faceSR} Face SR results with 4× scaling ($64 \rightarrow 256$) and 8× scaling ($64 \rightarrow 512$) on CelebA-HQ with SR3 (non-blind) and DiffBIR (blind means unknown degradation type) standalone and combined with YODA. 
    Note that RRDB \cite{wang2018esrgan} is also reported and that regression-based methods typically yield higher pixel-based scores (PSNR/SSIM) than generative approaches \cite{saharia2022image}.
    SR3 was trained for 1M steps and a reduced batch size of 4 and 8 instead of 256 to fit on a single A100-80GB GPU.
    Note that DiffBIR uses diffusion after an initial, regression-based predictor with only 50 sampling steps.
    Thus, smaller relative improvements are expected.
    }
\end{table}

\mysubparagraph{Results}
\autoref{tab:faceSR} shows significant improvements when SR3 (face SR) or DiffBIR (blind face SR) is coupled with YODA across all metrics.
DiffBIR applies a diffusion process following an initial regression-based predictor and uses only 50 sampling steps. 
Consequently, smaller relative improvements are expected, but it shows that YODA is also efficient for more complex degradation models.

We explain the poor performance of SR3 with a phenomenon also observed by other works \cite{wang2023exploiting, choi2022perception}: color shifting, which we attribute to the reduced batch size necessitated by limited hardware access.
Fitting SR3 on a single A100-80GB GPU required reducing the batch size from 256 to 8.
Examples are shown in \autoref{fig:image64by64} and in the appendix.
Color shifting manifests in pronounced deviation in pixel-based metrics (PSNR/SSIM) but only slightly decreased perceptual quality (LPIPS).
Our results suggest that YODA's role extends beyond mere performance enhancement. 
It actively mitigates color shifting and stabilizes training, especially when faced with hardware constraints. 
Due to selective refinement, YODA maintains more of the LR image in less complex areas, reducing the risk of introducing artifacts/shifts during denoising while enhancing details in complex regions.
With YODA, SR3 can be trained with a much smaller batch size.

\mysubparagraph{Results without color-shifting}
\autoref{tab:faceSR_color} shows the same experiments but with the channel-wise mean normalized by the ground-truth data, thus disentangling image quality improvements from color bias. 
While this procedure is not feasible in practice, it highlights YODA's significantly improved performance beyond reducing color shifting.

\begin{figure}[t!]
    \begin{center}
        \includegraphics[width=.88\linewidth]{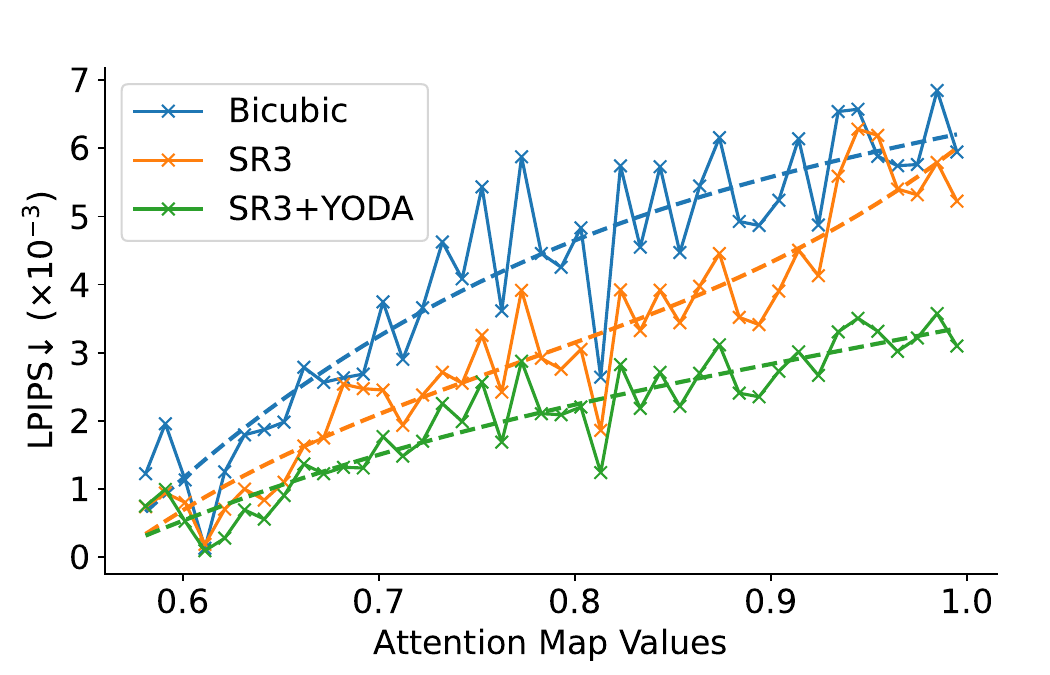}
        \caption{ \label{fig:lpips_with_polyfit}
        Regional LPIPS comparison across normalized attention values for CelebA, $64 \rightarrow 256$ (4×).
        We use 0.01 intervals and fit a polynomial through the means.
        High-attention areas are perceptually relevant and correspond to more difficult pixels (higher LPIPS).
        YODA reaches better scores, especially within high-attention areas.
        Note that dynamic masking stops around $t \approx 0.6 \cdot T$, see \autoref{fig:iterations}.
        }
    \end{center}
\end{figure}

\mysubparagraph{Analysis across attention regions}
In \autoref{fig:lpips_with_polyfit}, we analyze the LPIPS scores across different attention value intervals within a single attention map $\mathbf{A}$ (using the MAX aggregation, 0.01 interval size).
These intervals represent varying attention levels assigned by DINO to different regions of the image.
The high LPIPS scores associated with bicubic upsampling in high-attention areas underscore the effectiveness of DINO in capturing perceptually significant regions.
As a result, YODA significantly improves LPIPS scores across all attention regions, particularly in regions with higher attention values (highlighted by the trend curve).
Please refer to the appendix for more details.

\mysubparagraph{User study}
In addition to the quantitative results on $16 \rightarrow128$ (8×) provided in \autoref{fig:ablAttHeads} and inspired by Saharia et al., we conducted a user study.
We selected 50 random test images, asked 45 participants which SR prediction is preferred, and plotted the preferences per image (see appendix).
As a result, YODA was preferred 55.7\% of the time over SR3 (44.3\%). 
More details are in the appendix.

\begin{figure}[!t] 
\captionsetup[subfigure]{position=b}
   \begin{subfigure}[t]{.23\linewidth}
       \centering
       \includegraphics[width=\linewidth]{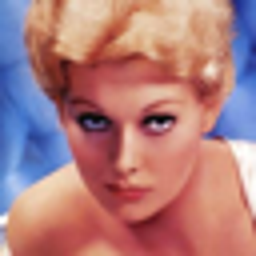}
       LR \\ $\uparrow$ PSNR: \\ $\uparrow$ SSIM: \\ $\downarrow$ LPIPS:
   \end{subfigure}
   \begin{subfigure}[t]{.23\linewidth}
       \centering
       \includegraphics[width=\linewidth]{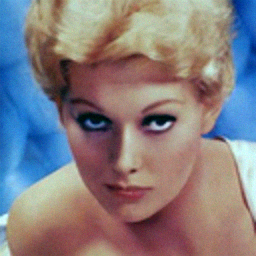}
       SR3 \\ 11.811 \\ 0.3311 \\ 0.2847
   \end{subfigure}
    \begin{subfigure}[t]{.23\linewidth}
       \centering
       \includegraphics[width=\linewidth]{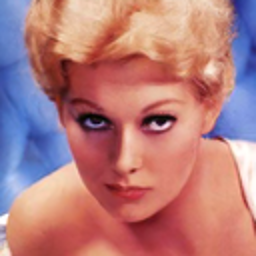}
       SR3+YODA \\ 18.069 \\ 0.6588 \\ 0.2628
   \end{subfigure}
   \begin{subfigure}[t]{0.23\linewidth}
       \centering
       \includegraphics[width=\linewidth]{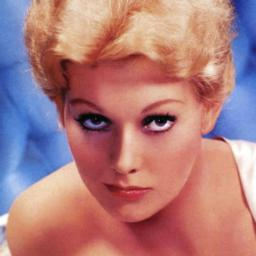}
       HR
    \end{subfigure}
   \caption{
   SR3 and SR3+YODA reconstructions, $64 \rightarrow 256$ (4×).
   SR3 suffers from color shifting, as also observed by \cite{wang2023exploiting, choi2022perception}.
   YODA solves this issue and produces higher-quality reconstructions.
   }
   \label{fig:image64by64}
\end{figure}

\begin{table}[t]
    \centering
    \resizebox{.85\columnwidth}{!}{%
    \begin{tabular}{c l c c c c}
    \toprule
    \textbf{Scaling} & \textbf{Model} & \textbf{PSNR} $\uparrow$ & \textbf{SSIM} $\uparrow$ & \textbf{LPIPS} $\downarrow$  & \textbf{FID} $\downarrow$\\
    \midrule
       4× & SR3 \cite{saharia2022image} & 28.3 & 0.67 & 0.10 & 74.99\\
       & \textbf{SR3 + YODA}  & \textbf{31.6} & \textbf{0.87} & \textbf{0.05}  & \textbf{58.34}
       \\
    \midrule
    8× & SR3 \cite{saharia2022image} & 29.3 & 0.71 & 0.10 & 55.94\\
       & \textbf{SR3 + YODA}  & \textbf{31.5} & \textbf{0.84} & \textbf{0.07} & \textbf{49.02}
       \\
    
    \bottomrule
    \end{tabular}
    }
    \caption{\label{tab:faceSR_color} Results without color-shifting on face SR by normalizing the channel-wise means to those of the HR ground-truth data for 4× scaling ($64 \rightarrow 256$) and 8× scaling ($64 \rightarrow 512$). }
\end{table}

\subsection{General Super-Resolution}
The experimental setup follows SRDiff \cite{li2022srdiff}, based on the setup of SRFlow and bicubic degradation \cite{lugmayr2020srflow}.
For training, we employed 800 2K resolution images from DIV2K \cite{agustsson2017ntire} and 2,650 from Flickr2K \cite{timofte2018ntire}.
For testing, we used the DIV2K validation set (100 images).
Moreover, we evaluated SR3, which was not originally tested on DIV2K. 

\begin{figure}[t!]
    \begin{center}
        \includegraphics[width=\linewidth]{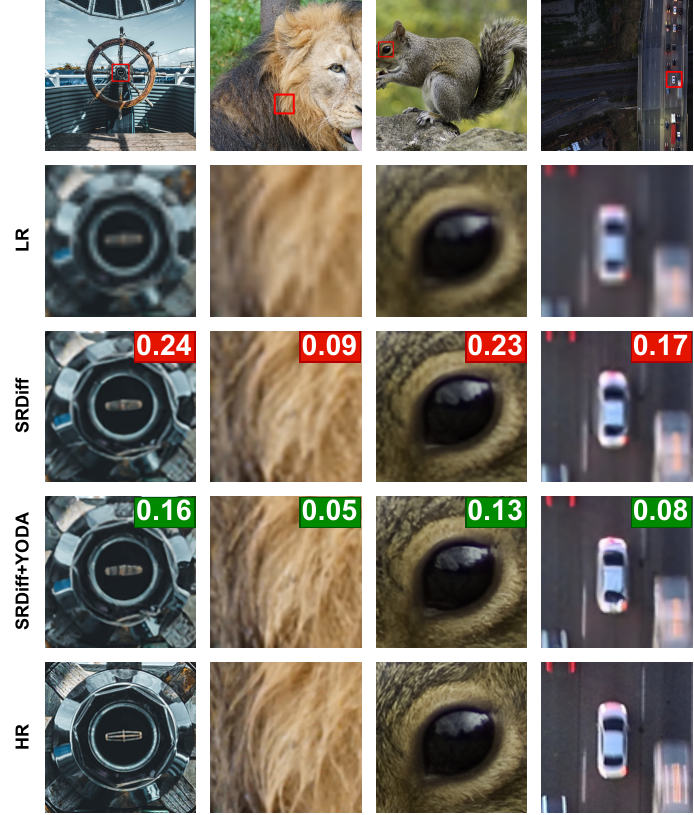}
        \caption{ \label{fig:div2k}
        Zoom-in regions of DIV2K images (first row). 
        LPIPS is reported in the boxes (the lower, the better). 
        YODA consistently produces better texture and more high-frequency details.
        }
    \end{center}
\end{figure}

\mysubparagraph{Results}
\autoref{tab:div2k_results} shows the 4× general SR results on the DIV2K val.
The reported values include regression-based methods, which typically yield higher pixel-based scores than generative models \cite{saharia2022image}.
The disparity is due to PSNR/SSIM penalizing misaligned high-frequency details, a known challenge in the wider SR field \cite{10041995}.
We observe that results from SR3 underperform without YODA.
We hypothesize that SR3 strongly depends on larger batch sizes and longer diffusion times.
Combining SRDiff with YODA improves the quality in PSNR (+0.21db) and SSIM (+0.01), with a minor deterioration in LPIPS (+0.01).
Nonetheless, when looking qualitatively at the predictions, one can observe that SRDiff strongly benefits from YODA, as shown in \autoref{fig:div2k}.
YODA produces much better LPIPS values for perceptually essential areas, i.e., hair and cars, but falls short in background areas, i.e., blurry grass or dark ground.
The appendix contains more visual results.

\mysubparagraph{Discussion}
Overall, YODA's strengths are more significant when combined with SR3 than with SRDiff.
A critical distinction between SR3 and SRDiff is the handling of conditional information, i.e., the LR image, which we identify as a potential contributor to the reduced perceptual score LPIPS.
SRDiff employs an LR encoder that generates an embedding during the denoising phase. 
Meanwhile, SR3 directly uses the LR image during the backward diffusion.

\begin{table}[t]
\center
\resizebox{.9\columnwidth}{!}{%
\begin{tabular}{l l c c c }
\toprule
\textbf{Type} & \textbf{Methods}& \textbf{PSNR} $\uparrow$ & \textbf{SSIM} $\uparrow$& \textbf{LPIPS} $\downarrow$ \\ 
\midrule
Interpolation & Bicubic & 26.70 & 0.77 & 0.409  \\
\midrule
Regression & EDSR \cite{lim2017enhanced} & 28.98 & 0.83 & 0.270  \\
& LIIF \cite{chen2021learning} & 29.24 & 0.84 & 0.239 \\
& RRDB \cite{wang2018esrgan} & 29.44 & 0.84 & 0.253  \\ 
\midrule
GAN &RankSRGAN \cite{zhang2019ranksrgan} & 26.55 & 0.75 & 0.128  \\ 
& ESRGAN \cite{wang2018esrgan} & 26.22 & 0.75 & 0.124  \\
\midrule
Flow & SRFlow \cite{lugmayr2020srflow} & 27.09 & 0.76 & 0.120  \\
& HCFlow \cite{liang2021hierarchical} & 27.02 & 0.76 & 0.124 \\
\midrule
Flow + GAN & HCFlow++ \cite{liang2021hierarchical} & 26.61 & 0.74 & 0.110 \\
\midrule
VAE + AR & LAR-SR \cite{guo2022lar} & 27.03 & 0.77 & 0.114 \\
\midrule
Diffusion & SR3 \cite{saharia2022image} & 14.14 & 0.15 & 0.753  \\
& \textbf{SR3 + YODA} & \textbf{27.24} & \textbf{0.77} & \textbf{0.127}  \\
\cmidrule{2-5}
& SRDiff \cite{li2022srdiff} & 27.41 & 0.79 & \textbf{0.136}  \\
& \textbf{SRDiff + YODA} & \textbf{27.62} & \textbf{0.80} & 0.146\\ 
\bottomrule
\end{tabular}
}
\caption{Quantitative results of 4× general SR on DIV2K val.
YODA improves across pixel-centric metrics such as PSNR and SSIM, but yields a marginal decline in LPIPS.
}
\label{tab:div2k_results}
\end{table}

\section{Conclusion}
In this work, we presented ``You Only Diffuse Areas" (YODA), an attention-guided diffusion-based image SR approach that emphasizes essential areas through time-dependent masking. 
YODA first extracts attention maps that reflect the pixel-wise saliency of each scene using a self-supervised, general-purpose vision encoder.
The attention maps are then used to guide the diffusion process by focusing on key regions in each time step while providing a fusion technique to ensure that masked and non-masked image regions are correctly connected between two successive time steps.
This targeting allows for a more efficient transition to high-resolution outputs, prioritizing areas that gain the most from iterative refinements, such as detail-intensive regions.
Beyond better performance, YODA stabilizes training and mitigates the color shift issue when a reduced batch size constraints the underlying diffusion model.
As a result, YODA consistently outperforms strong baselines like SR3, DiffBIR, and SRDiff while requiring less computational resources by using smaller batch sizes. 

\section{Limitations \& Future Work}
A notable constraint of this study is its dependence on a good saliency estimation. 
Even though DINO is trained to be a generic vision encoder, it has known limitations that will reflect on the quality of YODA \cite{darcet2023vision}. 
Additionally, DINO is explicitly trained on input image resolutions such as $224 \times 224$, which may not suffice for image SR applications with much larger spatial sizes of the LR image. 
Meanwhile, the modularity of YODA allows for the saliency model to be switched once a better one becomes available.
An ideal solution would be a scale-invariant extraction of attention maps, e.g., a more extended version of our SIFT-adapted approach. 
Lastly, YODA can be extended for other diffusion-based methods, e.g., latent-based methods like LDM \cite{rombach2022high} or unsupervised methods based on singular value decomposition like DDRM \cite{kawar2022denoising} or DDNM \cite{wang2022zero}, which is orthogonal to our work (see appendix for more details).

\section{Societal Impact}
YODA can significantly benefit fields like medical imaging and remote sensing by improving visual quality. 
Yet, high-quality SR can also be exploited maliciously by adding realism to deceptive or misleading media content.
Using SR methods responsibly and promoting transparency and ethical guidelines in deployment is crucial.
Also, the reliance on pre-trained models for attention maps, such as DINO, may introduce biases inherent in the training data. 
These biases could affect the quality and fairness of the SR results, particularly in diverse or underrepresented populations. 
Future work should aim to mitigate these biases.

\section*{Acknowledgment}
This work was supported by the EU project SustainML (Grant 101070408) and by Carl Zeiss Foundation through the Sustainable Embedded AI project (P2021-02-009).

{\small
\bibliographystyle{ieee_fullname}
\bibliography{egbib}
}

\newpage
\section*{Appendix}

In the main text, we presented YODA as an approach to dynamically focus the diffusion process to essential areas in the image. The supplementary material hereby gives further information and visualizations on YODA, such as a discussion on related work, details on DINO, training details, and complexity analysis. It supports understanding the main concepts and ideas examined in the main text.

\subsection*{Related Work}
In this section, we discuss other diffusion models that can be applied to YODA and content-aware SR methods that also focus on image content to optimize certain properties in the SR pipeline.

\subsubsection*{Other State-Of-The-Art Diffusion Models}
\begin{table}
\centering
    \caption{\label{tab:ldm_table}  $\times 4$ upscaling results on ImageNet-Val. ($256 \times 256$). Values directly derived from the original work of LDM \cite{rombach2022high}.}
    \resizebox{\columnwidth}{!}{
  \begin{tabular}{l c c c c }
  	\toprule
  	\textbf{Method} & IS $\uparrow$ & PSNR  $\uparrow$  & SSIM  $\uparrow$ \\
  	\midrule
  	Image Regression \cite{saharia2022image}& 121.1 & \textbf{27.9} & \textbf{0.801}\\
  	SR3 \cite{saharia2022image}& \textbf{180.1} & \underline{26.4} & \underline{0.762}\\
  	\midrule
  	LDM-4 (100 steps) \cite{rombach2022high}& 166.3 & 24.4\tiny{$\pm $3.8} & 0.69\tiny{$\pm $0.14}\\
  	LDM-4 (big, 100 steps) \cite{rombach2022high}& \underline{174.9} & 24.7\tiny{$\pm $4.1} & 0.71\tiny{$\pm $0.15}\\
  	LDM-4 (50 steps, guiding) \cite{rombach2022high}& 153.7 & 25.8\tiny{$\pm $3.7} & 0.74\tiny{$\pm $0.12}\\

  	\bottomrule
  \end{tabular}
  }
\end{table}
\noindent
As shown in the study of Moser et al. \cite{moser2024diffusion}, many approaches apply to image SR. 
In this section, we want to discuss their potential in combination with YODA and possible limitations for future work.

\textbf{Latent Diffusion Models.} Despite the significant advancements brought by Latent Diffusion Models (LDMs) \cite{rombach2022high}, their efficacy in the realm of image SR competes closely with that of SR3 \cite{saharia2022image}, as shown in \autoref{tab:ldm_table}.
Unfortunately, recent research in this direction focused primarily on text-to-image tasks \cite{frolov2021adversarial}, which makes further comparisons with image SR methods challenging, e.g., SDXL \cite{podell2023sdxl}, MultiDiffusion \cite{bar2023multidiffusion}, or DemoFusion \cite{du2023demofusion}.
Nevertheless, their potential for image SR is undeniable.
Concerning YODA, we also see great research avenues in combination with LDMs. 
A critical prerequisite for this synergy is the conversion of attention maps from pixel to latent representations.
This aspect has to be investigated in more detail in future work.
Our preliminary StableSR (10 epochs in stage 2) results show that our method improves LPIPS from 0.1242 to 0.1239.

\textbf{Unsupervised Diffusion Models.} Another interesting research avenue is unsupervised diffusion models for image SR, exemplified by DDRM \cite{kawar2022denoising} or DDNM \cite{wang2022zero}. 
Interestingly, they use a pre-trained diffusion model to solve any linear inverse problem, including image SR, but they rely on singular value decomposition (SVD).
Similar to the challenges faced with LDMs, integrating YODA into unsupervised diffusion models presents another interesting research avenue. 
The core of this challenge lies in devising a method for effectively translating attention maps into a format compatible with the SVD process used by these models. 
This transformation is crucial for harnessing the power of attention-based enhancements in unsupervised diffusion frameworks for image SR. 
Future work needs to conceptualize and implement a seamless integration strategy that combines the dynamic attention modulation offered by YODA with SVD.

\textbf{Alternative Corruption Spaces.}
Applying YODA with alternative corruption spaces (not pure Gaussian noise), such as used in InDI \cite{delbracio2023inversion}, I$^2$SB \cite{liu20232}, CCDF \cite{chung2022come}, or ColdDiffusion \cite{bansal2024cold} is also an interesting future research direction.
Although our primary focus has been refining and enhancing the diffusion process of specific models through attention-guided masks, we acknowledge the orthogonal potential these approaches represent within the broader context of SR.

\subsubsection*{Other Content-Aware SR Methods}
One category of approaches that draw similarities to YODA is dataset pruning for image SR.
Commonly, image SR methods are trained on sub-images cropped from higher-resolution counterparts, such as those found in DIV2K \cite{agustsson2017ntire}. 
The central premise behind dataset pruning strategies is the observation that not all sub-images contribute equally to training efficacy. 
These approaches employ content-aware detectors to prune training data based on metrics like loss values selectively \cite{moser2024study} or Sobel norms \cite{ding2023not}.

A related strategy, ClassSR \cite{wang2022adaptive}, categorizes different image regions into three levels of reconstruction difficulty - easy, medium, and hard. 
They propose training specialized models for each category.
Related to ClassSR are RAISR \cite{romano2016raisr}, SFTGAN \cite{wang2018recovering}, RL-Restore \cite{yu2018crafting}, and PathRestore \cite{yu2021path}.
RAISR \cite{romano2016raisr} assigns image patches into clusters and employs a tailored filter to each cluster, utilizing an efficient hashing technique to streamline the clustering process. 
SFTGAN \cite{wang2018recovering} introduces a spatial feature transform layer that embeds high-level semantic priors, enabling nuanced processing of different image regions with different parameters (i.e., different models). 
Similarly, RL-Restore \cite{yu2018crafting} and PathRestore \cite{yu2021path} divide images into sub-images and employ reinforcement learning to determine the optimal processing pathway for each sub-image. 
Unlike YODA, which focuses on refining specific image regions with one model, the presented works aim to optimize training datasets by reduction or categorization (thereby employing multiple models tailored to varied reconstruction difficulties).

Similarly, the Multiple-in-One Image Restoration (MiO IR) strategy introduces a novel approach to handling diverse image restoration tasks within a single model \cite{kong2024towards}. 
MiO IR employs sequential learning, which allows the model to learn different tasks incrementally and optimize for diverse objectives. 
Additionally, it utilizes prompt learning - both explicit and adaptive - to guide the model in adapting to various tasks dynamically.
While YODA focuses on refining specific image regions, MiO IR’s versatility across tasks offers a potential avenue for expanding YODA's application to more generalized scenarios.

SkipDiff \cite{luo2024skipdiff} presents another content-aware SR approach in the context of diffusion models. 
This method operates through two primary phases: a coarse skip approximation and a fine skip refinement. 
SkipDiff constructs a preliminary high-resolution image approximation in the first phase in a single step. 
In the second phase, this image is refined using the classical diffusion pipeline with an adaptive noise schedule. 
For this, they employ a schedule driven by the characteristics of the input image. Reinforcement learning is integral to this process, as it is trained to find optimal diffusion steps for this phase. 
This adaptability enables SkipDiff to tailor diffusion to the entire content of an image, contrasting with YODA’s targeted refinement of specific regions.
Future research might explore the potential synergies between YODA and SkipDiff, combining their strengths to further enhance content-specific image SR.

Another recent approach in lightweight SR is the Self-Feature Learning (SFL) method proposed by Xiao et al. \cite{xiao2021self}. 
This method introduces a locally adaptive involution technique that reduces computational costs by dynamically generating convolutional kernels based on local image content. 
The SFL model achieves a remarkable trade-off between performance and model complexity by avoiding inter-channel redundancy, making it particularly suitable for resource-constrained devices. 
Unlike YODA, which focuses on targeted region refinement, SFL employs a dual-path residual module to ensure efficient feature extraction across the entire image, potentially complementing YODA’s targeted strategy.

\subsection*{Details: DINO}
DINO \cite{caron2021emerging} is a self-supervised learning approach. involving a teacher and student network.
While both networks share the same architecture, their parameters differ.
The student network is optimized to match the teacher's output via cross-entropy loss.
During training, both receive two random views of the same input image: the teacher is trained on global views, i.e., $224 \times 224$ crops, while the student receives local views, i.e., $96 \times 96$ crops.
This setup encourages the student to learn ``local-to-global'' correspondences.
In other words, by predicting the teacher's output, the student learns to infer global information from local views.
To prevent mode collapse, the teacher's parameters are updated as a moving average of the student's parameters.

\subsection*{Details: Training Parameters}
For SR3, we adopted the AdamW \cite{loshchilov2017decoupled} optimizer, using a weight decay of 0.0001 and a learning rate of 5e-5. 
The number of sampling steps is set to $T_{\text{train}}=500$.
The number of sampling steps is set to $T_{\text{eval}}=200$.
Concerning the denoising architecture, our approach aligns with the SR3 model \cite{saharia2022image}, but we employed residual blocks \cite{he2016deep} proposed by Ho et al. instead of those used in BigGAN \cite{ho2020denoising, brock2018large}. 
Specifically, the configuration includes three ResBlocks, an initial channel size of 64, and a channel multiplier array of [1, 2, 4, 8, 8]. We also employed a NormGroup with a size of 32.
For SRDiff, we extracted $40\times40$ sub-images with a batch size of 16, AdamW \cite{loshchilov2017decoupled}, a channel size of 64 with channel multipliers [1, 2, 2, 4] and $T=100$.

\subsection*{Details: Complexity of YODA}
We discuss the resource implications of the core components of YODA: Identifying key regions, time-dependent masking, and the guided diffusion process. Additionally, we explore potential avenues for future enhancements aimed at optimizing computational efficiency.

\textbf{Identifying Key Regions.} To avoid the computational burden of on-the-fly generation, we pre-compute the attention maps prior to training.
\autoref{tab:archs} shows the parameter count and throughput of different DINO backbones.
While YODA introduces additional complexity for setting up the attention maps, the overhead is minimal. 
For instance, generating all attention maps for our face SR experiments (i.e., 120,000 images) needed less than two minutes. 

\textbf{Time-Dependent Masking and Guided Diffusion Process.} The integration of attention masks within the diffusion framework introduces minimal computational overhead, thanks to the inherently parallelizable nature of element-wise multiplication and addition, as demonstrated in the methodology (see time-dependent masking). 
Consequently, the predominant factor influencing the overall computational complexity remains the choice of diffusion model, whether it be SR3, SRDiff, or another variant.

\textbf{Potential Future Improvements.} YODA notably decreases computational requirements by enabling the use of smaller batch sizes during training, which in turn reduces VRAM usage without compromising performance. 
Looking ahead, YODA paves the way for leveraging sparse diffusion techniques. 
Such approaches promise further computational savings by focusing computation efforts on selectively identified regions (through YODA), thereby streamlining the diffusion process.
Currently, in PyTorch, applying masks to regions within a matrix does not result in computational savings.

\begin{table}[h]
\centering
\caption{
Details of different DINO backbones, values directly extracted from the original work \cite{caron2021emerging}.
Throughput was measured with a NVIDIA V100 GPU.
}
	\label{tab:archs}
  \setlength{\tabcolsep}{2pt}
  \begin{tabular}{l c c}
\toprule
	  Model & Parameters & Throughput \\
        & [M] & [img/s] \\
    \midrule
	  ResNet-50 & 23M & 1237 \\
	  ViT-S/8 & 21M & 180 \\
\bottomrule
  \end{tabular}
\end{table}

This examination of YODA's complexity highlights its efficiency and the strategic decisions made to balance performance with computational demands. 

\subsection*{Details: Analysis Across Attention Regions}
This section explains how we created the figure associated with the subsection \textbf{Analysis across attention regions}.
The used attention maps are derived from DINO+ResNet. 
We apply the max aggregation explained in the main paper to create a single attention map per image.
The aggregated map is then divided into several attention-value ranges (bins/intervals). 
Specifically, the attention values range from 0 to 1 (see methodology), and we divide this range into small bins with a step size of 0.01. 
For each bin, we analyze the regions of the image where the attention values fall within the bin range (e.g., 0.01-0.02, 0.02-0.03, etc.). 
This allows us to observe how different attention levels correlate with the LPIPS scores and how each SR model performs in these non-overlapping and separated regions.

For each attention value bin, we calculate the LPIPS score between the reference HR image and the output images generated by the different SR techniques (SR3, SR+YODA, and LR/bicubic upsampling). This is done as follows:
\setlist{nolistsep}
\begin{itemize}[noitemsep]
    \item \textbf{Mask Creation:} For each attention bin, we create a binary mask where pixels in the attention map that fall \underline{within} the bin are set to 1, and all others are set to 0.
    \item \textbf{Region Extraction:} Using this binary mask, we extract the corresponding regions from the HR and SR images.
    \item \textbf{LPIPS Calculation:} We compute the LPIPS score between the HR image's masked regions and the generated images' masked regions. 
    This process is repeated for each bin across the entire attention range.
    Masked-out regions do not influence LPIPS because LPIPS is a distance measure of features and masked-out regions in HR, and any SR image leads to an LPIPS of 0.
\end{itemize} 

We calculate the mean LPIPS error within attention bins across the images.
A polynomial curve (degree 3) is fitted to the LPIPS scores for each model across the attention value ranges to visualize trends more clearly. This allows us to smooth out potential noise and outliers in the data and observe how each model's performance changes as we move from high to low-attention regions.
Key observations:
\setlist{nolistsep}
\begin{itemize}[noitemsep]
    \item \textbf{Higher LPIPS in High Attention Regions:} Bicubic upsampling performs poorly in high-attention regions, as indicated by the high LPIPS values. This shows that detail-rich and perceptually important regions are indeed reflected by attention values, as simple upsampling cannot adequately capture the fine details.
    \item \textbf{YODA's Improvement:} YODA shows the most significant improvement in high-attention regions, confirming that YODA refines detail-rich regions more effectively.
    Moreover, it also shows that YODA is improving every region, thereby truly super-resolving every pixel instead of just replacing the LR and leaving it unchanged.
\end{itemize} 
As DINO+ResNet starts to refine the whole image without dynamic masking at around 0.6, the figure stops there (as it investigated all separated regions by then).

\subsection*{More Visualizations}
\begin{figure*}[t!] 
\captionsetup[subfigure]{position=b}
    
    \begin{subfigure}[t]{0.11\linewidth}
       \centering
       \includegraphics[width=\linewidth]{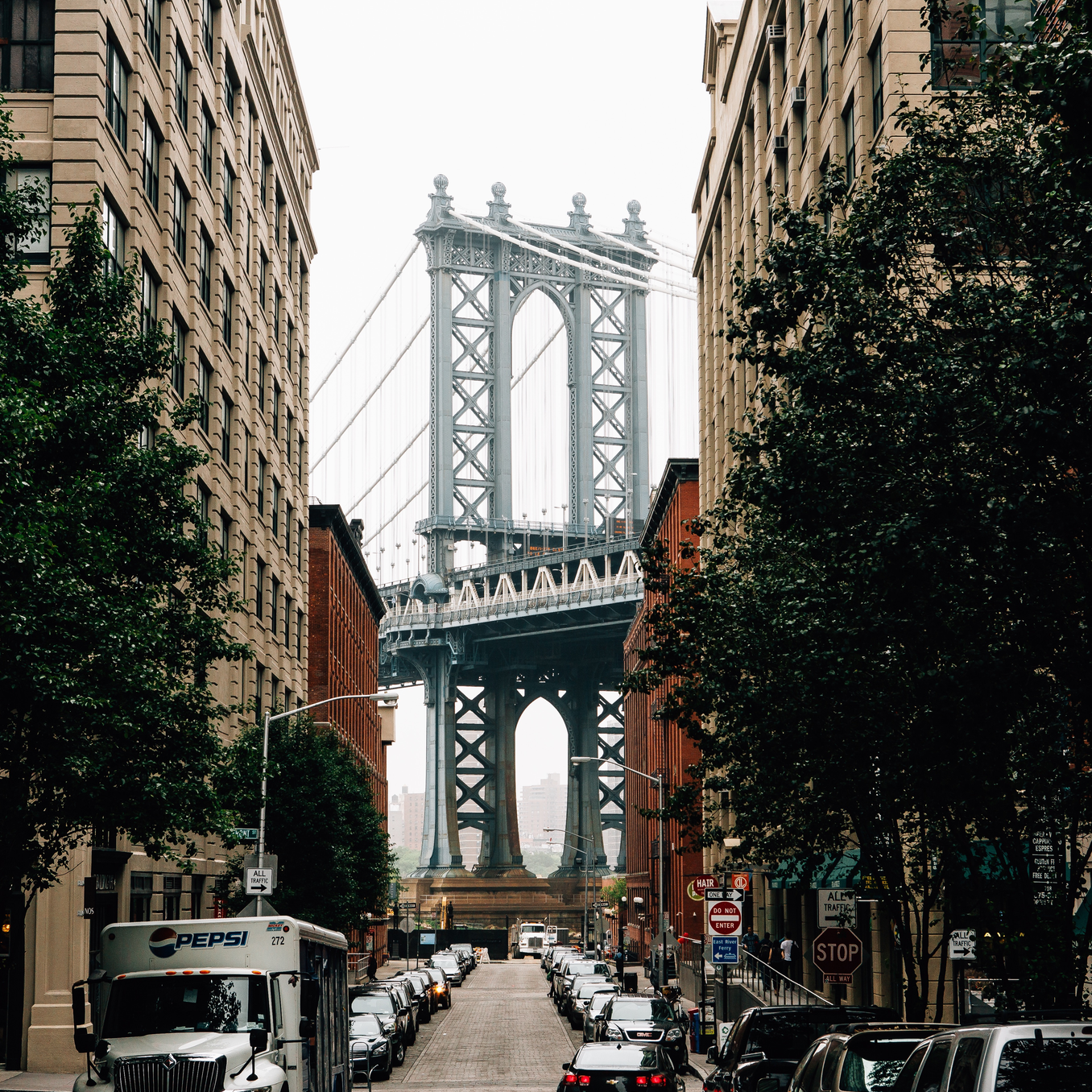}
   \end{subfigure}
   \begin{subfigure}[t]{0.11\linewidth}
       \centering
       \includegraphics[width=\linewidth]{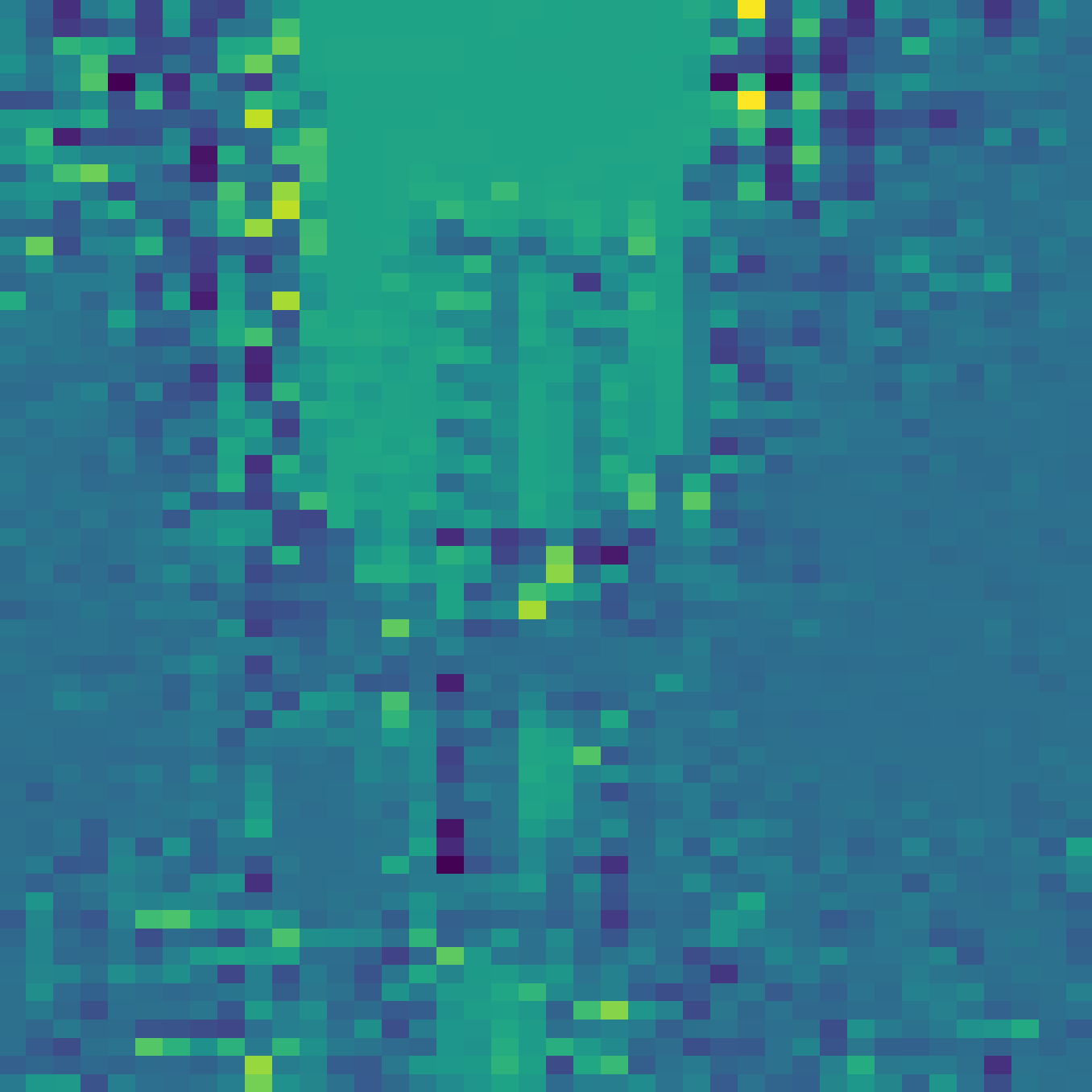}
   \end{subfigure}
   \begin{subfigure}[t]{0.11\linewidth}
       \centering
       \includegraphics[width=\linewidth]{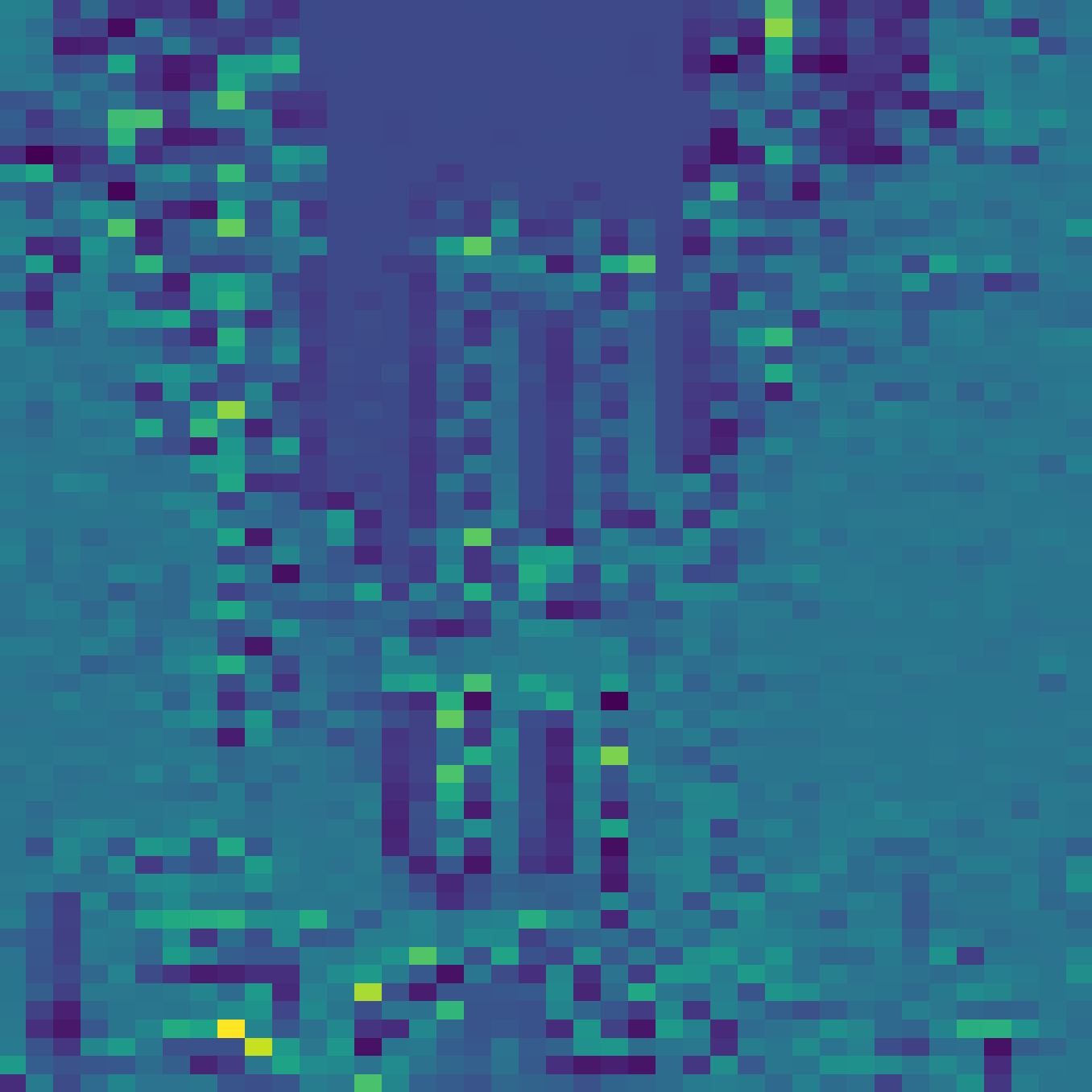}
   \end{subfigure}
   \begin{subfigure}[t]{0.11\linewidth}
       \centering
       \includegraphics[width=\linewidth]{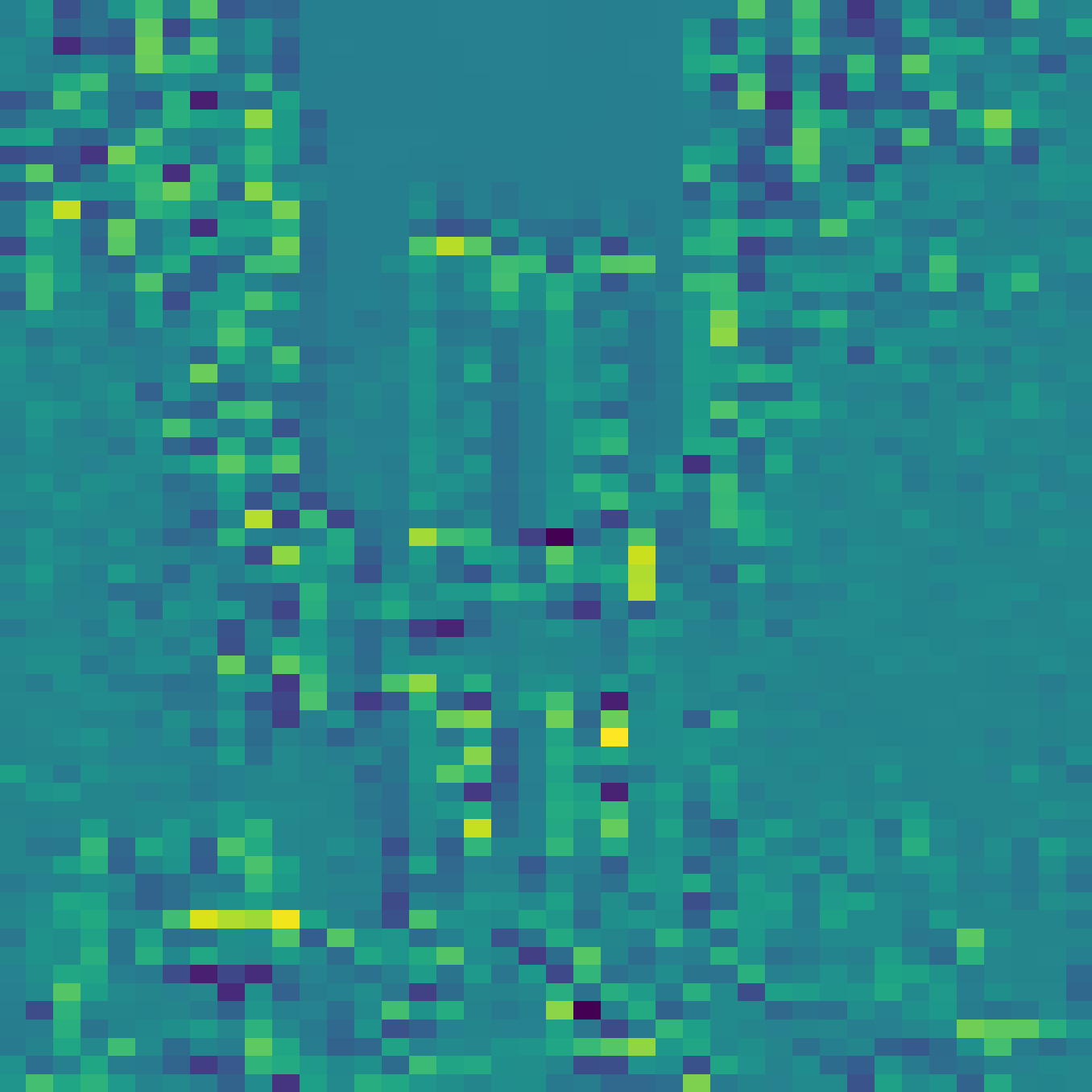}
   \end{subfigure}
   \begin{subfigure}[t]{0.11\linewidth}
       \centering
       \includegraphics[width=\linewidth]{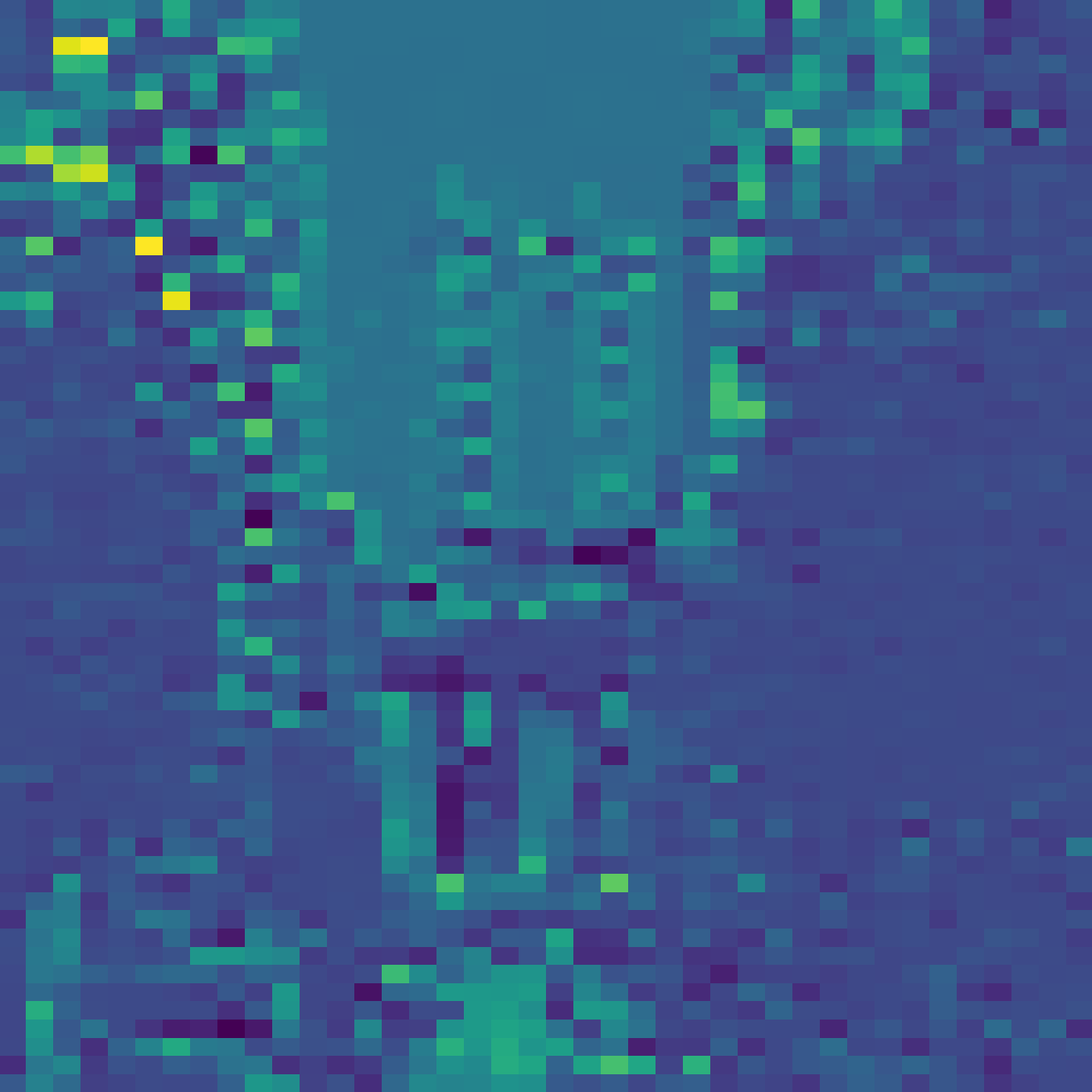}
   \end{subfigure}
   \begin{subfigure}[t]{0.11\linewidth}
       \centering
       \includegraphics[width=\linewidth]{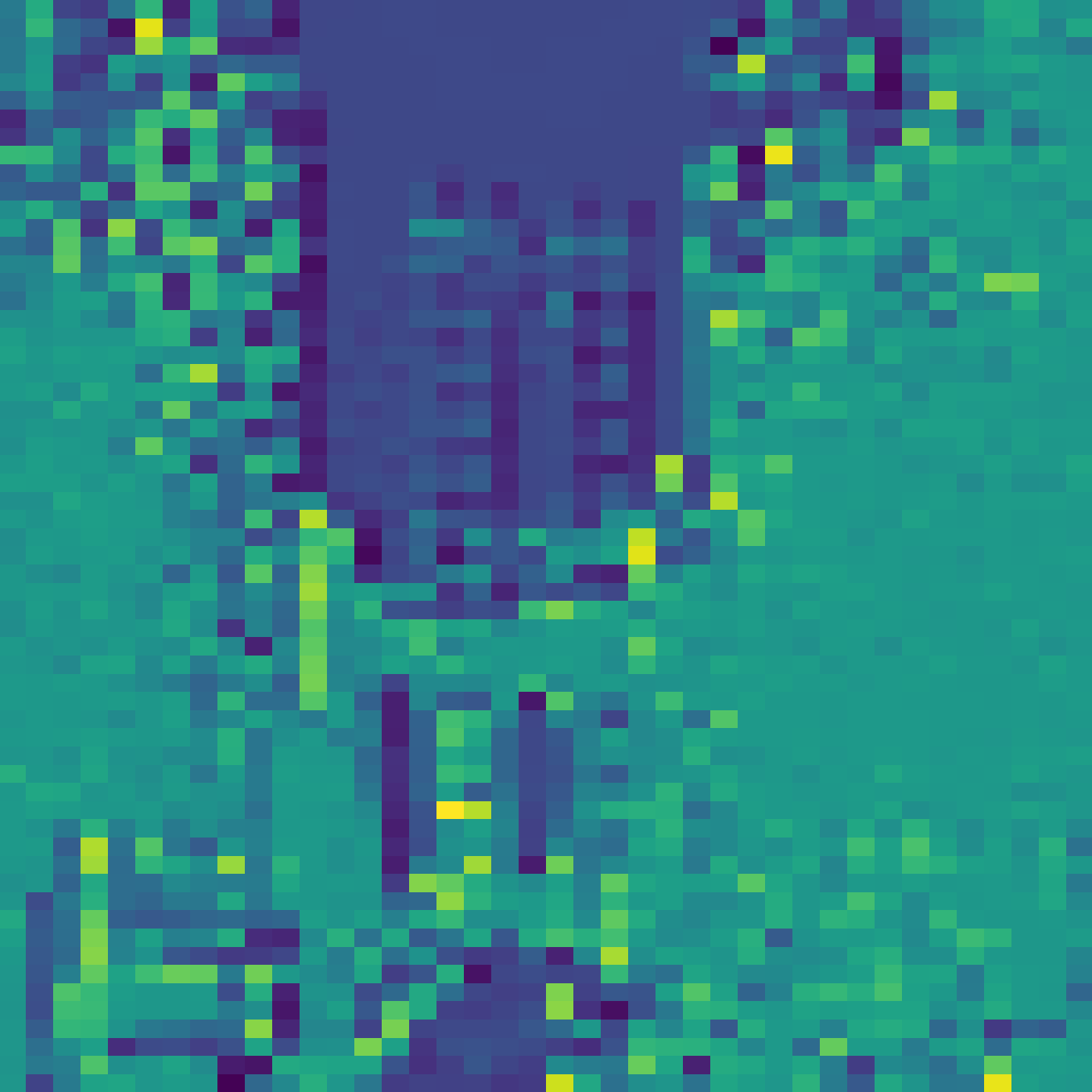}
   \end{subfigure}
   \begin{subfigure}[t]{0.11\linewidth}
       \centering
       \includegraphics[width=\linewidth]{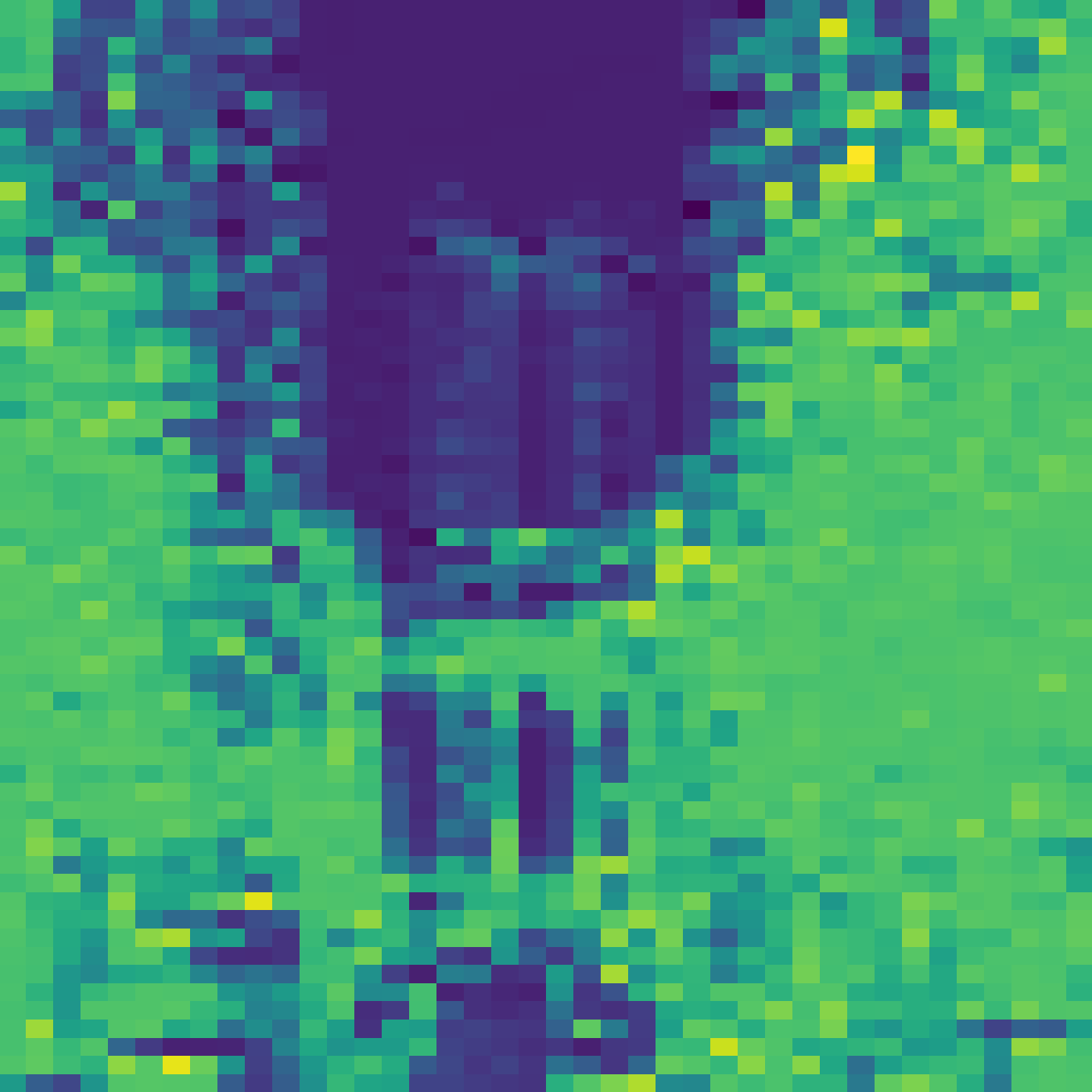}
   \end{subfigure}
   \begin{subfigure}[t]{0.11\linewidth}
       \centering
       \includegraphics[width=\linewidth]{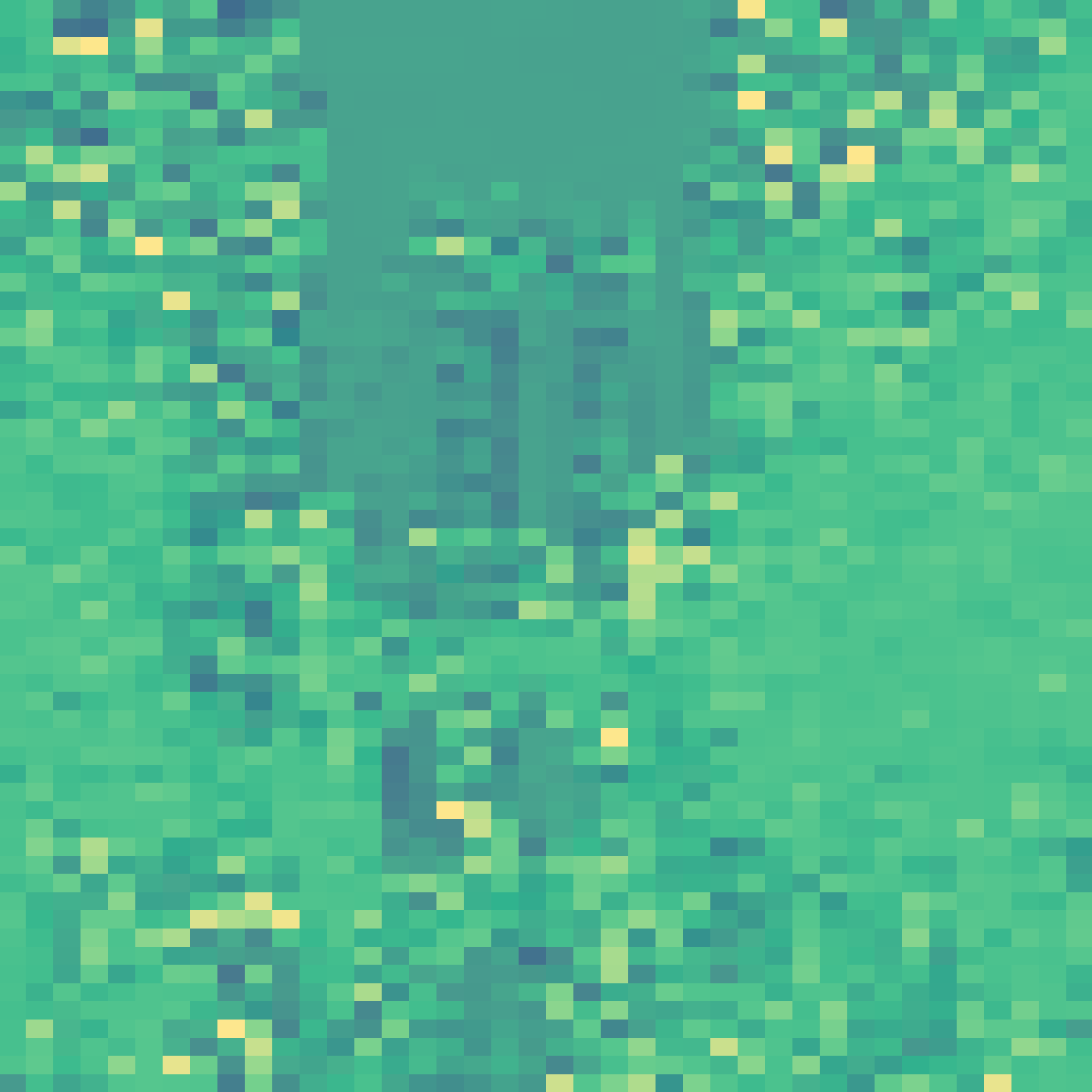}
   \end{subfigure}
    \begin{subfigure}[t]{0.11\linewidth}
       \centering
       \includegraphics[width=\linewidth]{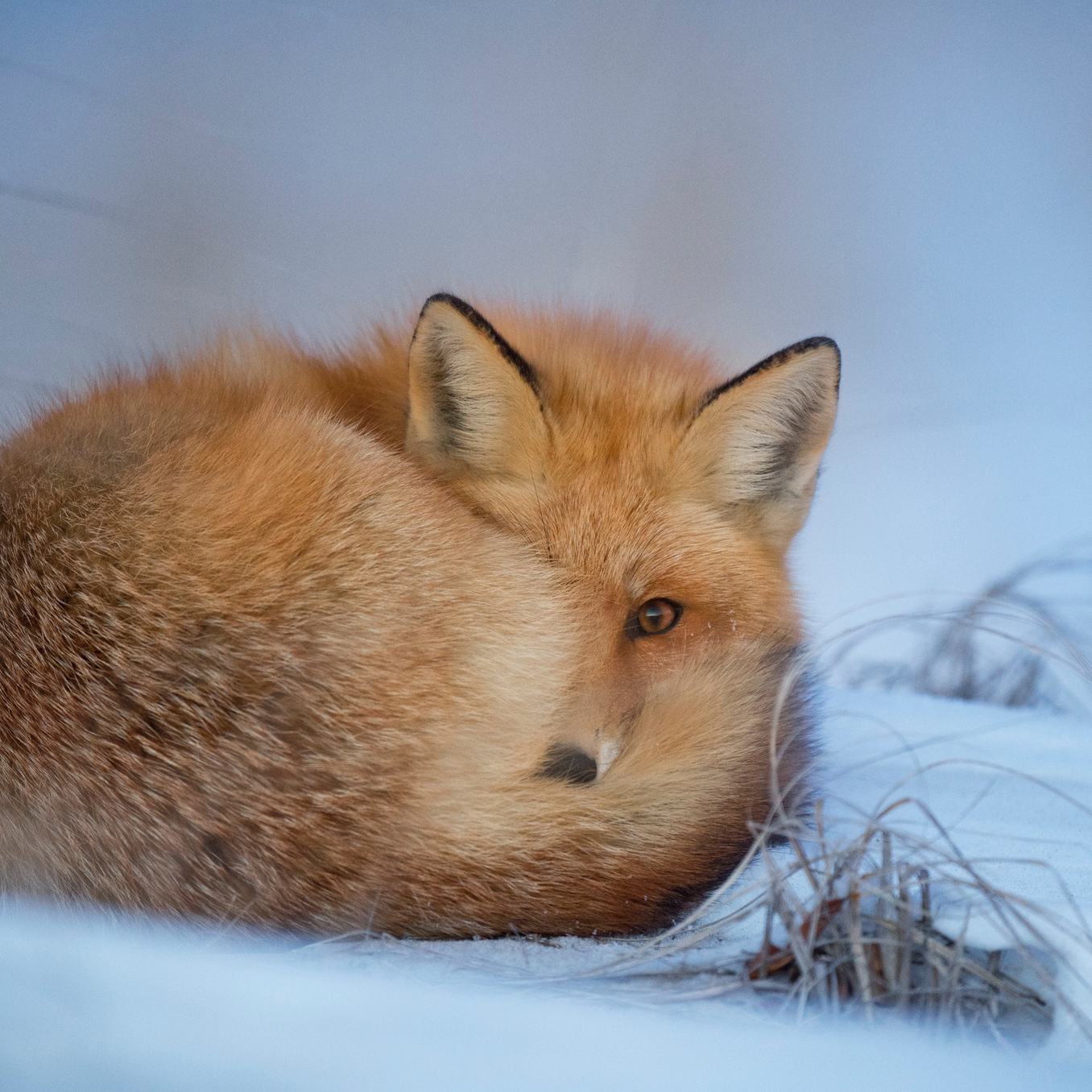}
   \end{subfigure}
   \begin{subfigure}[t]{0.11\linewidth}
       \centering
       \includegraphics[width=\linewidth]{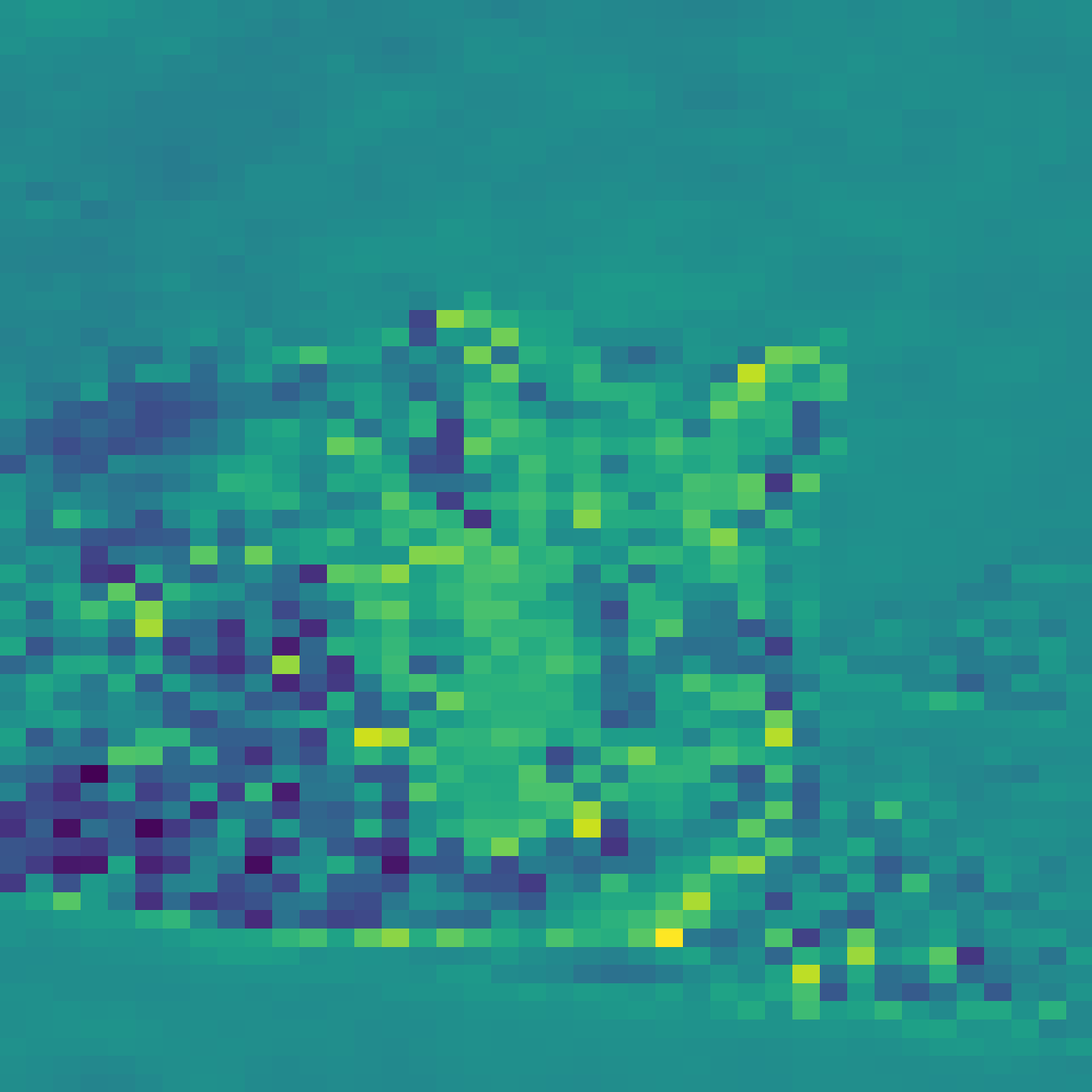}
   \end{subfigure}
   \begin{subfigure}[t]{0.11\linewidth}
       \centering
       \includegraphics[width=\linewidth]{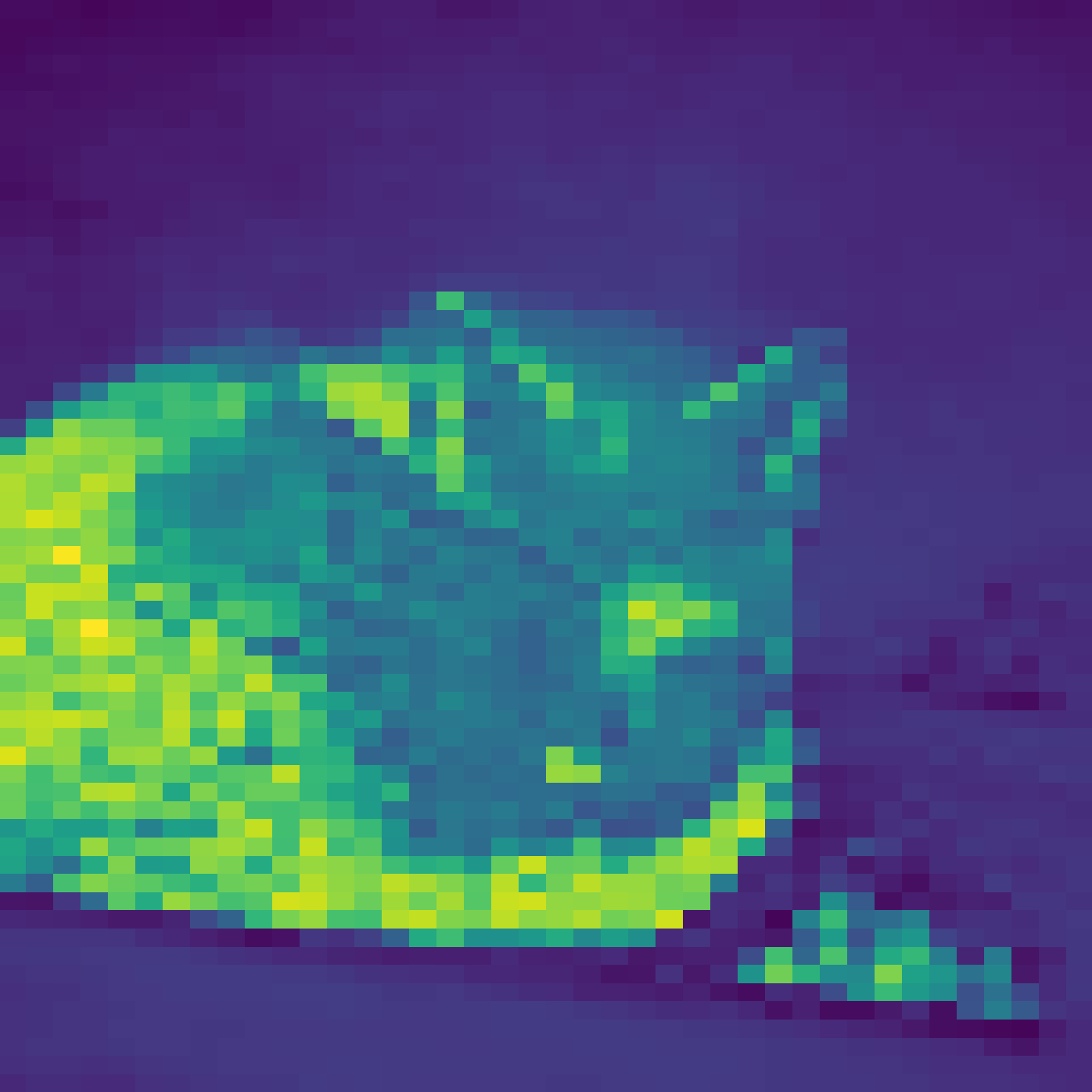}
   \end{subfigure}
   \begin{subfigure}[t]{0.11\linewidth}
       \centering
       \includegraphics[width=\linewidth]{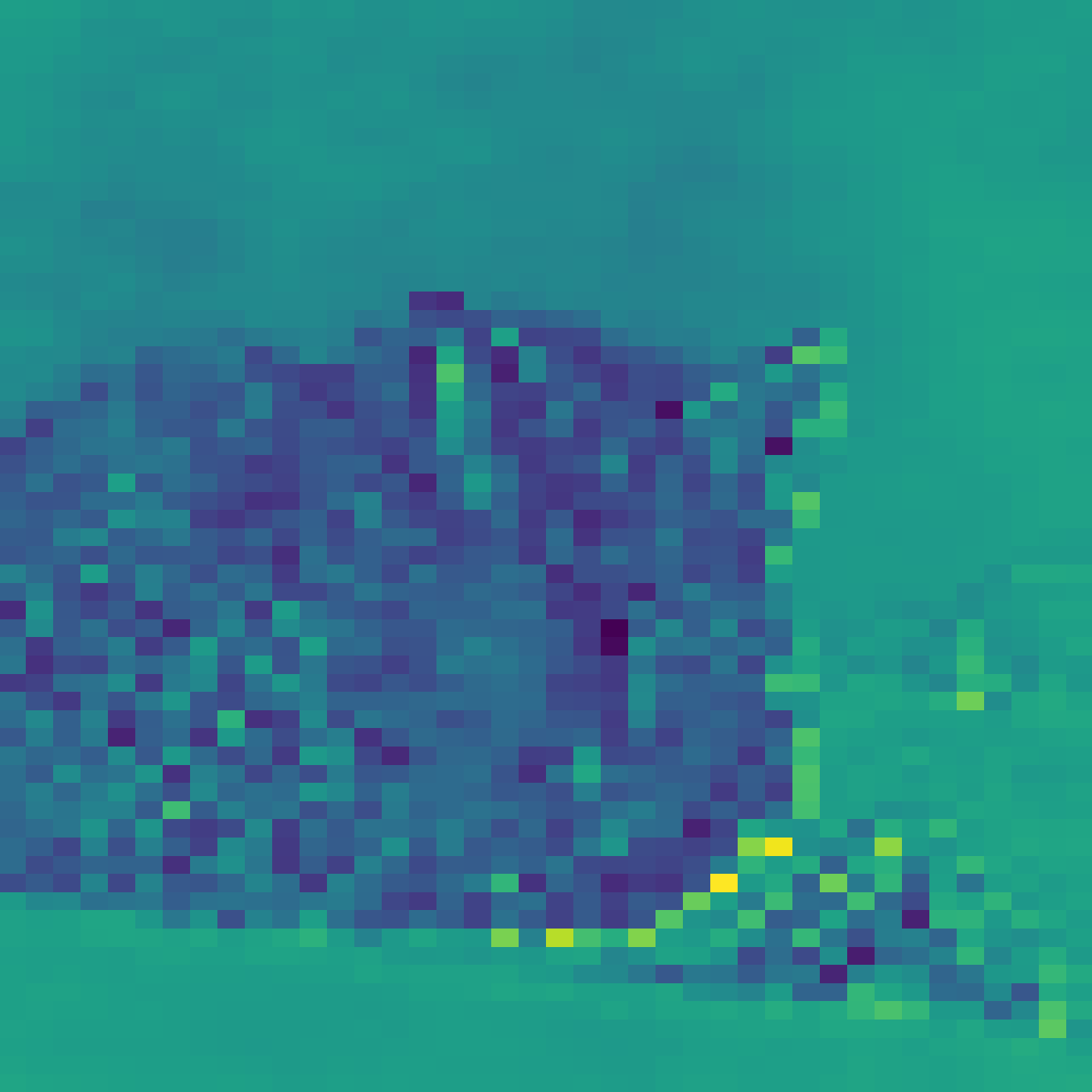}
   \end{subfigure}
   \begin{subfigure}[t]{0.11\linewidth}
       \centering
       \includegraphics[width=\linewidth]{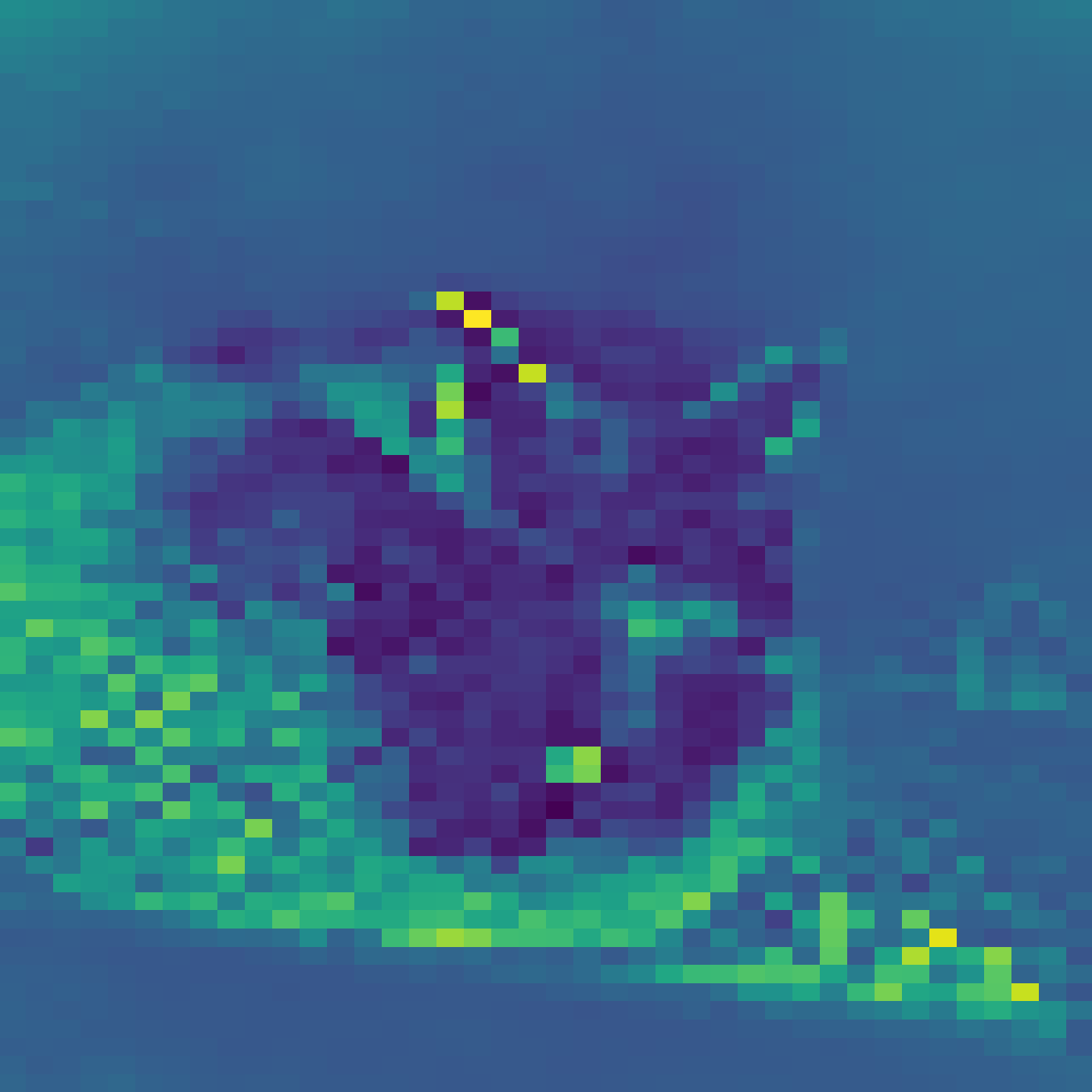}
   \end{subfigure}
   \begin{subfigure}[t]{0.11\linewidth}
       \centering
       \includegraphics[width=\linewidth]{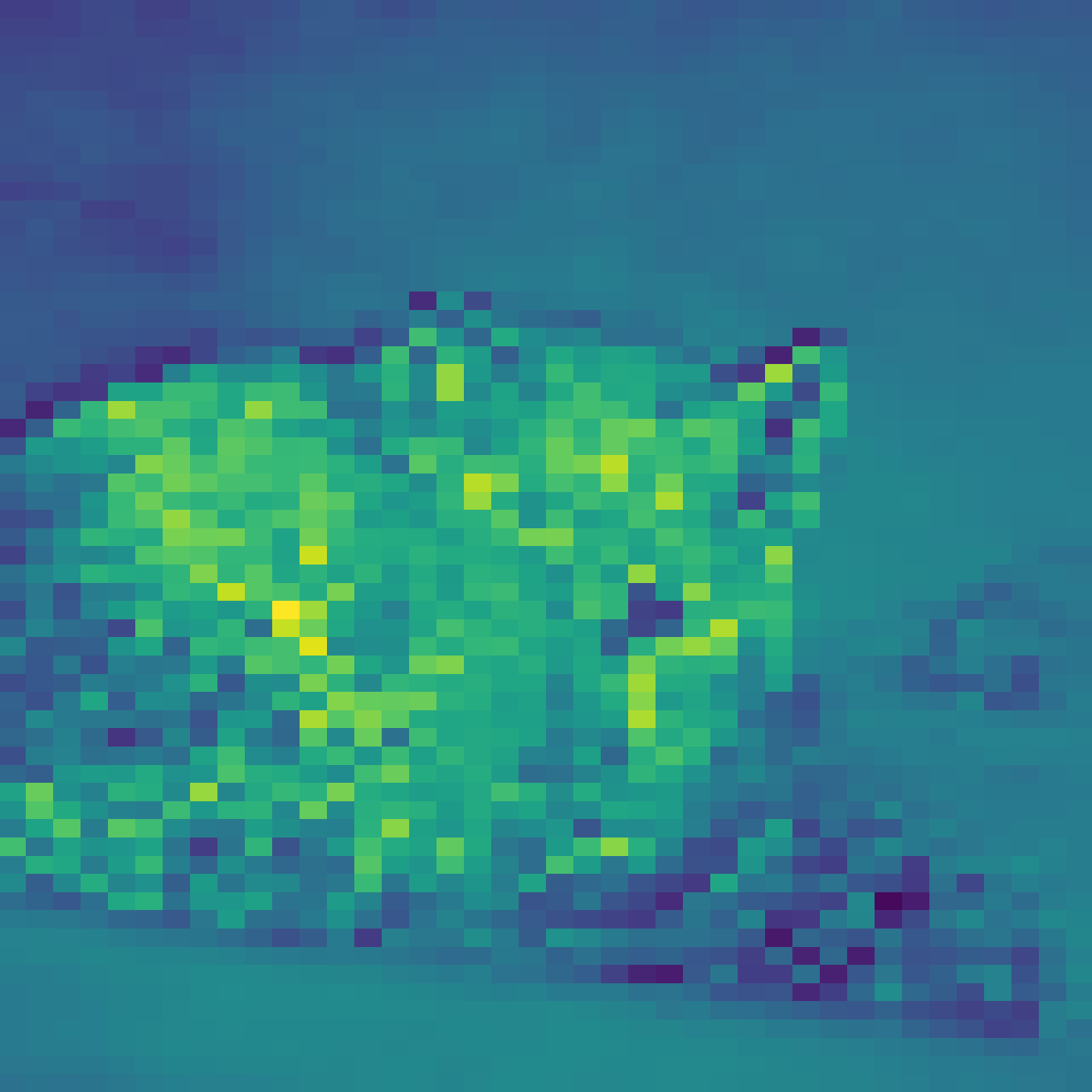}
   \end{subfigure}
   \begin{subfigure}[t]{0.11\linewidth}
       \centering
       \includegraphics[width=\linewidth]{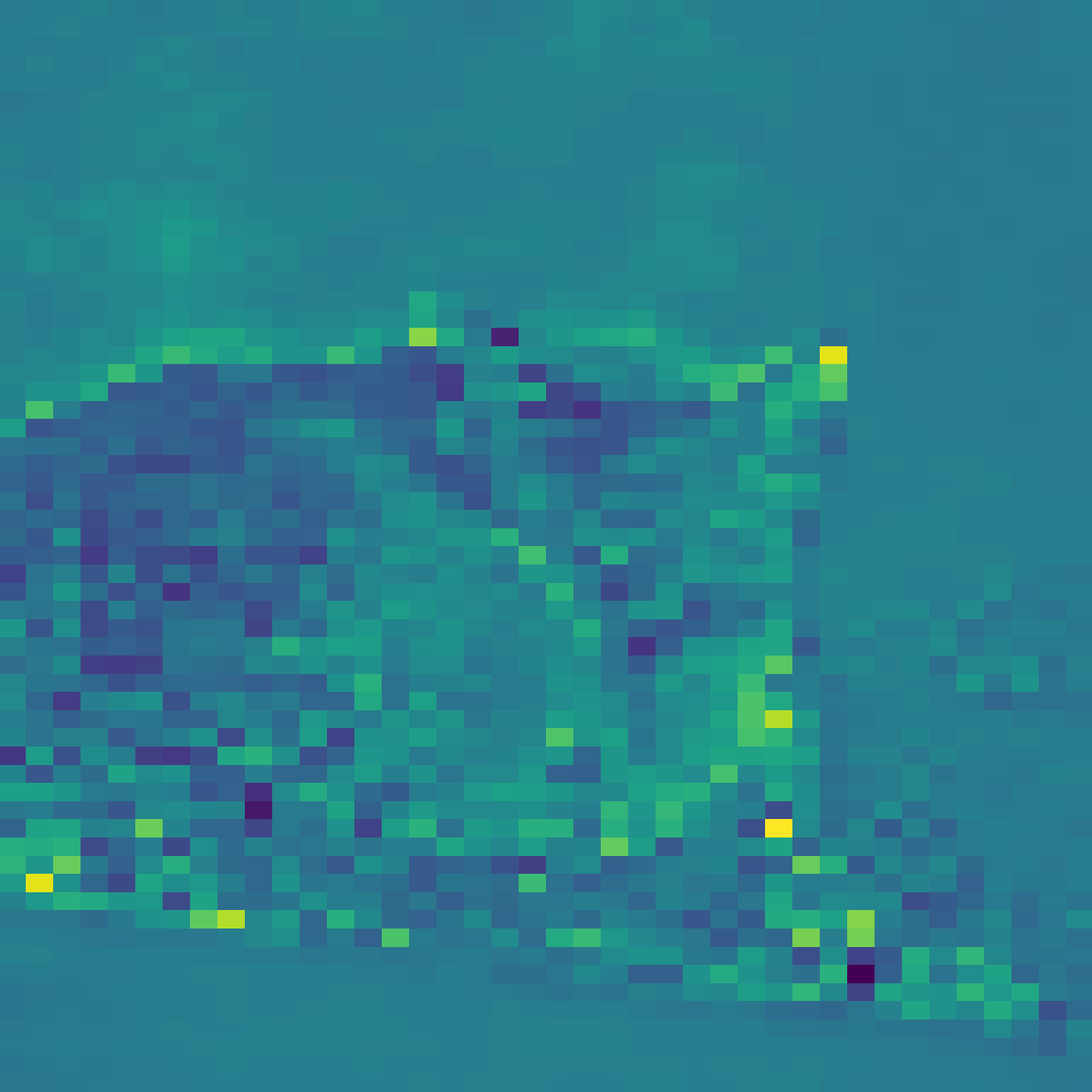}
   \end{subfigure}
   \begin{subfigure}[t]{0.11\linewidth}
       \centering
       \includegraphics[width=\linewidth]{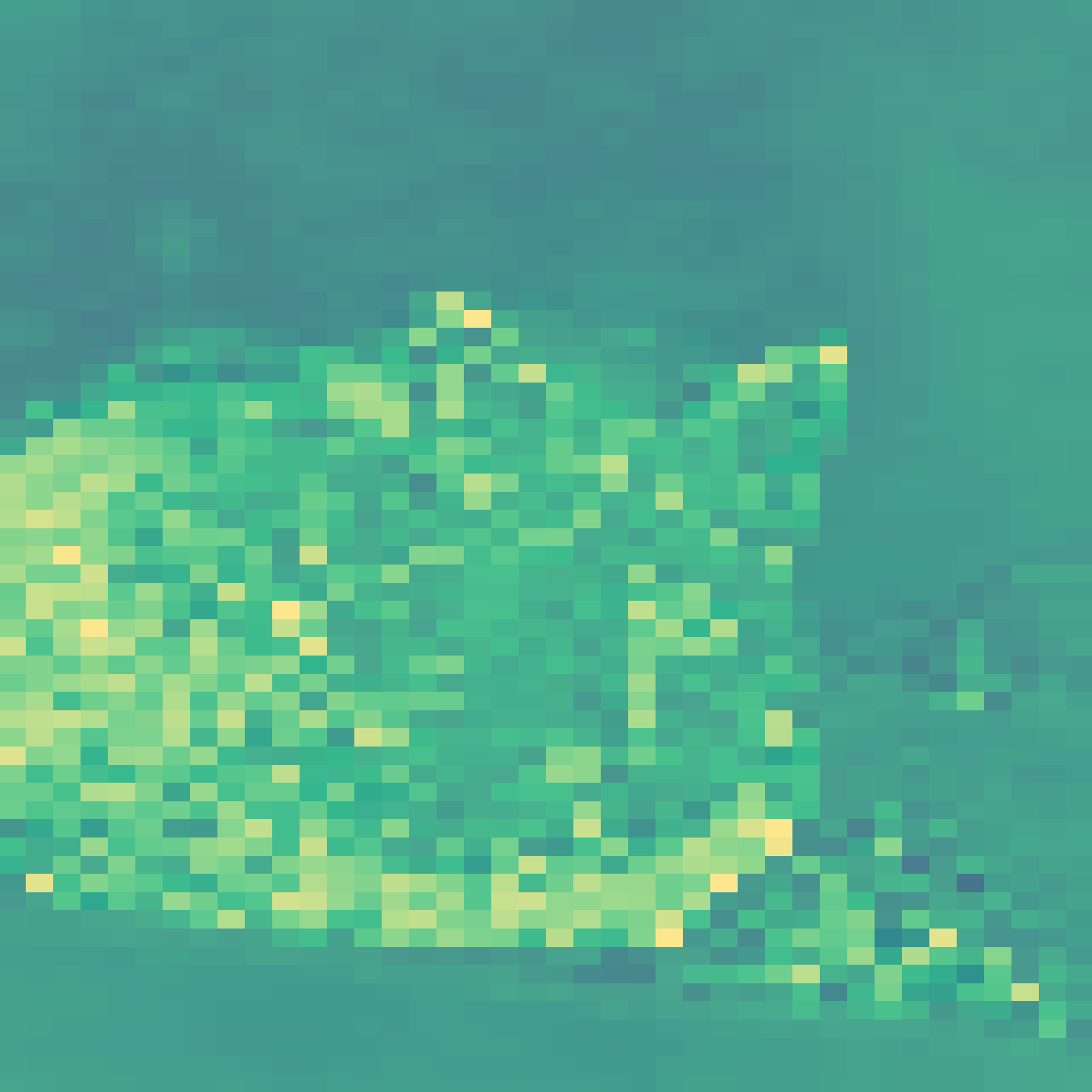}
   \end{subfigure}
    \begin{subfigure}[t]{0.11\linewidth}
       \centering
       \includegraphics[width=\linewidth]{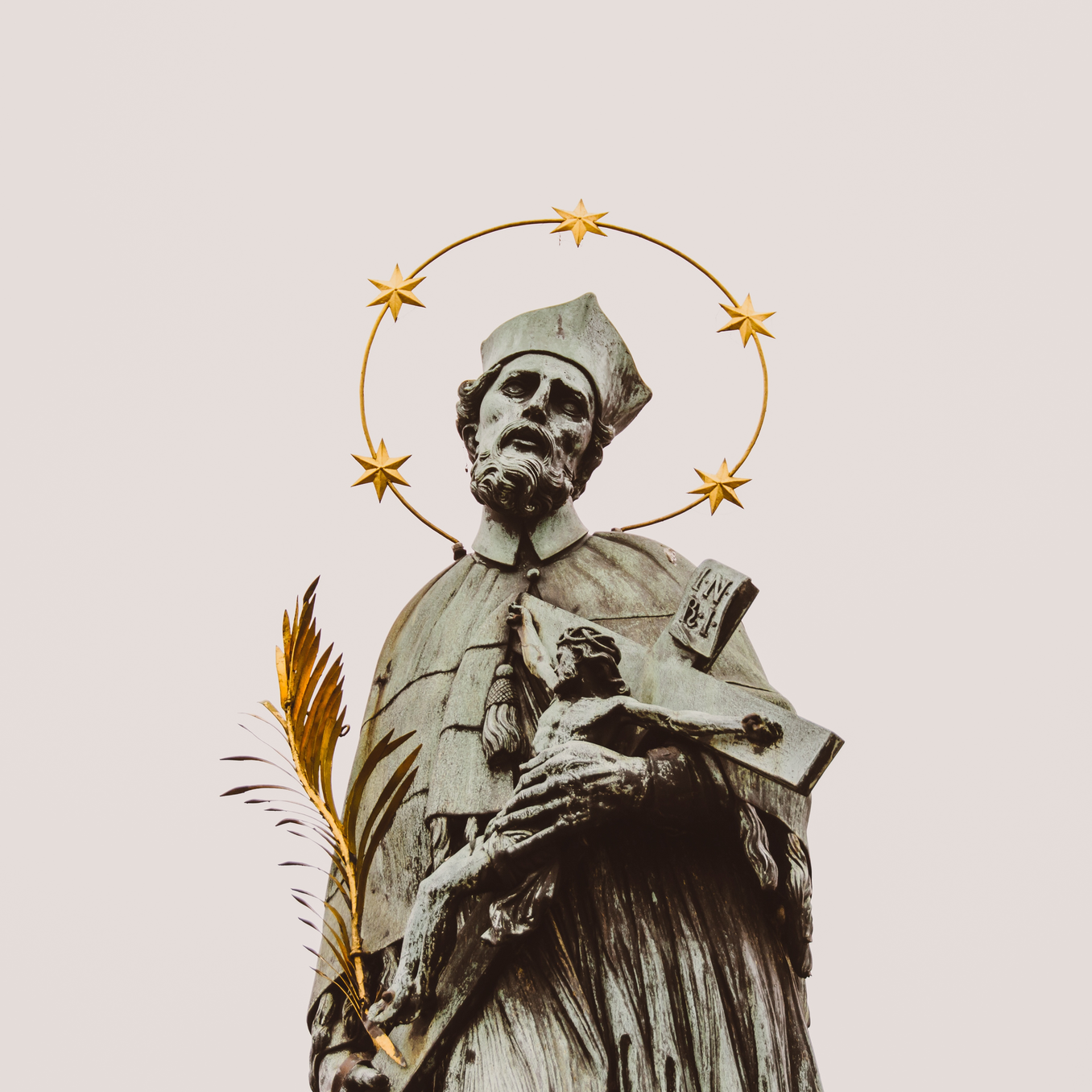}
   \end{subfigure}
   \begin{subfigure}[t]{0.11\linewidth}
       \centering
       \includegraphics[width=\linewidth]{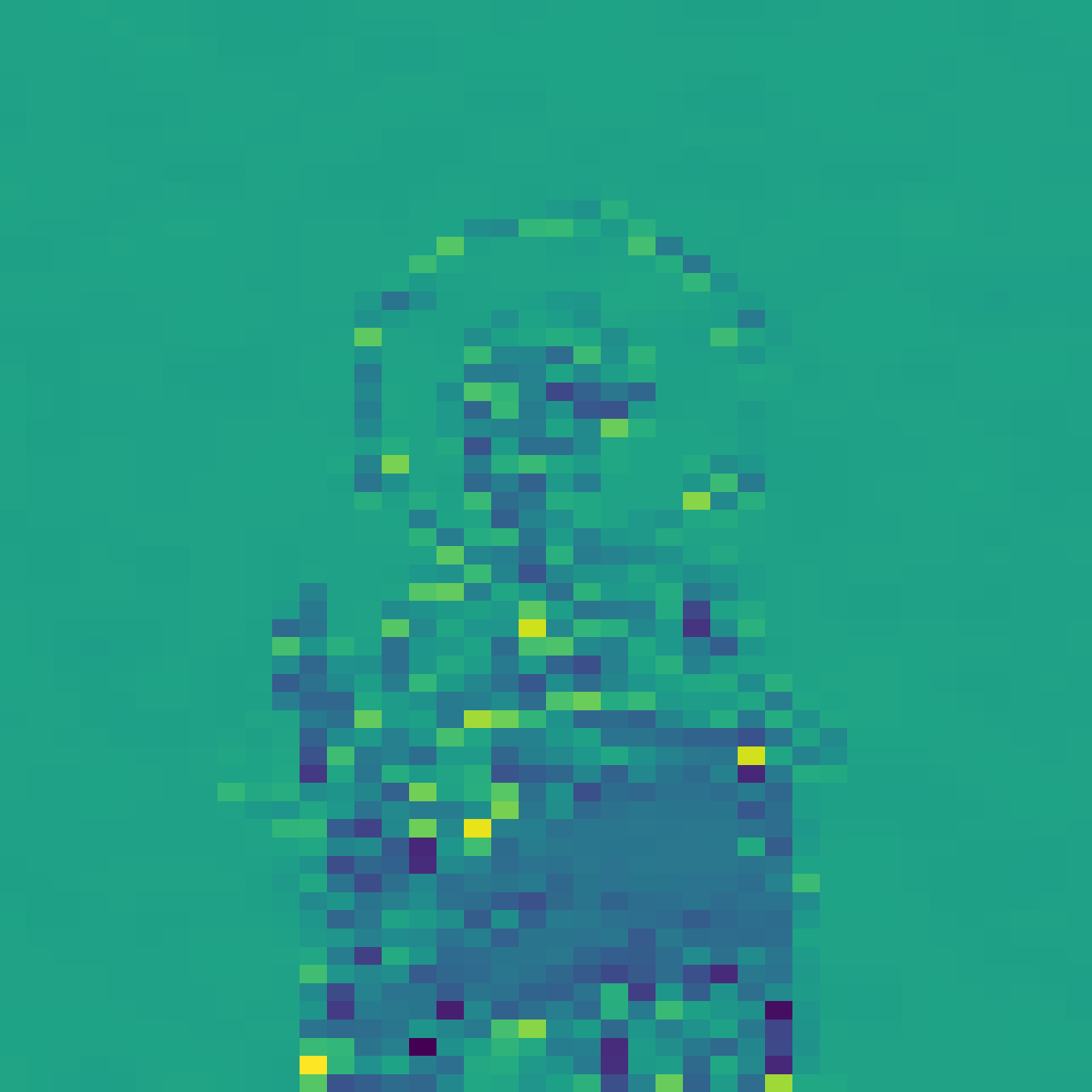}
   \end{subfigure}
   \begin{subfigure}[t]{0.11\linewidth}
       \centering
       \includegraphics[width=\linewidth]{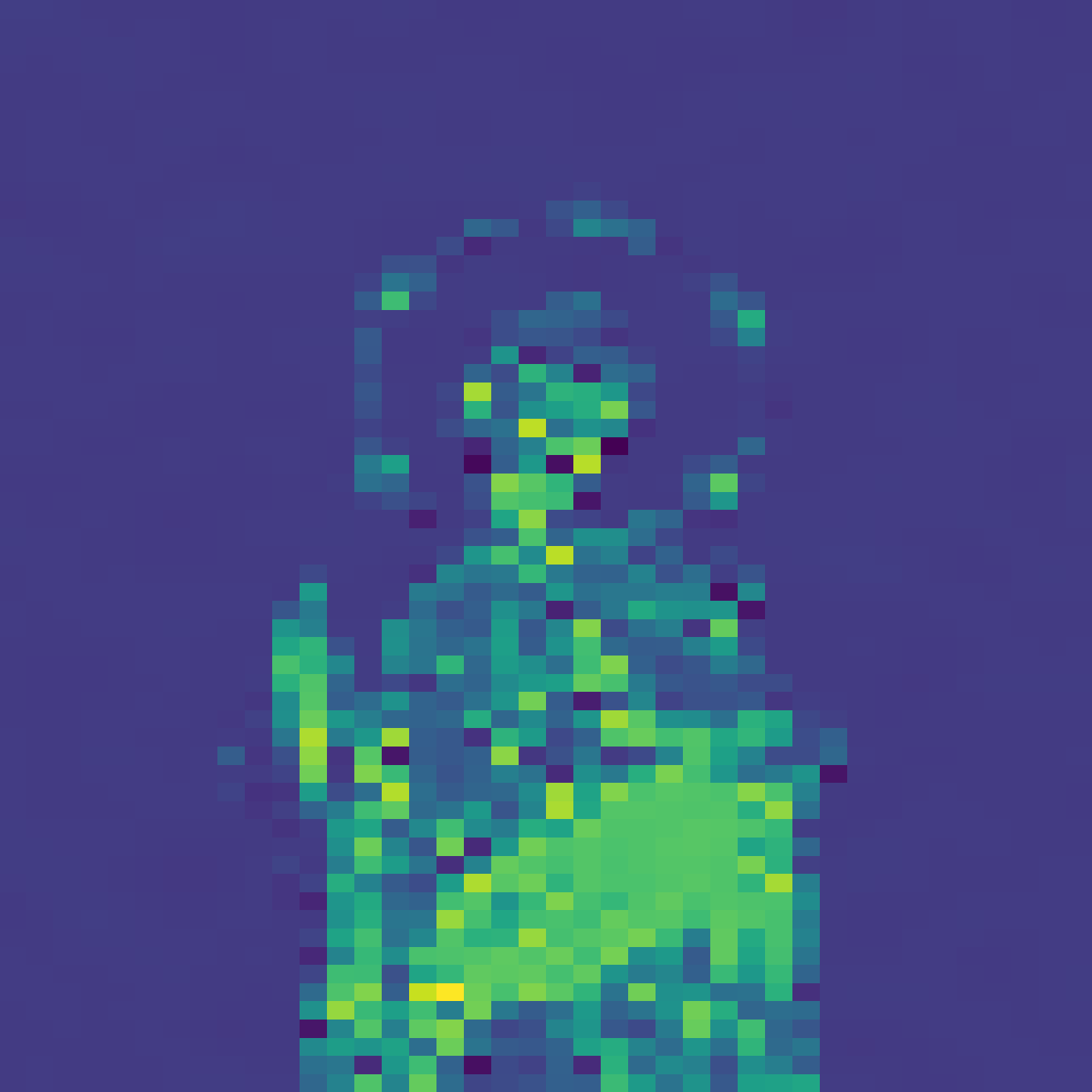}
   \end{subfigure}
   \begin{subfigure}[t]{0.11\linewidth}
       \centering
       \includegraphics[width=\linewidth]{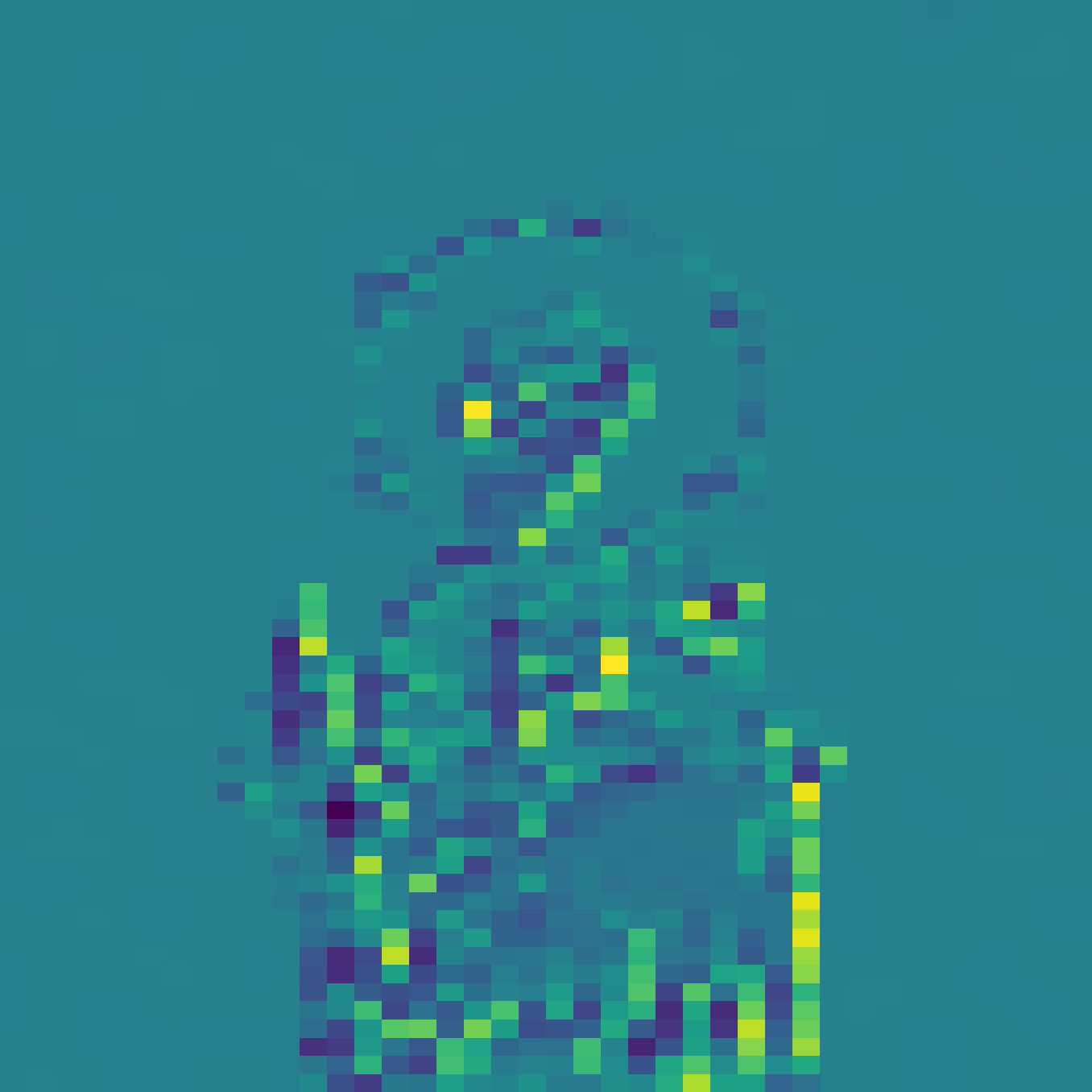}
   \end{subfigure}
   \begin{subfigure}[t]{0.11\linewidth}
       \centering
       \includegraphics[width=\linewidth]{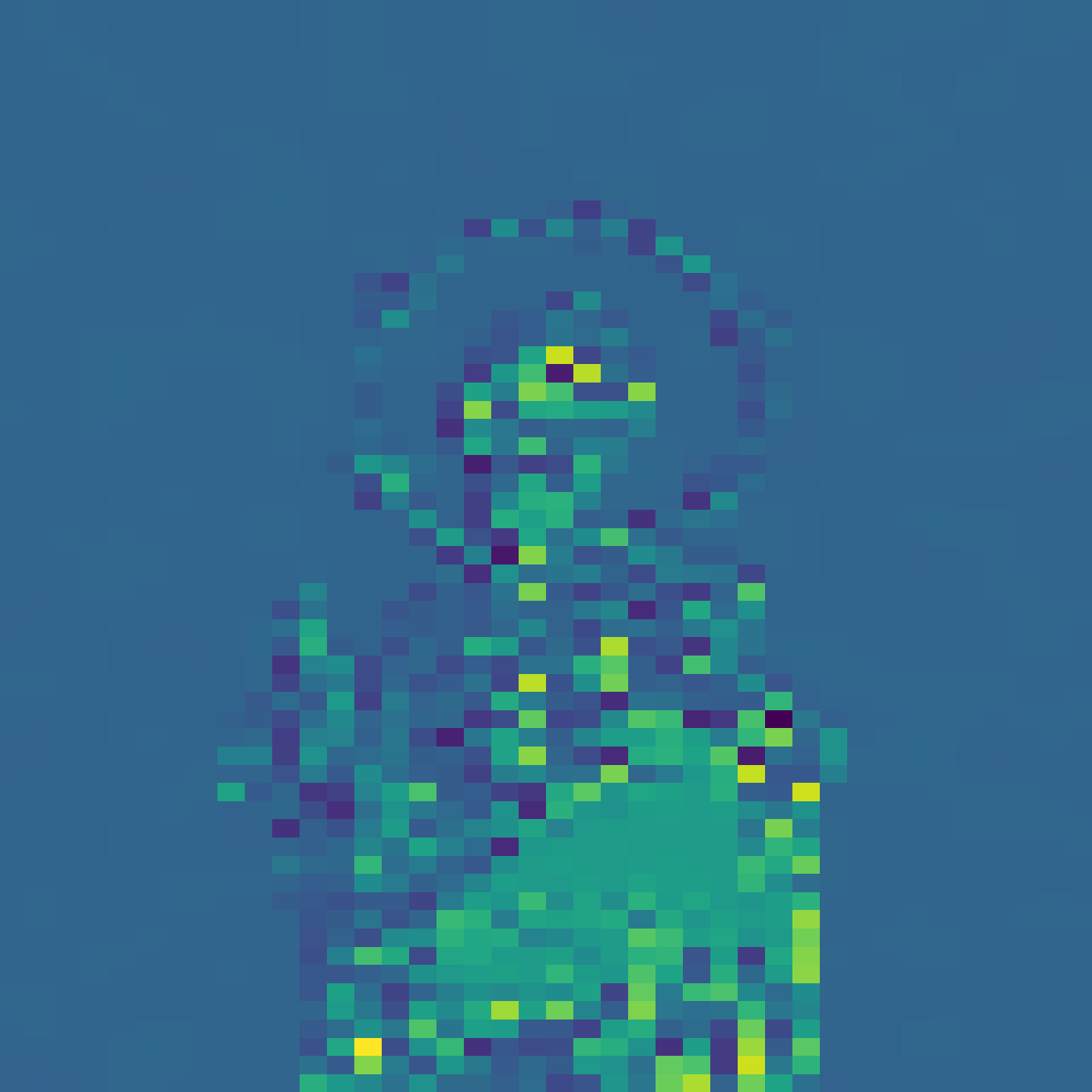}
   \end{subfigure}
   \begin{subfigure}[t]{0.11\linewidth}
       \centering
       \includegraphics[width=\linewidth]{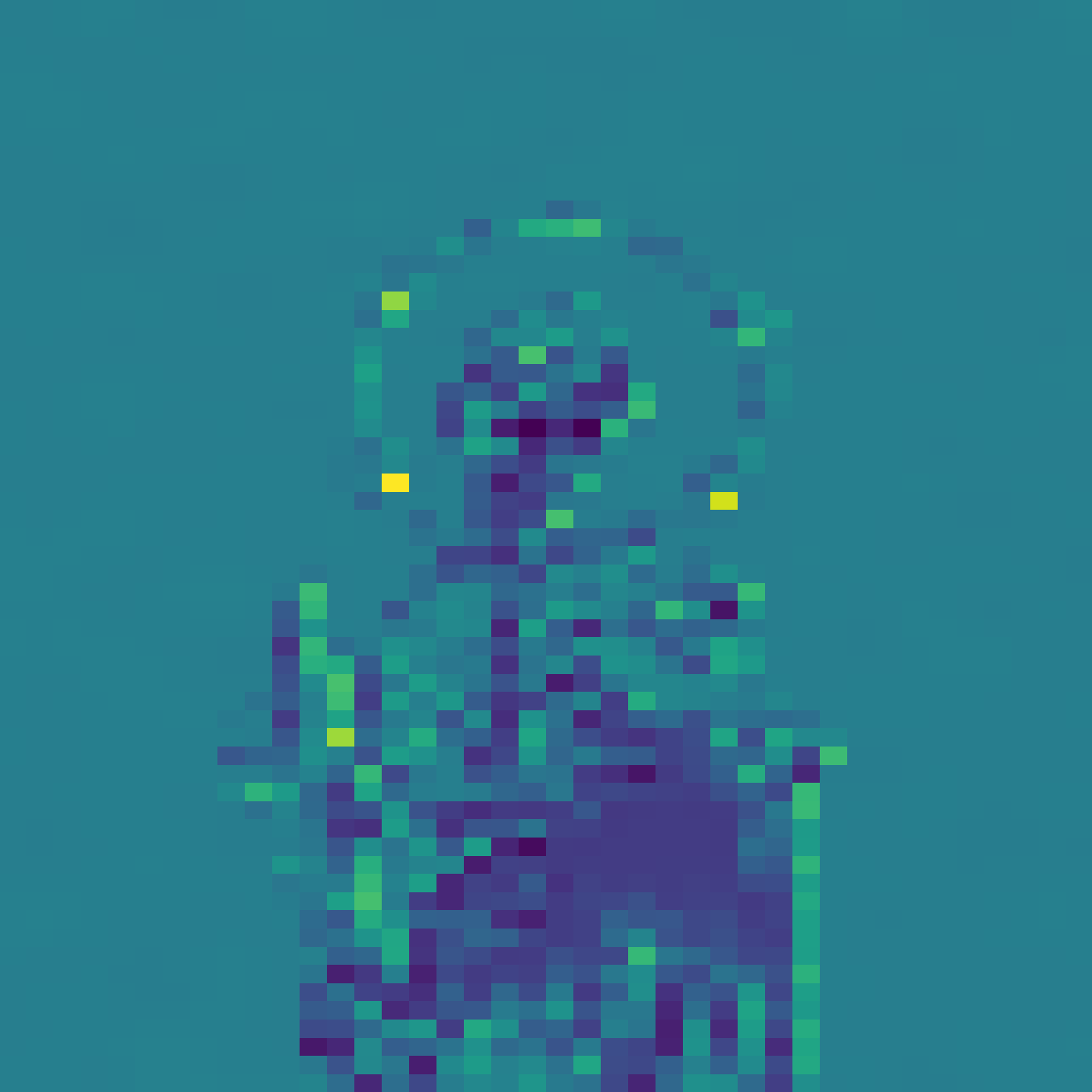}
   \end{subfigure}
   \begin{subfigure}[t]{0.11\linewidth}
       \centering
       \includegraphics[width=\linewidth]{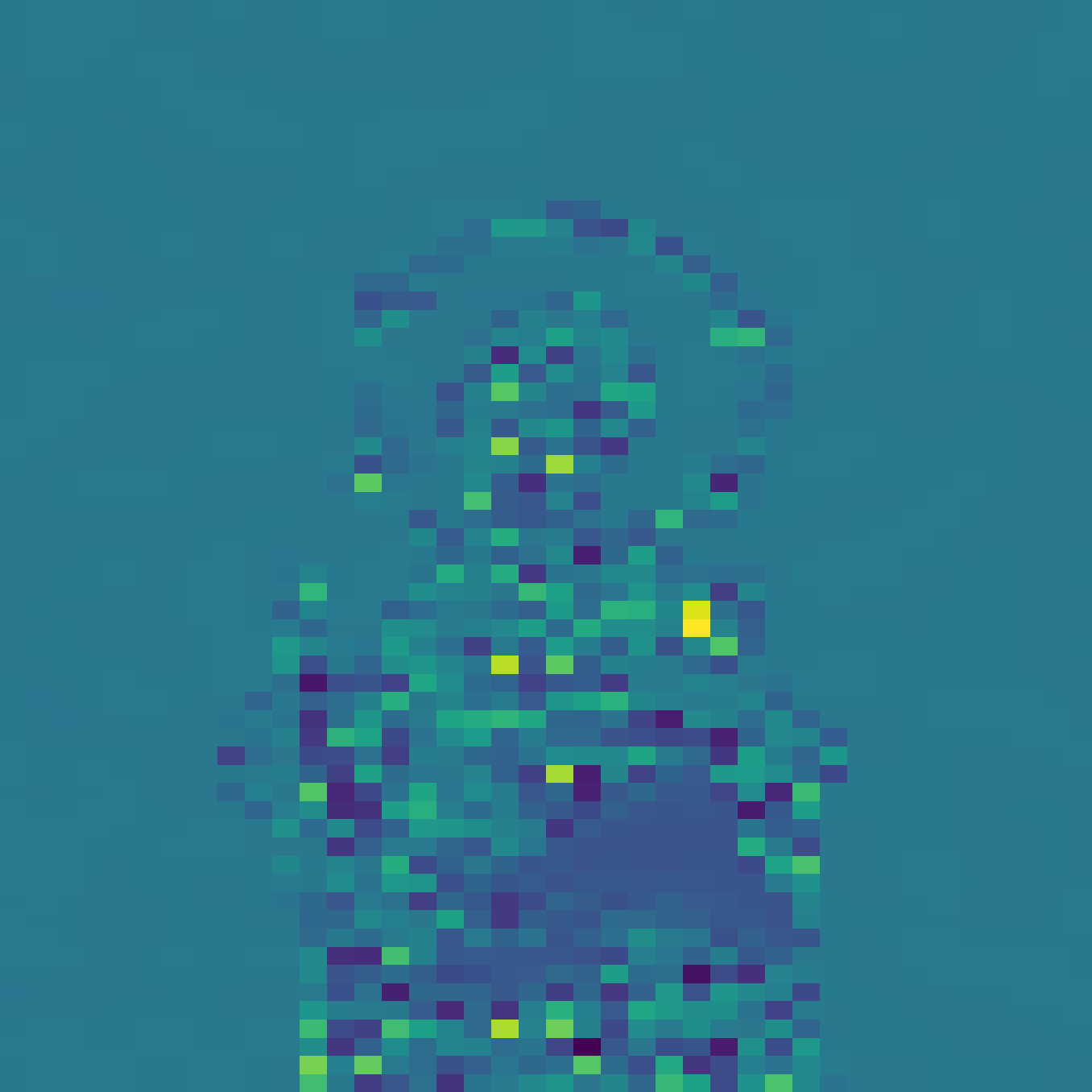}
   \end{subfigure}
   \begin{subfigure}[t]{0.11\linewidth}
       \centering
       \includegraphics[width=\linewidth]{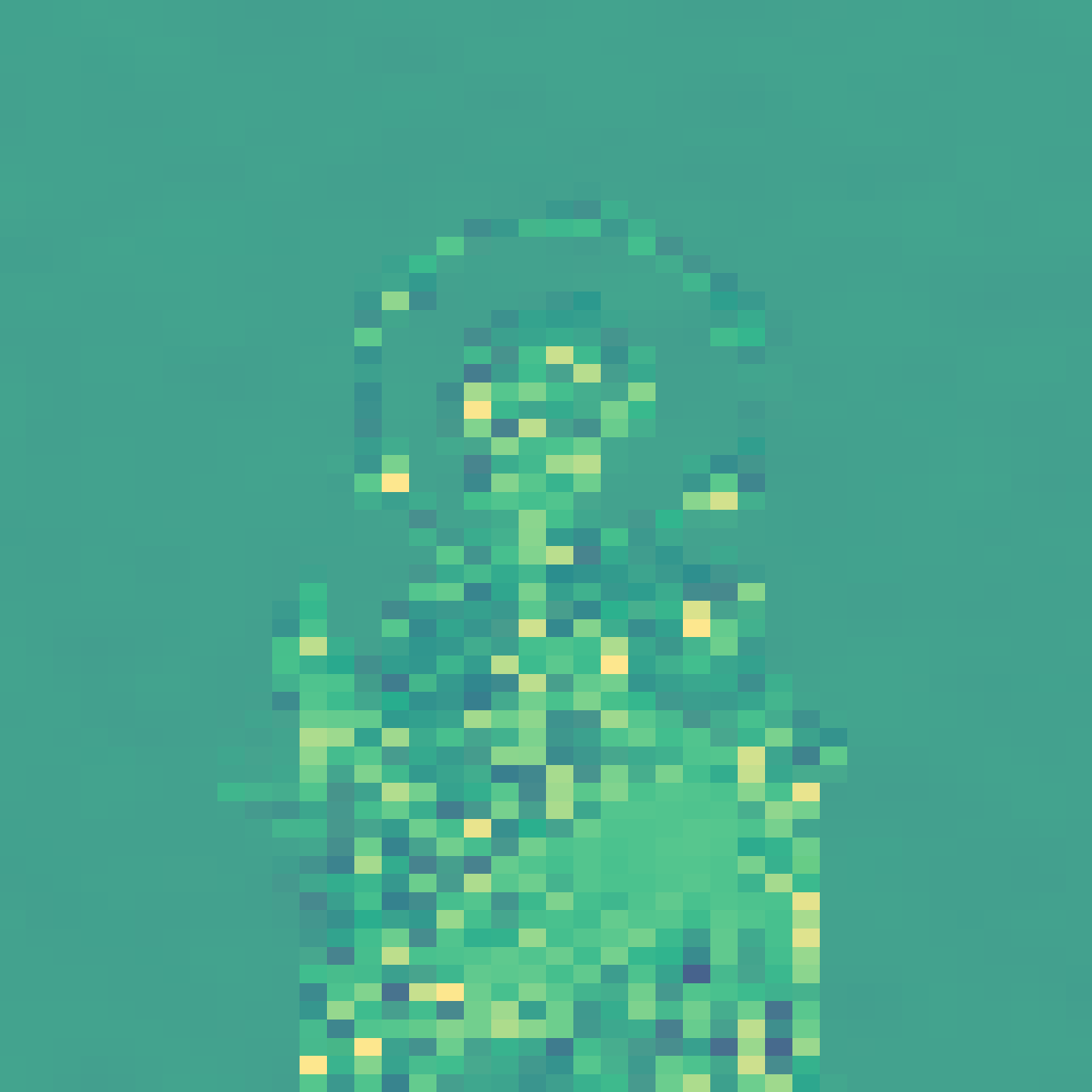}
   \end{subfigure}
    \begin{subfigure}[t]{0.11\linewidth}
       \centering
       \includegraphics[width=\linewidth]{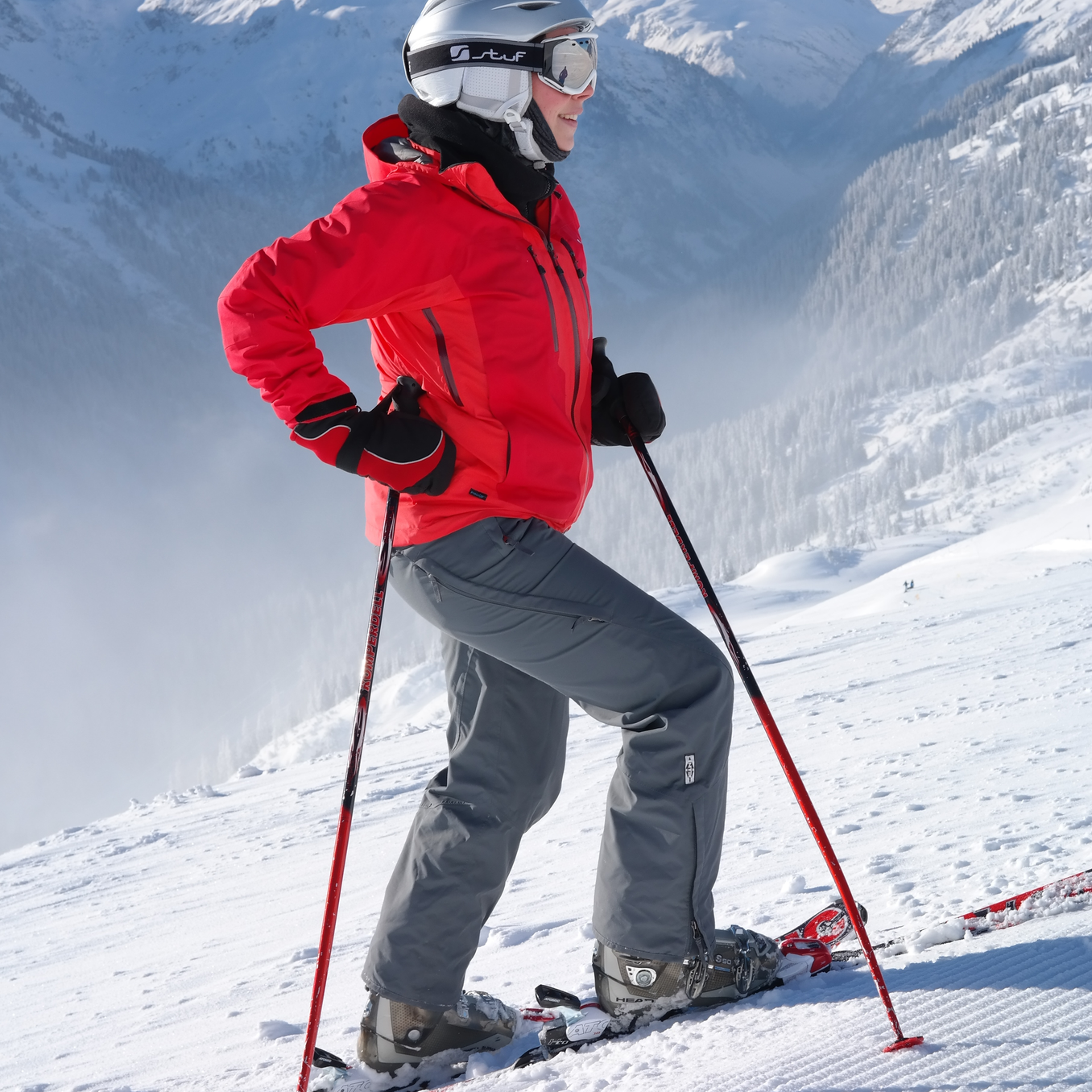}
   \end{subfigure}
   \begin{subfigure}[t]{0.11\linewidth}
       \centering
       \includegraphics[width=\linewidth]{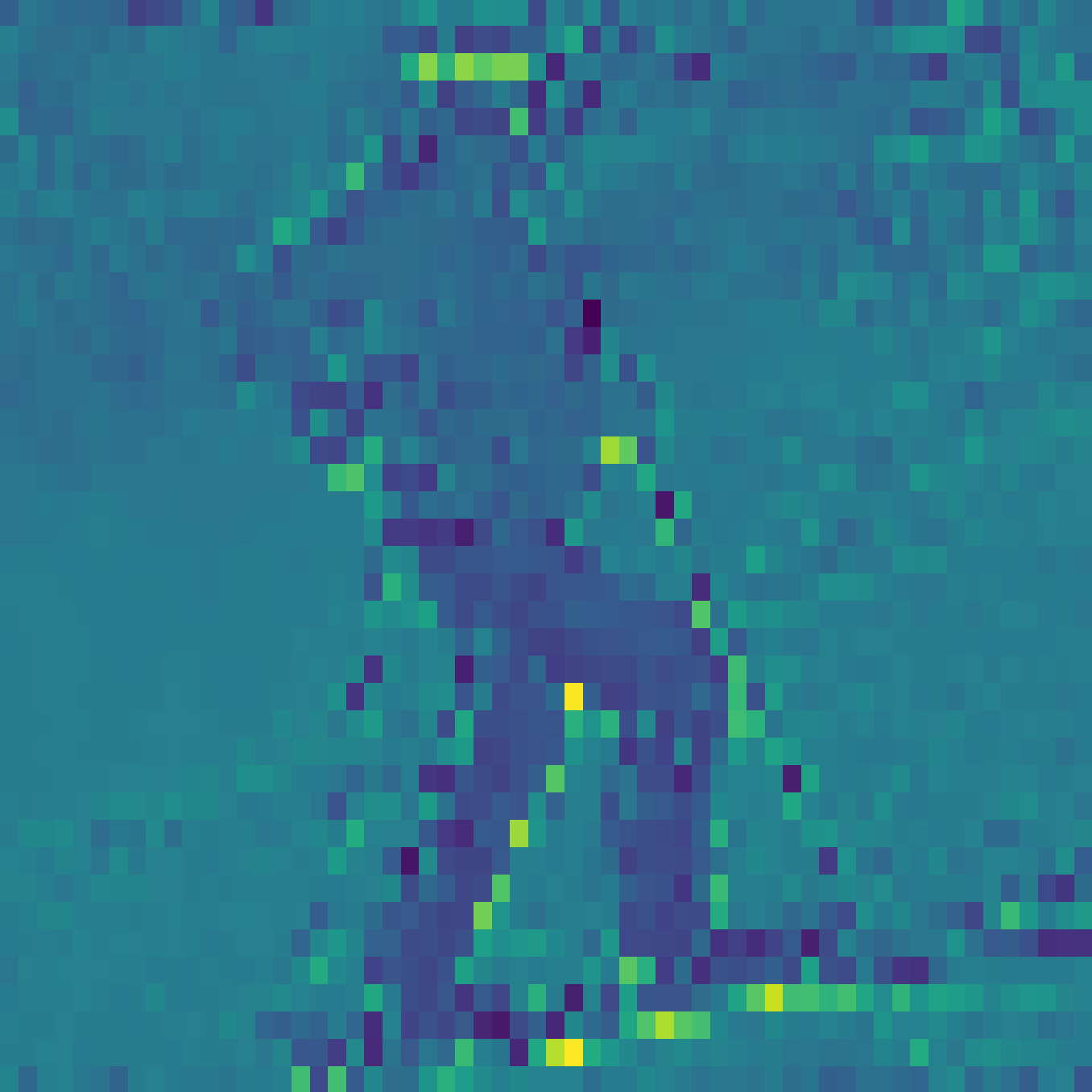}
   \end{subfigure}
   \begin{subfigure}[t]{0.11\linewidth}
       \centering
       \includegraphics[width=\linewidth]{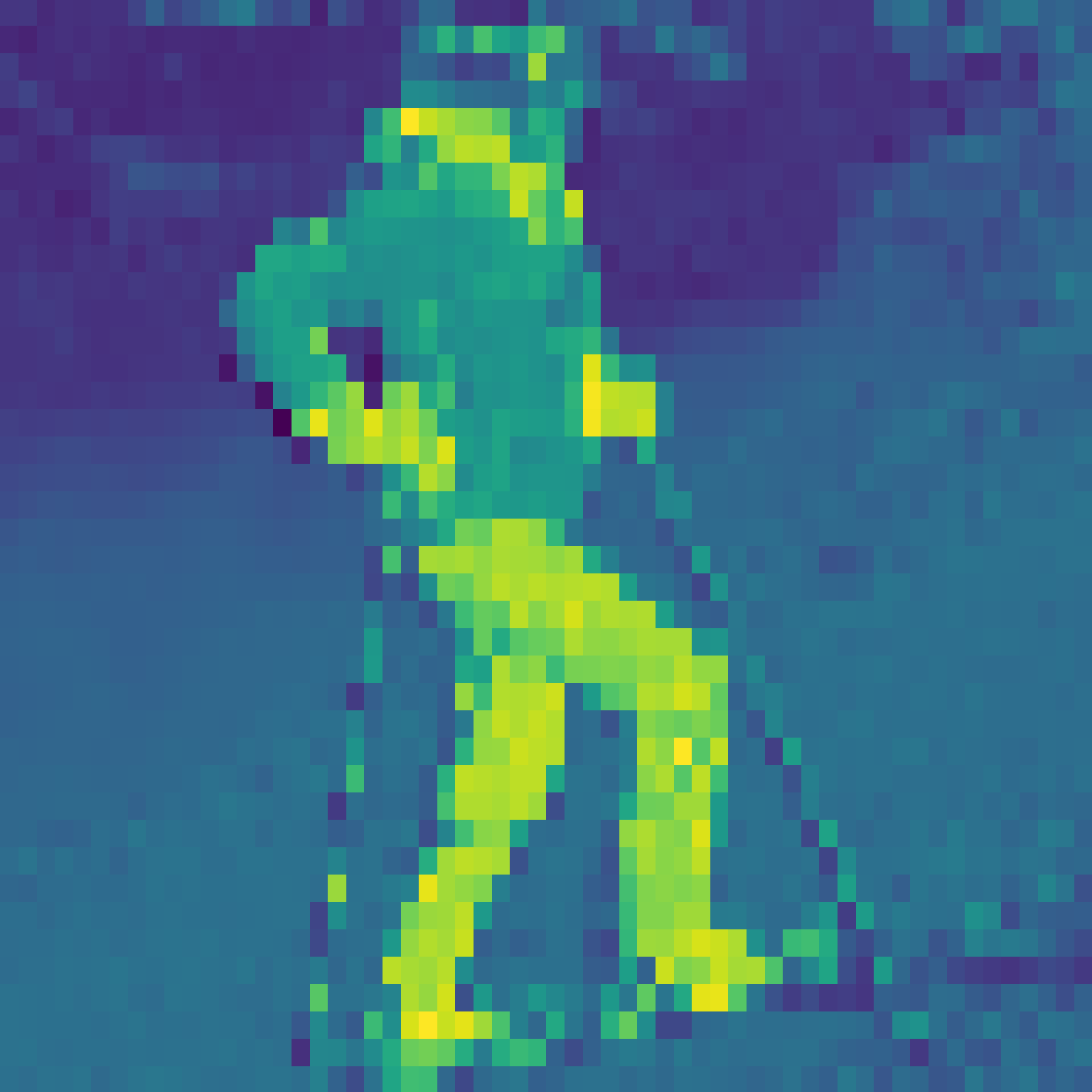}
   \end{subfigure}
   \begin{subfigure}[t]{0.11\linewidth}
       \centering
       \includegraphics[width=\linewidth]{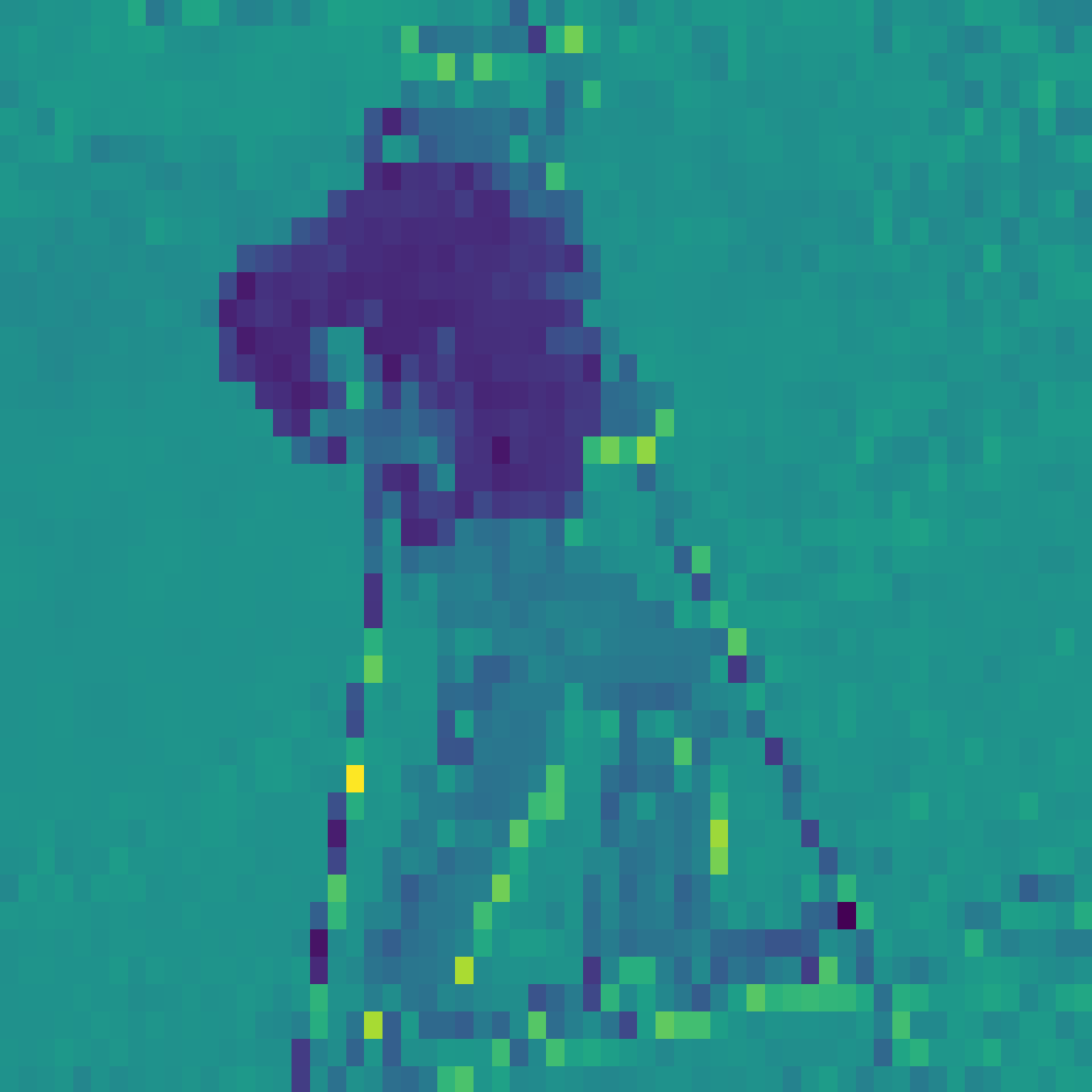}
   \end{subfigure}
   \begin{subfigure}[t]{0.11\linewidth}
       \centering
       \includegraphics[width=\linewidth]{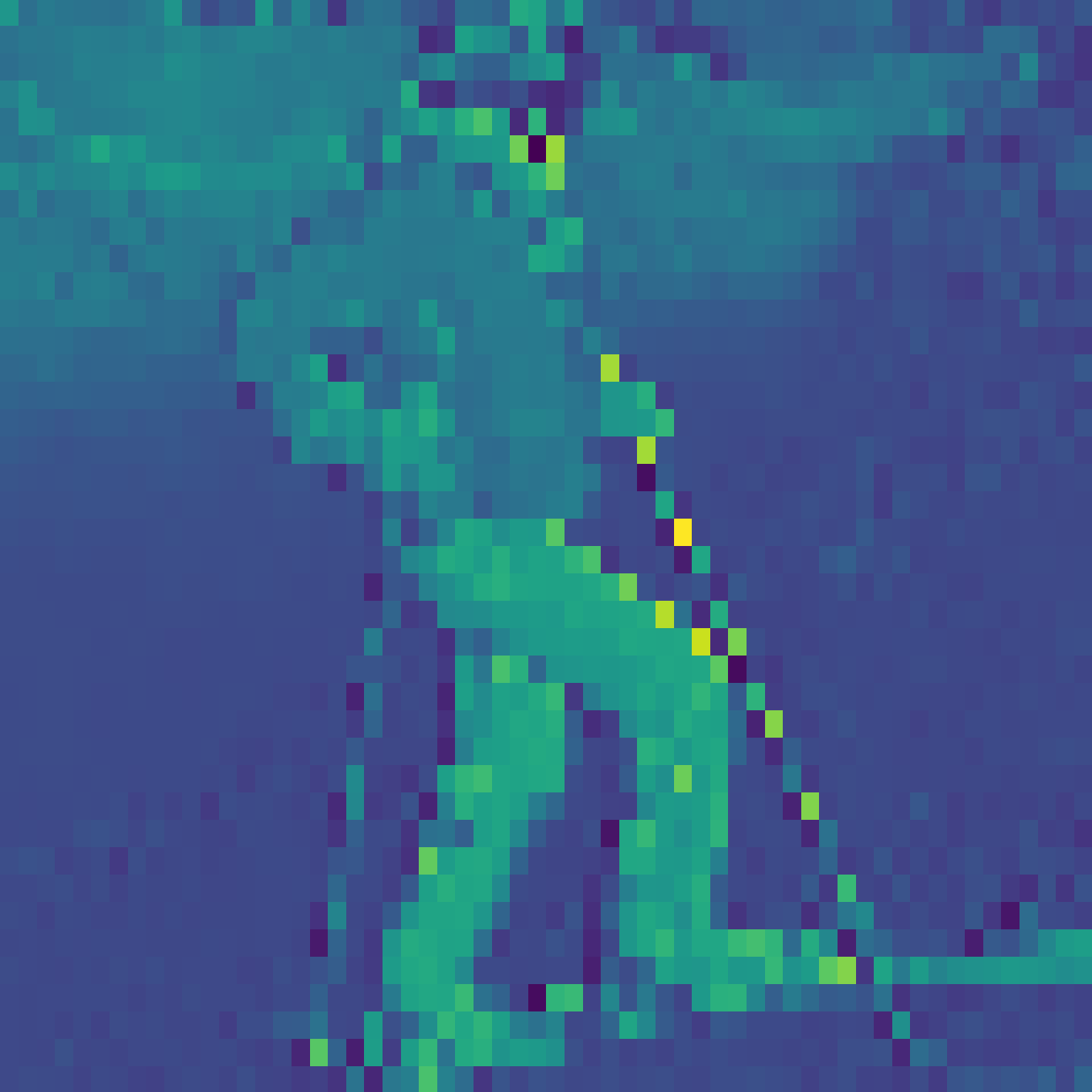}
   \end{subfigure}
   \begin{subfigure}[t]{0.11\linewidth}
       \centering
       \includegraphics[width=\linewidth]{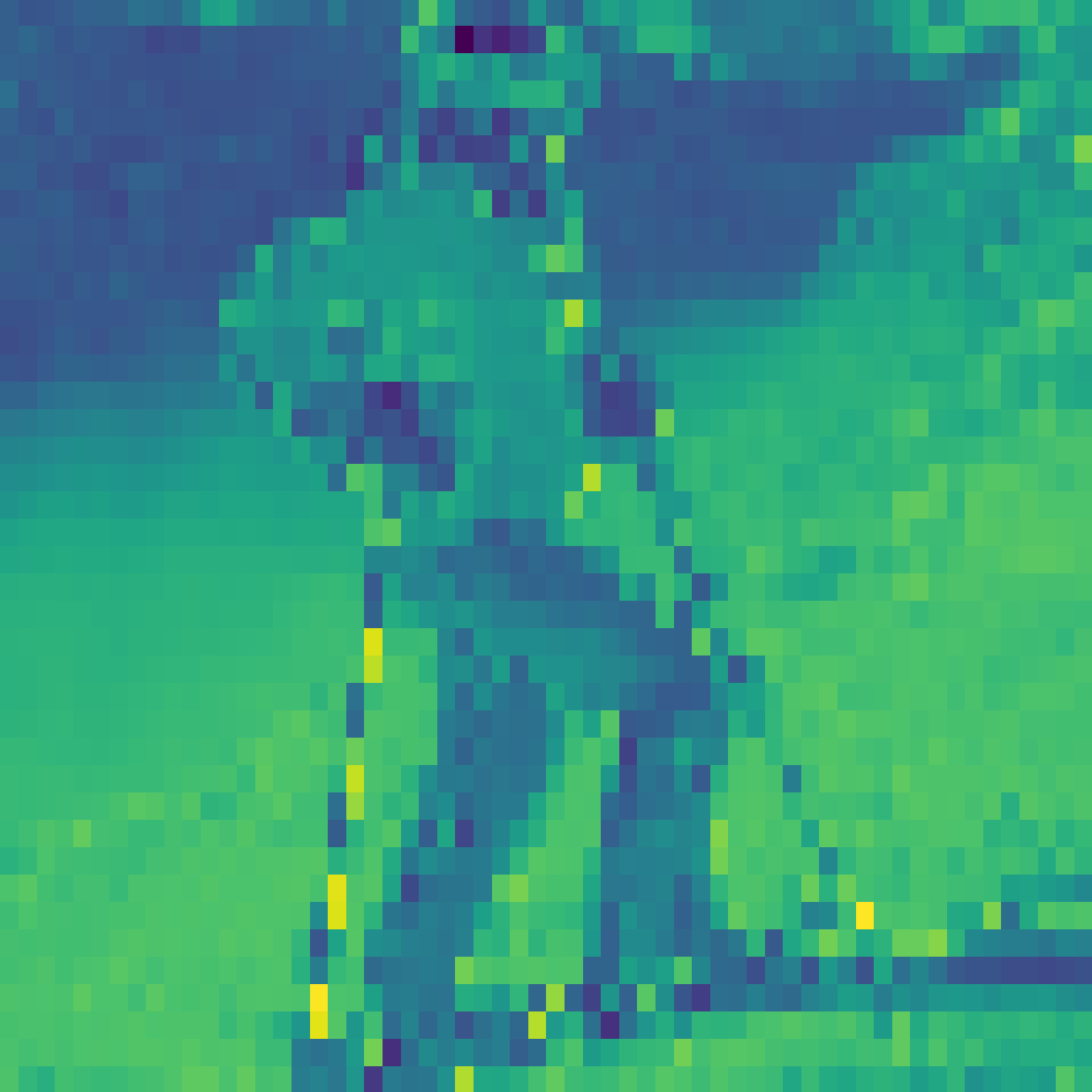}
   \end{subfigure}
   \begin{subfigure}[t]{0.11\linewidth}
       \centering
       \includegraphics[width=\linewidth]{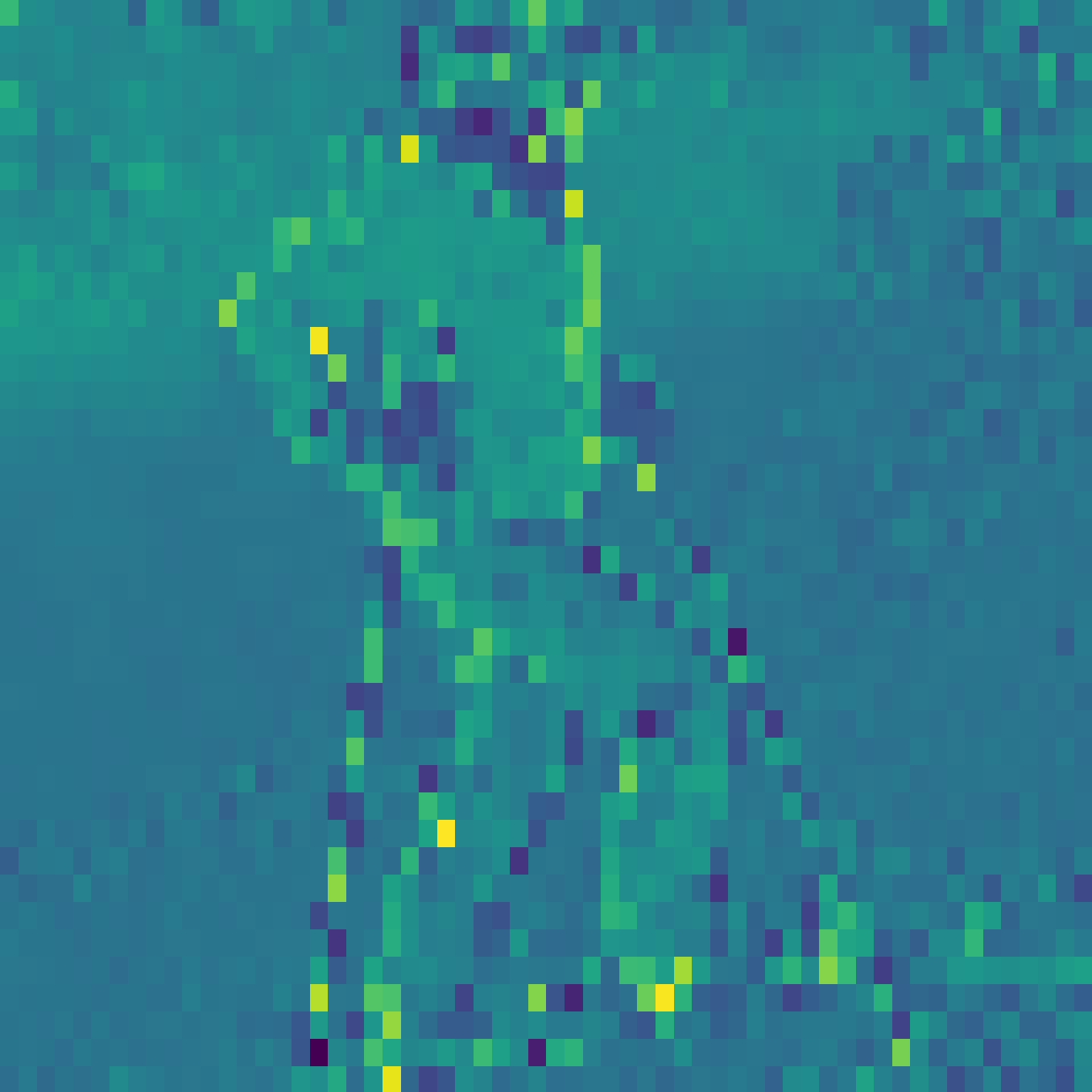}
   \end{subfigure}
   \begin{subfigure}[t]{0.11\linewidth}
       \centering
       \includegraphics[width=\linewidth]{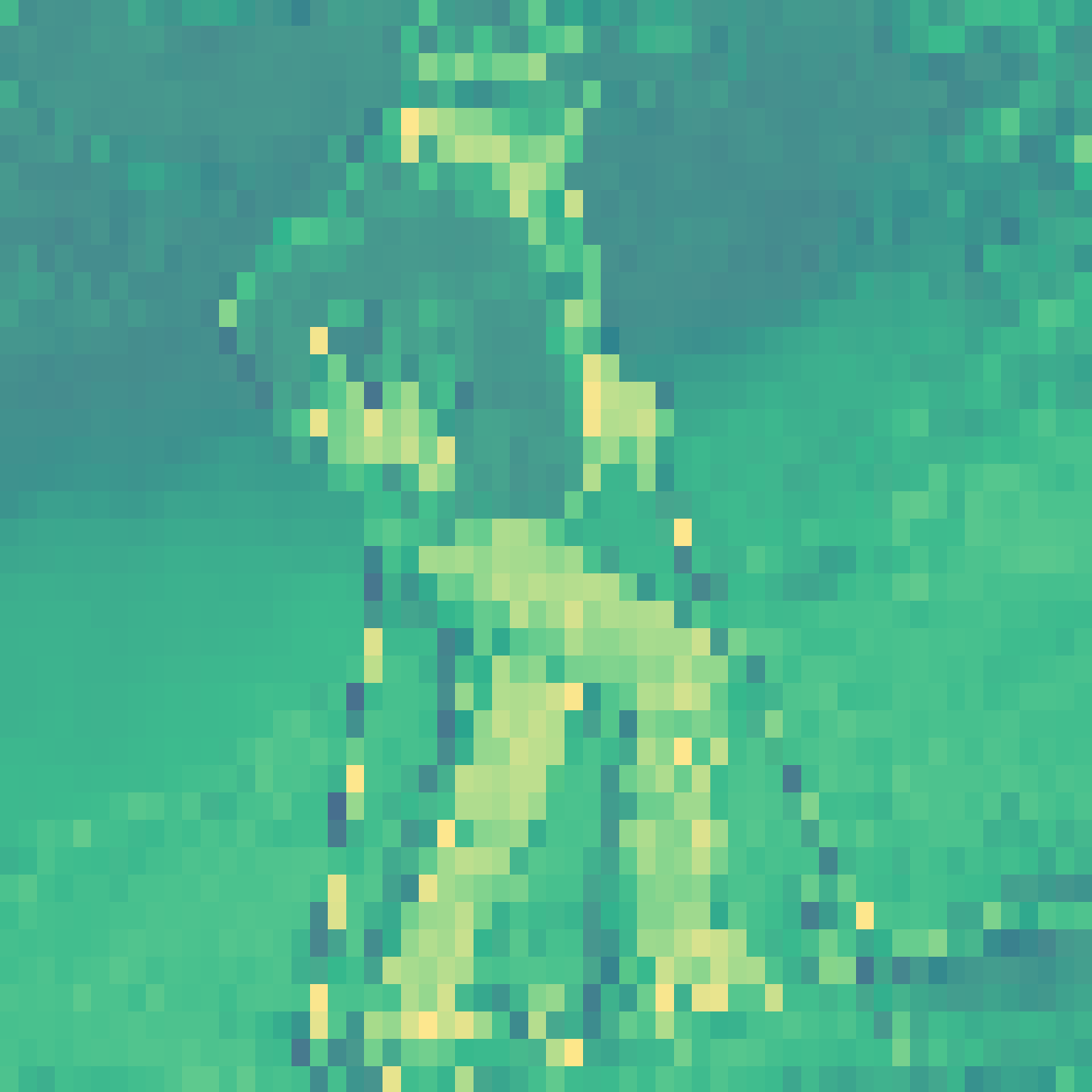}
   \end{subfigure}
   \begin{subfigure}[t]{0.11\linewidth}
       \centering
       \includegraphics[width=\linewidth]{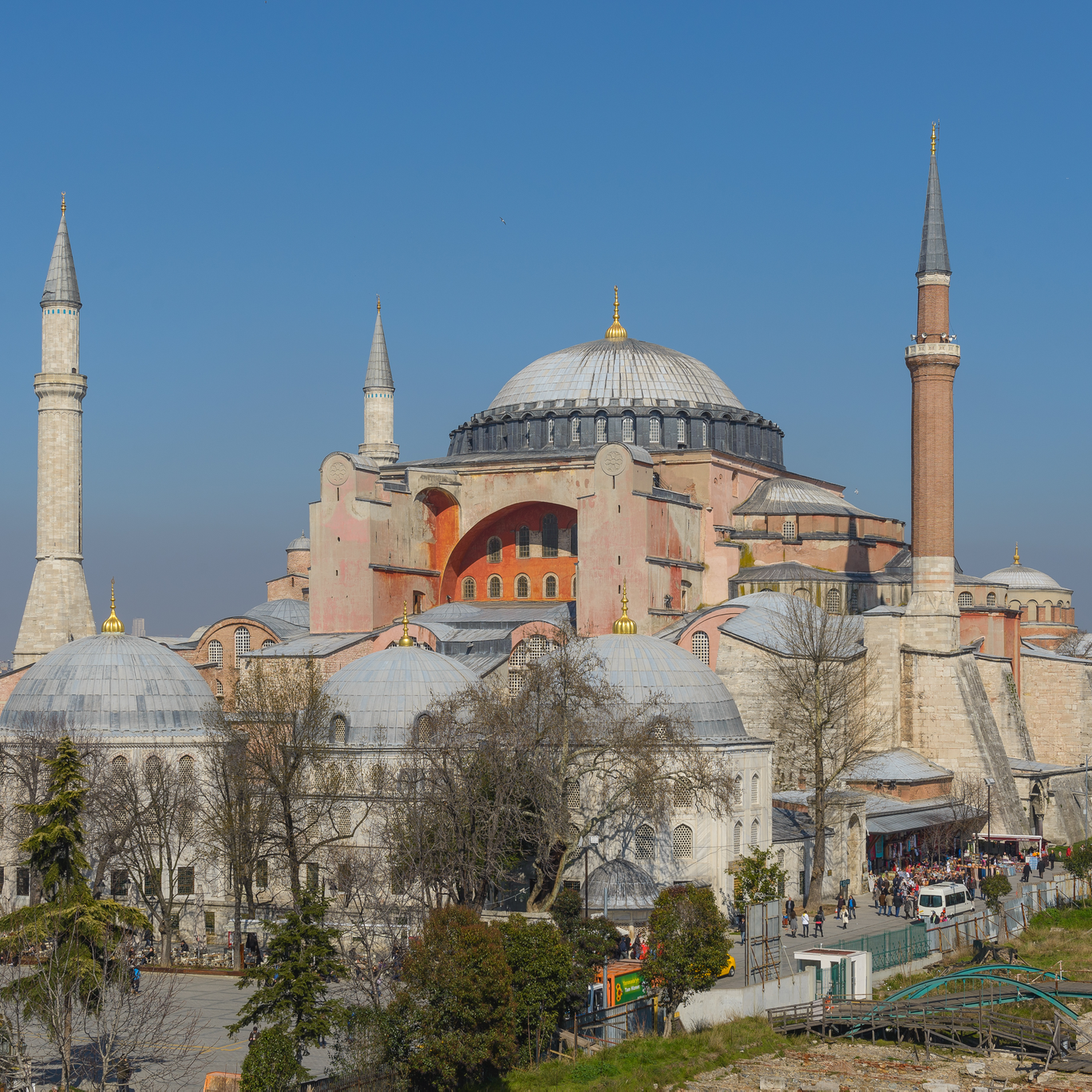}
       Input
   \end{subfigure}
\hfill 
   \begin{subfigure}[t]{0.11\linewidth}
       \centering
       \includegraphics[width=\linewidth]{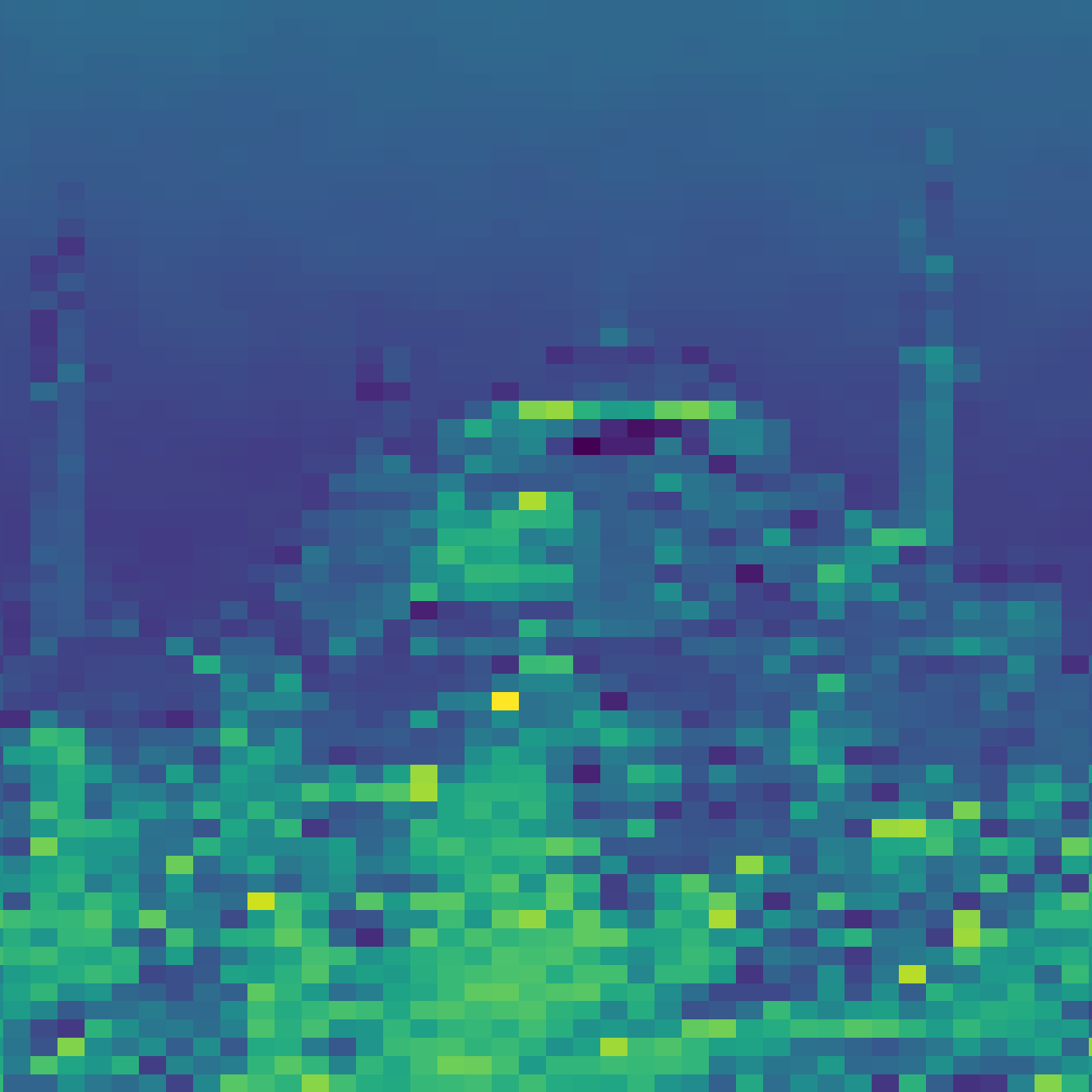}
       Att.-Head 0
   \end{subfigure}
\hfill 
   \begin{subfigure}[t]{0.11\linewidth}
       \centering
       \includegraphics[width=\linewidth]{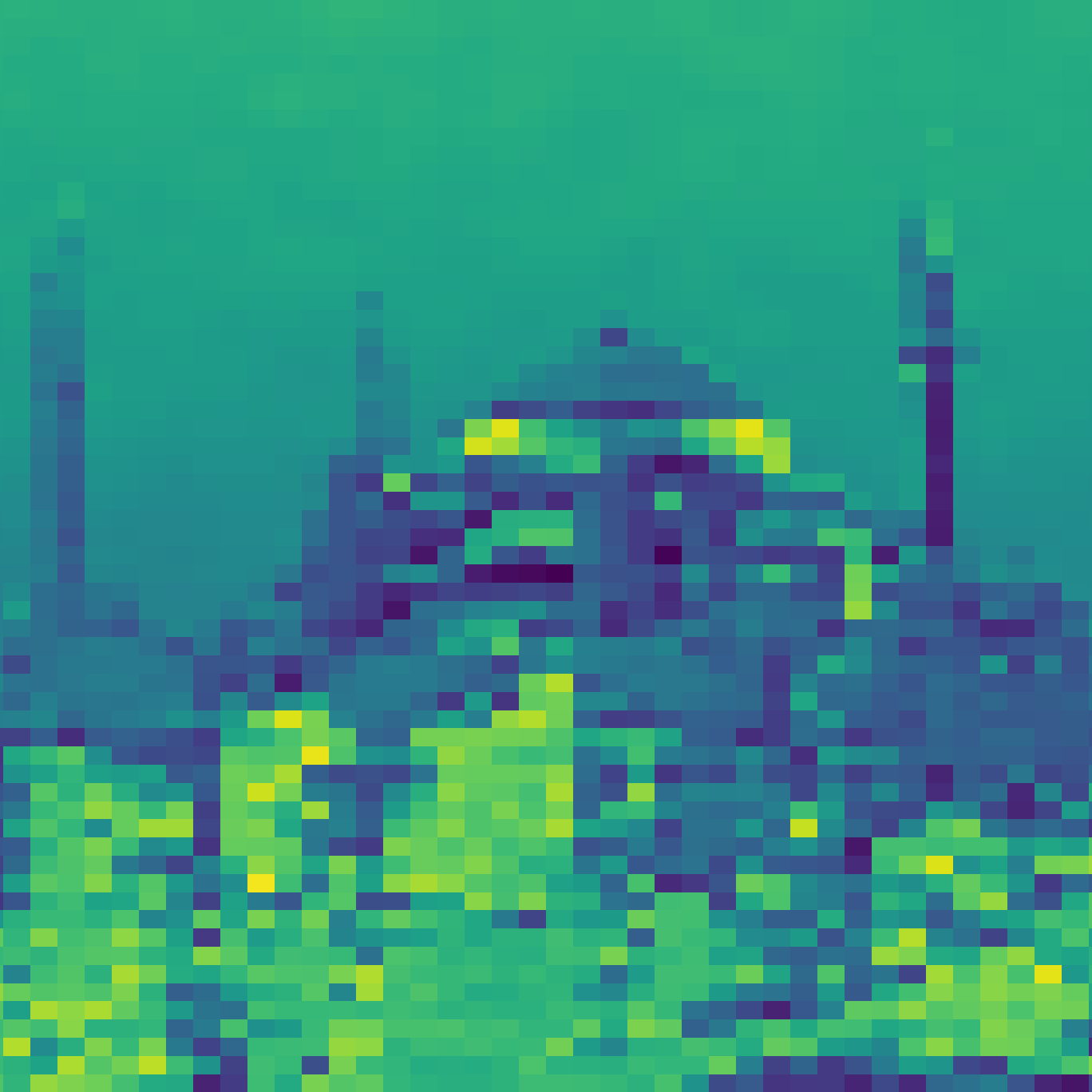}
       Att.-Head 1
   \end{subfigure}
\hfill 
   \begin{subfigure}[t]{0.11\linewidth}
       \centering
       \includegraphics[width=\linewidth]{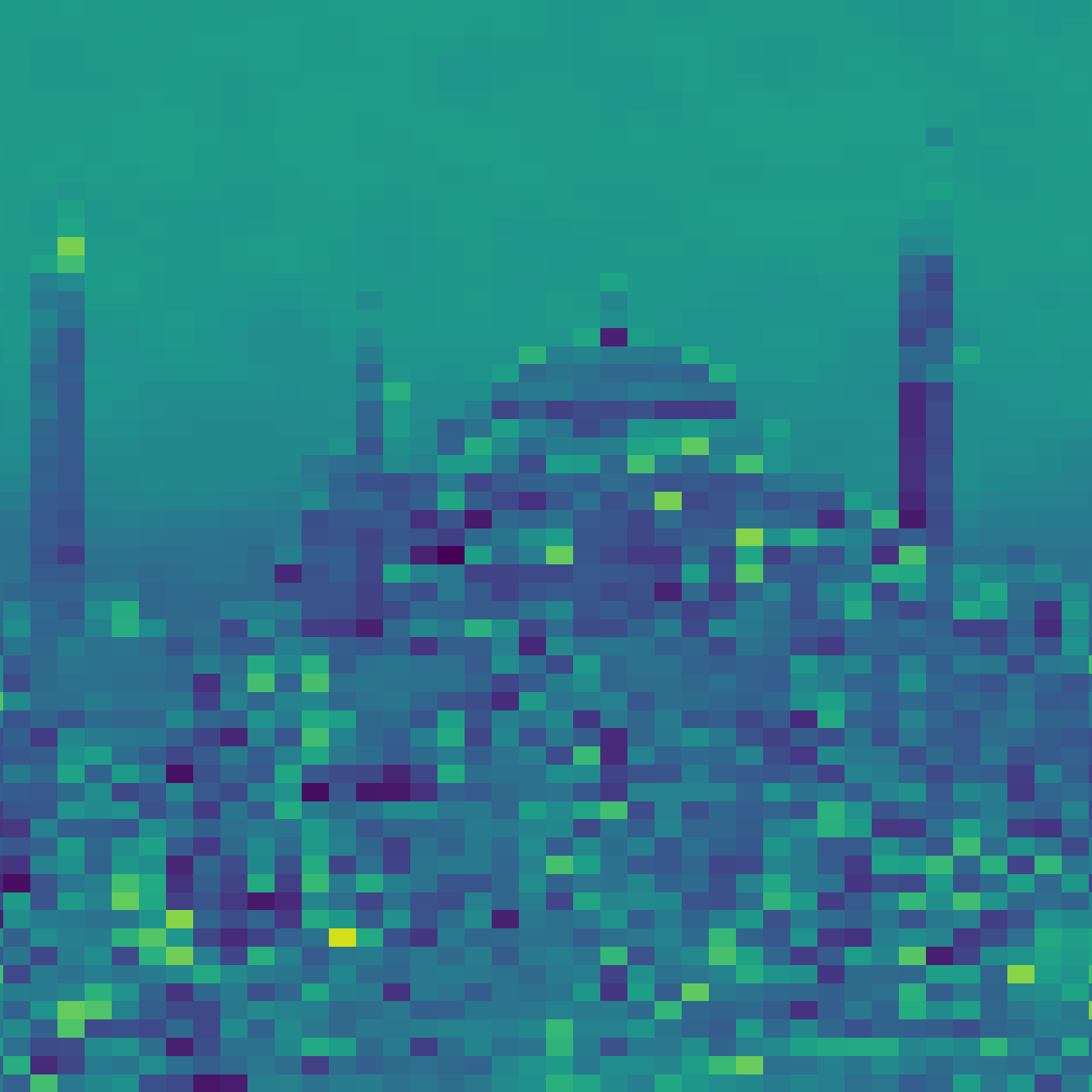}
       Att.-Head 2
   \end{subfigure}
   \hfill 
   \begin{subfigure}[t]{0.11\linewidth}
       \centering
       \includegraphics[width=\linewidth]{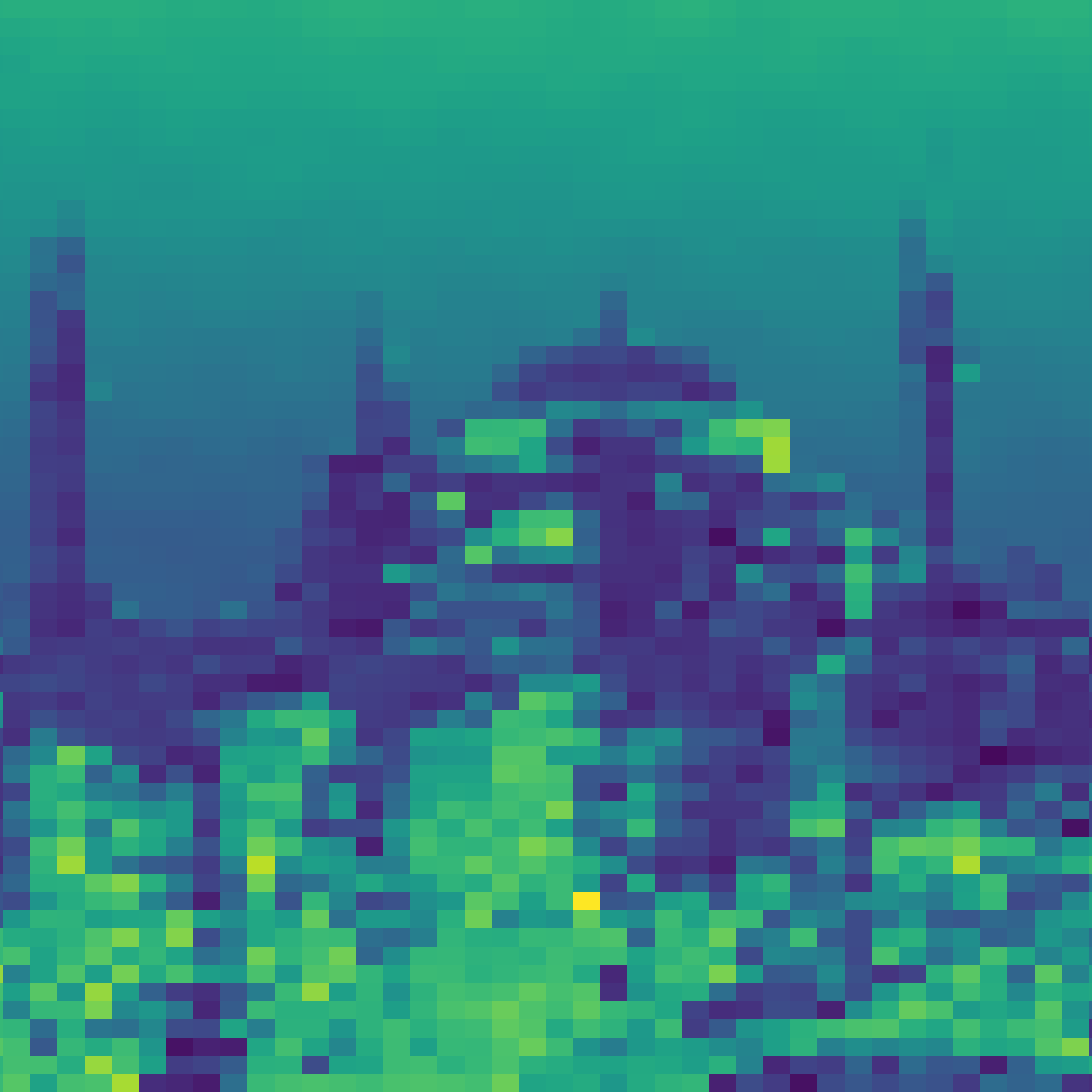}
       Att.-Head 3
   \end{subfigure}
   \hfill 
   \begin{subfigure}[t]{0.11\linewidth}
       \centering
       \includegraphics[width=\linewidth]{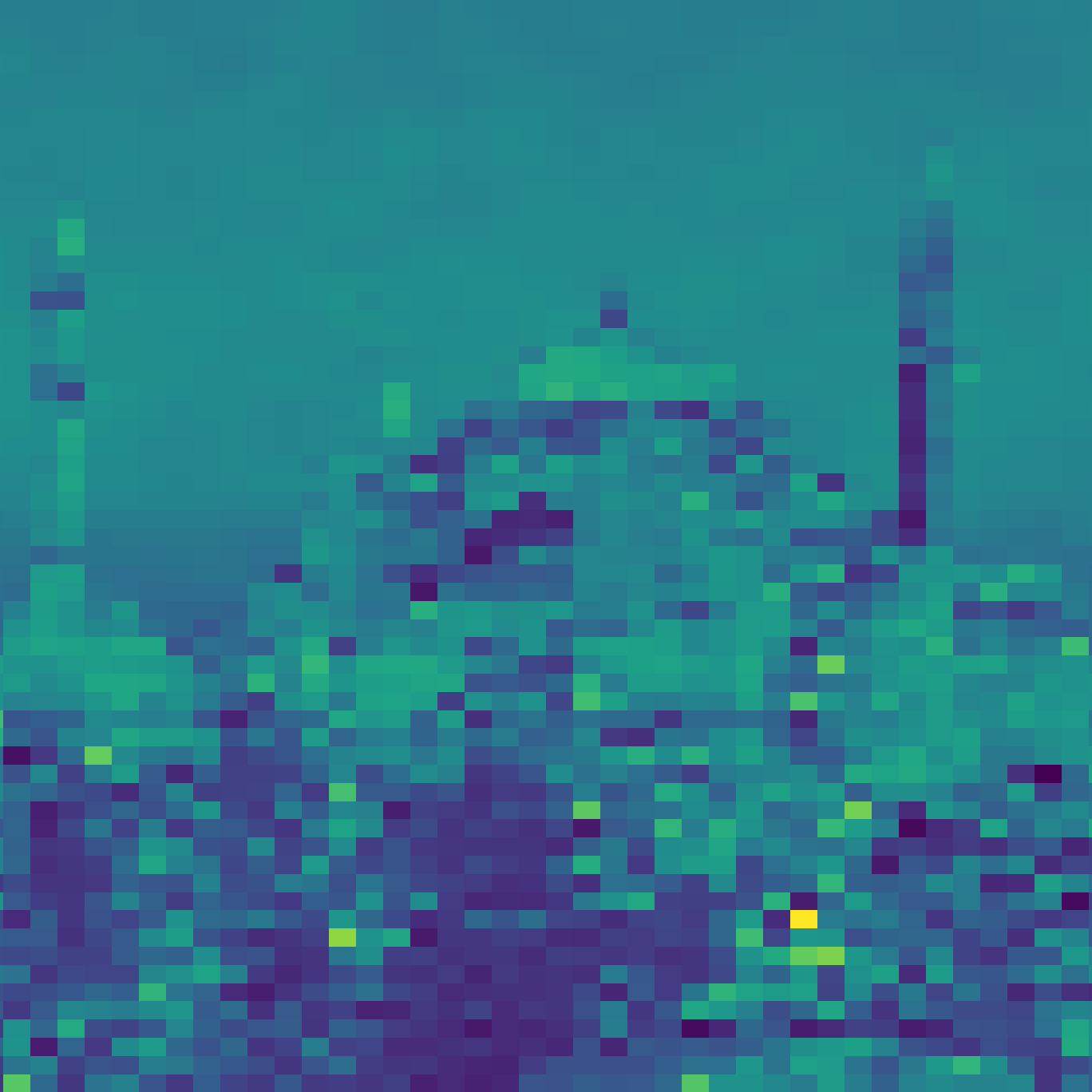}
       Att.-Head 4
   \end{subfigure}
   \hfill 
   \begin{subfigure}[t]{0.11\linewidth}
       \centering
       \includegraphics[width=\linewidth]{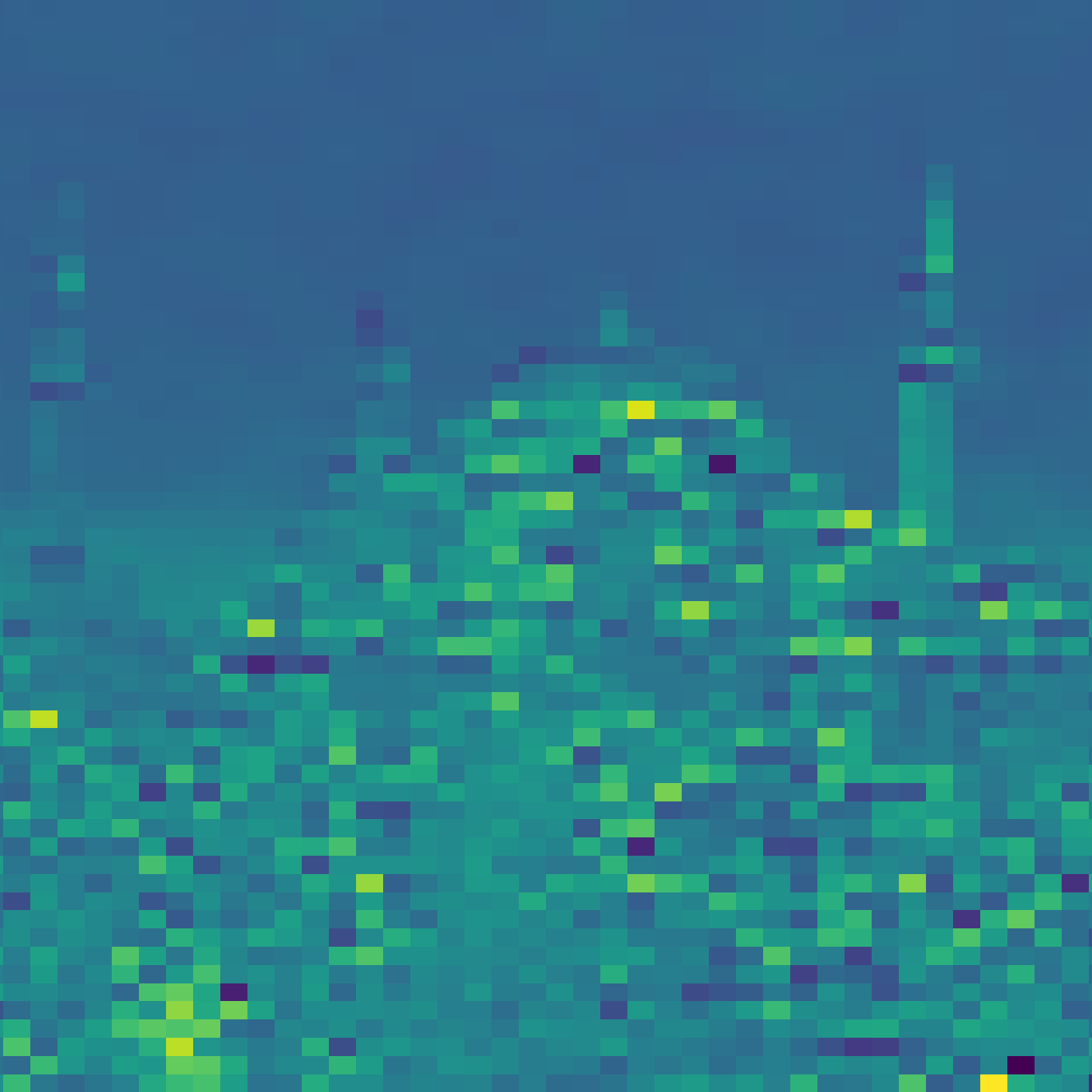}
       Att.-Head 5
   \end{subfigure}
   \hfill 
   \begin{subfigure}[t]{0.11\linewidth}
       \centering
       \includegraphics[width=\linewidth]{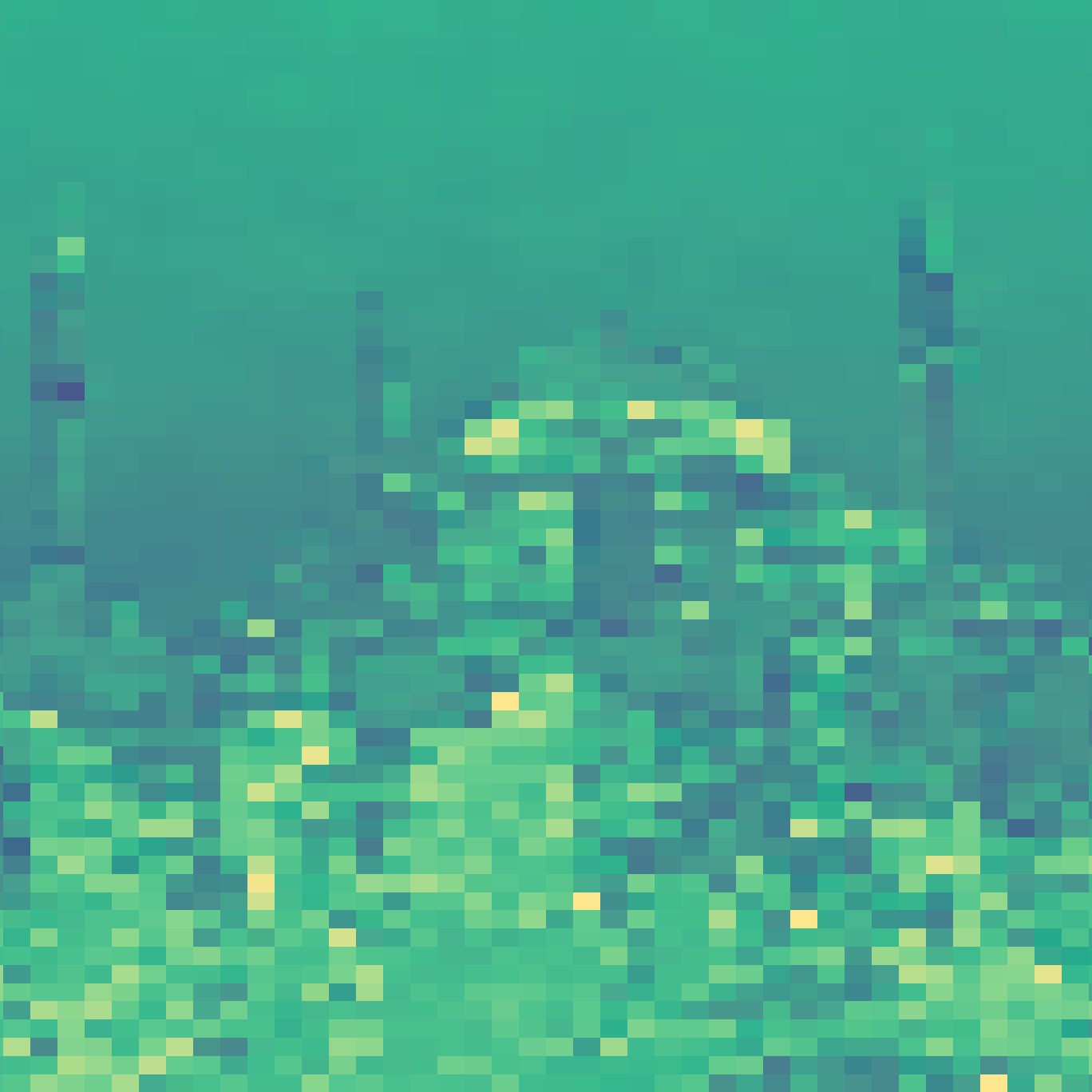}
       MAX
   \end{subfigure}
   
   \caption{\label{fig:more_att_heads}
Comparison of attention maps derived from different DINO heads.
   }
\end{figure*}

\begin{figure*}[!t] 
\captionsetup[subfigure]{position=b}
   \begin{subfigure}[t]{0.24\linewidth}
       \centering
       \includegraphics[width=\linewidth]{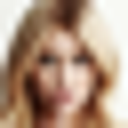}
       LR \\ $\uparrow$ PSNR: \\ $\uparrow$ SSIM: \\ $\downarrow$ LPIPS:
    \end{subfigure}
    \begin{subfigure}[t]{0.24\linewidth}
       \centering
       \includegraphics[width=\linewidth]{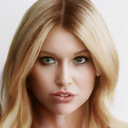}
       SR3 \\ 23.061 \\ 0.6208 \\ 0.0504
   \end{subfigure}
\hfill 
    \begin{subfigure}[t]{0.24\linewidth}
       \centering
       \includegraphics[width=\linewidth]{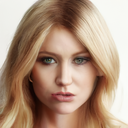}
       SR3+YODA \\ 23.289 \\ 0.6334 \\ 0.0502
   \end{subfigure}
   \begin{subfigure}[t]{0.24\linewidth}
       \centering
       \includegraphics[width=\linewidth]{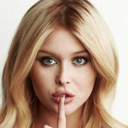}
       HR
    \end{subfigure}
   \caption{
   SR3 and SR3+YODA reconstructions, $16 \rightarrow 128$ (8x).
   The color shift in SR3 can still be observed (e.g., see background).
   YODA produces higher-quality images without color shift issues.
   }
   \label{fig:image16by16}
\end{figure*}

\begin{figure*}[h!]
    \begin{center}
        \includegraphics[width=.95\textwidth]{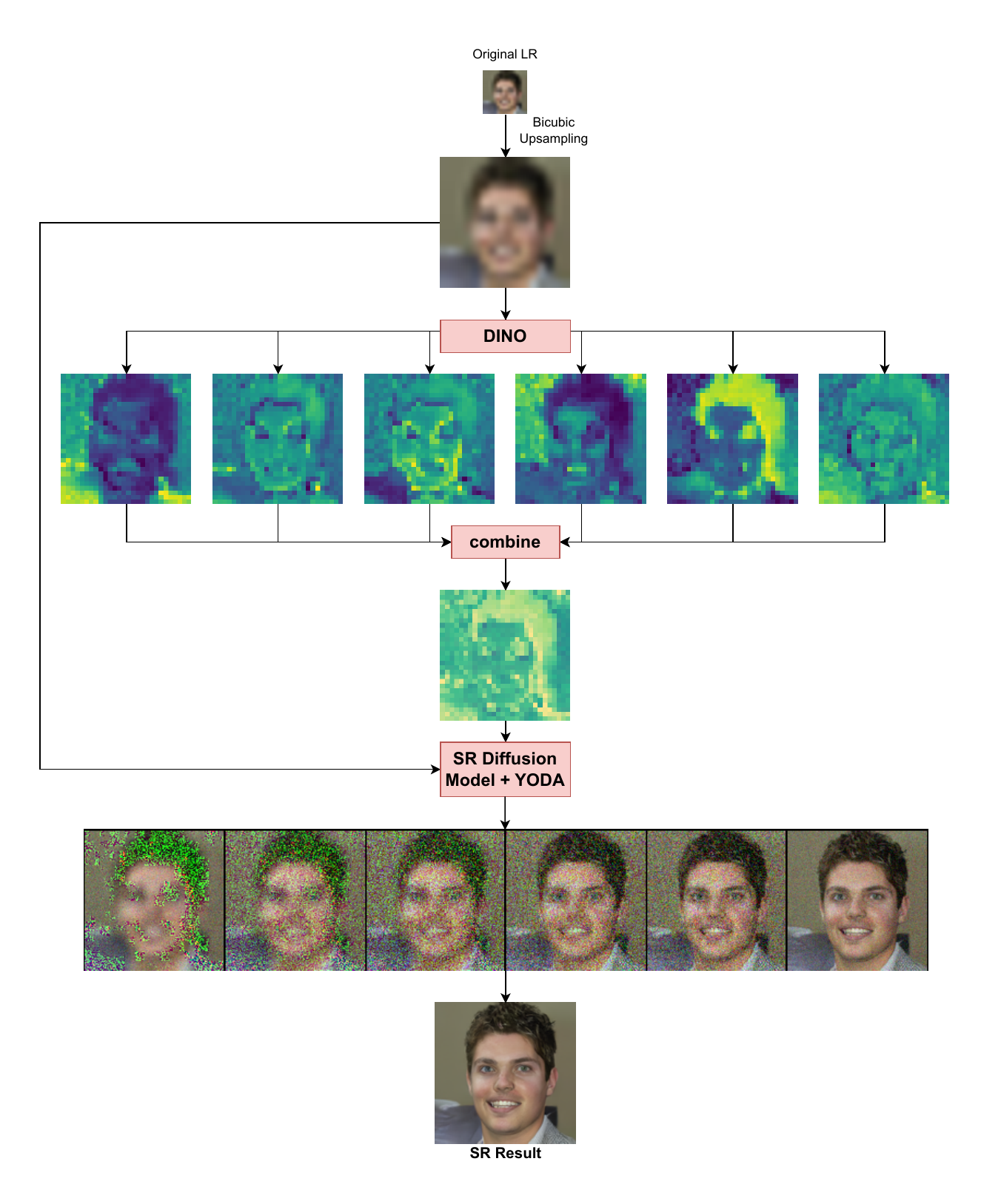}
        \caption{\label{fig:more_working_pipeline}
        An Overview of integrating YODA and DINO within SR diffusion models. Our process begins with using DINO to extract multiple attention maps. These maps are then combined to form a singular comprehensive attention map, denoted as $\mathbf{A}$. Subsequently, leveraging $\mathbf{A}$, YODA defines a unique diffusion schedule through time-dependent masks $\mathbf{M}(t)$ for every spatial location, as detailed within our method section.
        }
    \end{center}
\end{figure*}

\begin{figure*}[h!]
    \begin{center}
        \includegraphics[width=\linewidth]{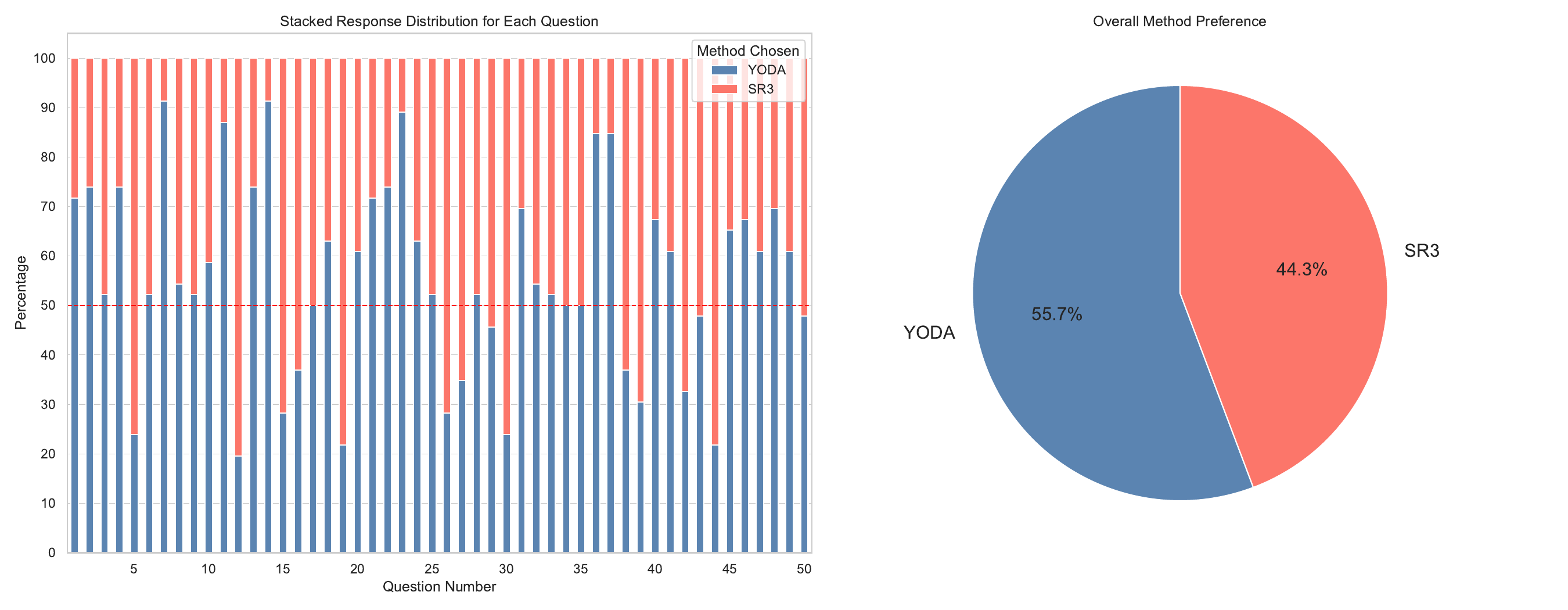}
        \caption{\label{fig:userstudy}
        Results of our user study. We randomly selected 50 images from the CelebA-HQ dataset and let 45 participants decide which SR predictions are preferred with respect to the given LR image. The scaling of the tested SR prediction task was $16 \times 16 \rightarrow 128 \times 128$ (8x scaling).
        As a result, YODA+SR3 was preferred in 55.7\% of all cases. 
        }
    \end{center}
\end{figure*}

\begin{figure*}[h!]
    \begin{center}
        \includegraphics[width=.7\textwidth]{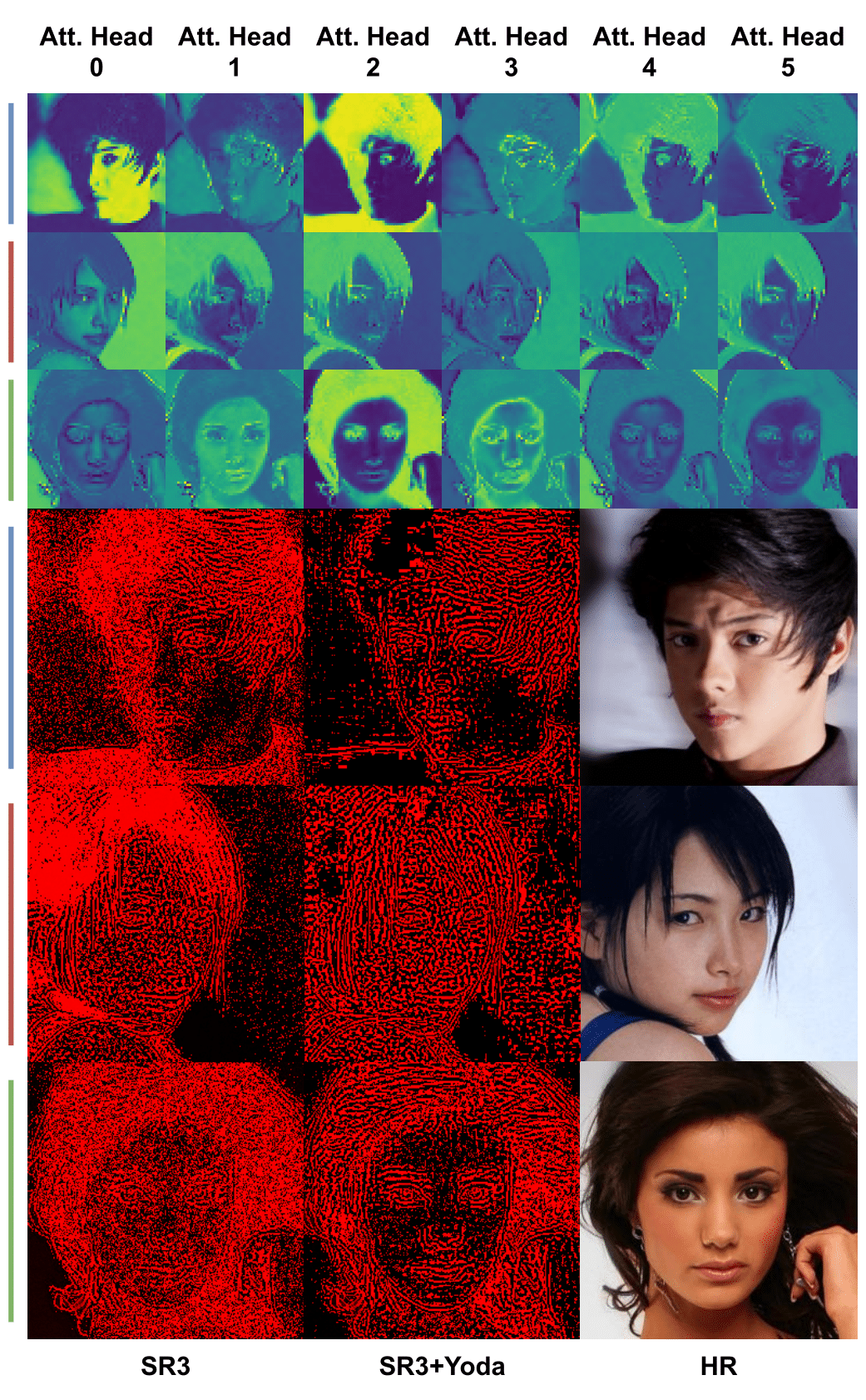}
        \caption{\label{fig:more_error maps}
        Error Maps of SR3 with and without YODA. The brighter, the higher the error.
        The attention maps generated by DINO are shown above.
        YODA produces smaller errors, especially in important areas such as face details and hair.
        }
    \end{center}
\end{figure*}

\begin{figure*}[!h] 
\captionsetup[subfigure]{position=b}
    \begin{subfigure}[t]{0.24\linewidth}
       \includegraphics[width=\linewidth]{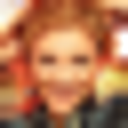}
    \end{subfigure}
    \begin{subfigure}[t]{0.24\linewidth}
       \includegraphics[width=\linewidth]{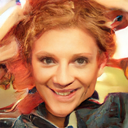}
   \end{subfigure}
\hfill
    \begin{subfigure}[t]{0.24\linewidth}
       \includegraphics[width=\linewidth]{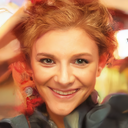}
   \end{subfigure}
   \begin{subfigure}[t]{0.24\linewidth}
       \includegraphics[width=\linewidth]{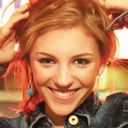}
    \end{subfigure}
\hfill
       \begin{subfigure}[t]{0.24\linewidth}
       \includegraphics[width=\linewidth]{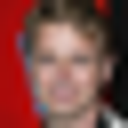}
    \end{subfigure}
    \begin{subfigure}[t]{0.24\linewidth}
       \includegraphics[width=\linewidth]{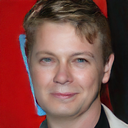}
   \end{subfigure}
\hfill
    \begin{subfigure}[t]{0.24\linewidth}
       \includegraphics[width=\linewidth]{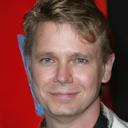}
   \end{subfigure}
   \begin{subfigure}[t]{0.24\linewidth}
       \includegraphics[width=\linewidth]{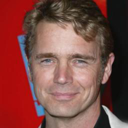}
    \end{subfigure}
\hfill
       \begin{subfigure}[t]{0.24\linewidth}
       \includegraphics[width=\linewidth]{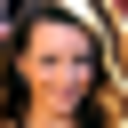}
    \end{subfigure}
    \begin{subfigure}[t]{0.24\linewidth}
       \includegraphics[width=\linewidth]{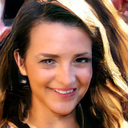}
   \end{subfigure}
\hfill
    \begin{subfigure}[t]{0.24\linewidth}
       \includegraphics[width=\linewidth]{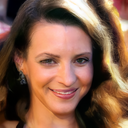}
   \end{subfigure}
   \begin{subfigure}[t]{0.24\linewidth}
       \includegraphics[width=\linewidth]{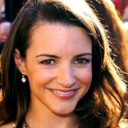}
    \end{subfigure}
\hfill
       \begin{subfigure}[t]{0.24\linewidth}
       \includegraphics[width=\linewidth]{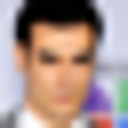}
    \end{subfigure}
    \begin{subfigure}[t]{0.24\linewidth}
       \includegraphics[width=\linewidth]{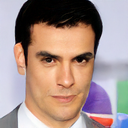}
   \end{subfigure}
\hfill
    \begin{subfigure}[t]{0.24\linewidth}
       \includegraphics[width=\linewidth]{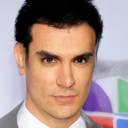}
   \end{subfigure}
   \begin{subfigure}[t]{0.24\linewidth}
       \includegraphics[width=\linewidth]{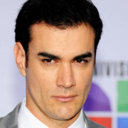}
    \end{subfigure}
\hfill
   \begin{subfigure}[t]{0.24\linewidth}
       \centering
       \includegraphics[width=\linewidth]{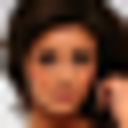}
       LR
    \end{subfigure}
    \begin{subfigure}[t]{0.24\linewidth}
       \centering
       \includegraphics[width=\linewidth]{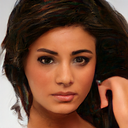}
       SR3
   \end{subfigure}
\hfill
    \begin{subfigure}[t]{0.24\linewidth}
       \centering
       \includegraphics[width=\linewidth]{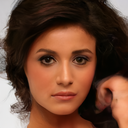}
       SR3+YODA
   \end{subfigure}
   \begin{subfigure}[t]{0.24\linewidth}
       \centering
       \includegraphics[width=\linewidth]{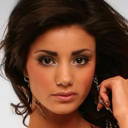}
       HR
    \end{subfigure}
   \caption{\label{fig:more_comparison_16}
   A comparison of LR, HR, SR3, and SR3+YODA images for $16 \rightarrow 128$.
   }
 
\end{figure*}

\begin{figure*}[!h] 
\captionsetup[subfigure]{position=b}
    \begin{subfigure}[t]{0.19\linewidth}
       \centering
       Original\\
       \includegraphics[width=.95\linewidth]{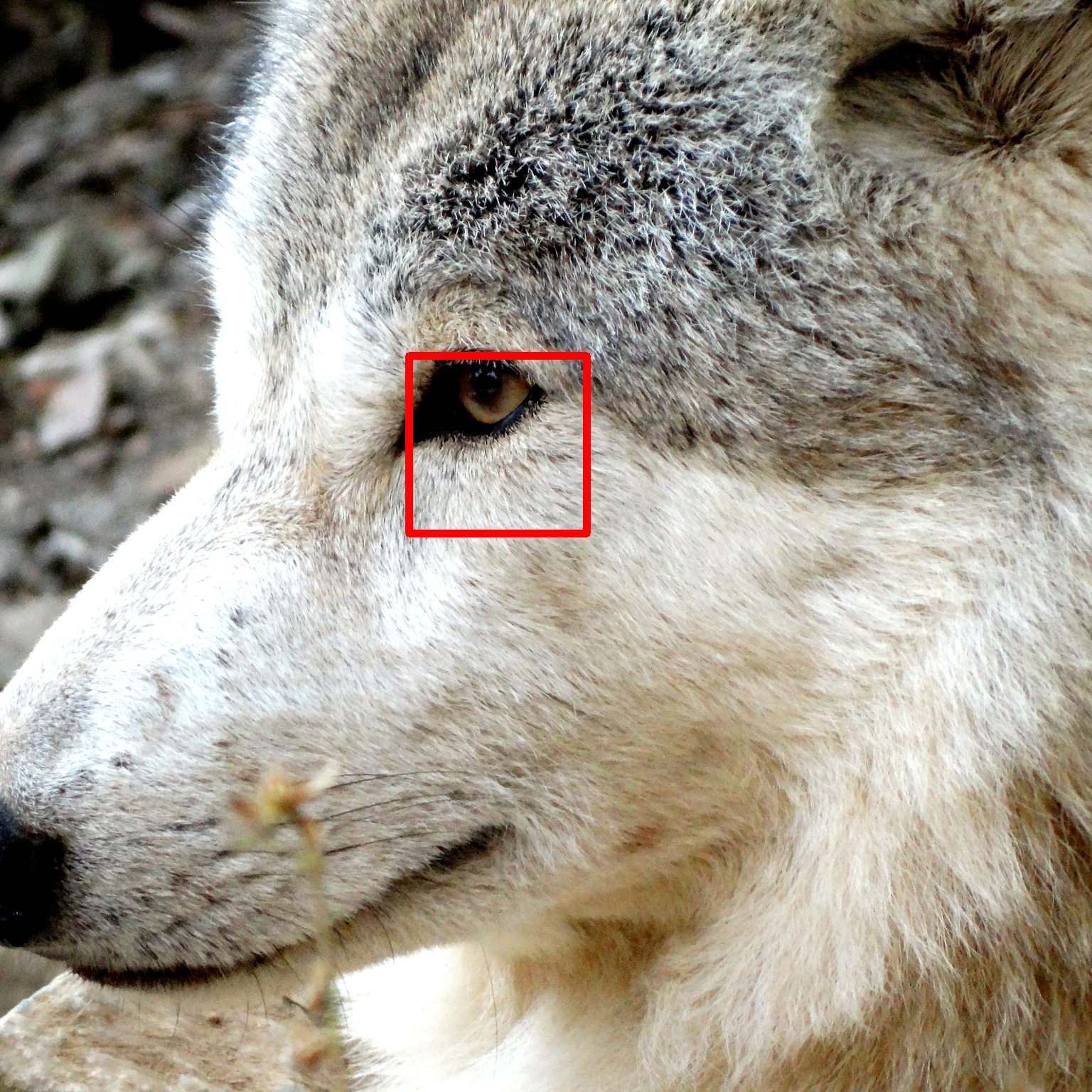}
       \text{}
    \end{subfigure}
    \begin{subfigure}[t]{0.19\linewidth}
       \centering
       LR \\
       \includegraphics[width=.95\linewidth]{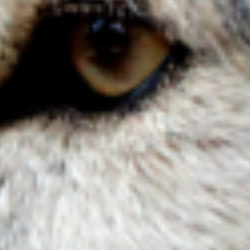}
       \text{}
    \end{subfigure}
    \begin{subfigure}[t]{0.19\linewidth}
       \centering
       SRDiff\\
       \includegraphics[width=.95\linewidth]{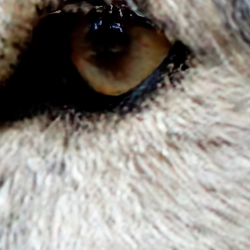}
       LPIPS: 0.2746
    \end{subfigure}
    \begin{subfigure}[t]{0.19\linewidth}
       \centering
       SRDiff+YODA\\
       \includegraphics[width=.95\linewidth]{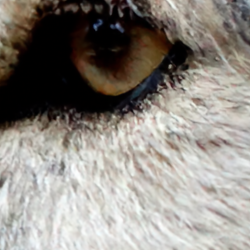}
       LPIPS: 0.1683
    \end{subfigure}
    \begin{subfigure}[t]{0.19\linewidth}
       \centering
       HR \\
       \includegraphics[width=.95\linewidth]{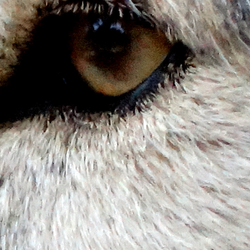}
       \text{}
    \end{subfigure}
    \begin{subfigure}[t]{0.19\linewidth}
       \centering
       \includegraphics[width=.95\linewidth]{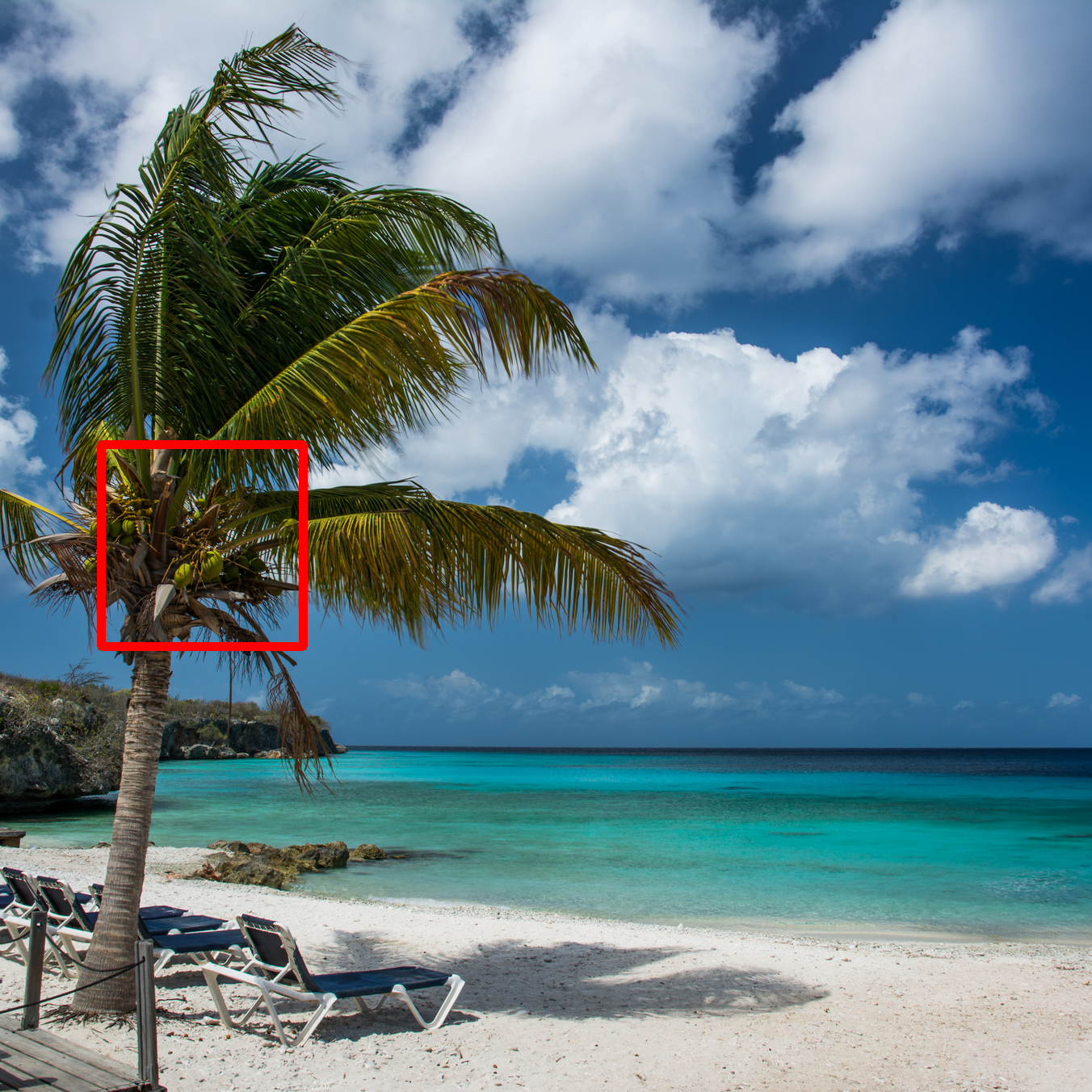}
       \text{}
    \end{subfigure}
    \begin{subfigure}[t]{0.19\linewidth}
       \centering
       \includegraphics[width=.95\linewidth]{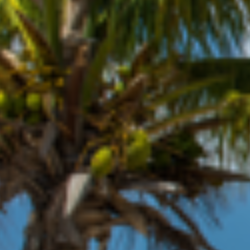}
       \text{}
    \end{subfigure}
    \begin{subfigure}[t]{0.19\linewidth}
       \centering
       \includegraphics[width=.95\linewidth]{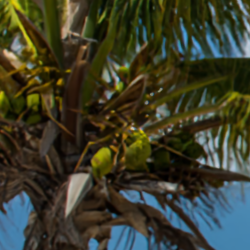}
       LPIPS: 0.2579
    \end{subfigure}
    \begin{subfigure}[t]{0.19\linewidth}
       \centering
       \includegraphics[width=.95\linewidth]{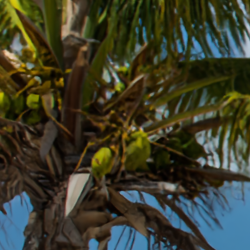}
       LPIPS: 0.2276
    \end{subfigure}
    \begin{subfigure}[t]{0.19\linewidth}
       \centering
       \includegraphics[width=.95\linewidth]{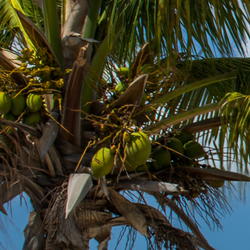}
       \text{}
    \end{subfigure}
    \begin{subfigure}[t]{0.19\linewidth}
       \centering
       \includegraphics[width=.95\linewidth]{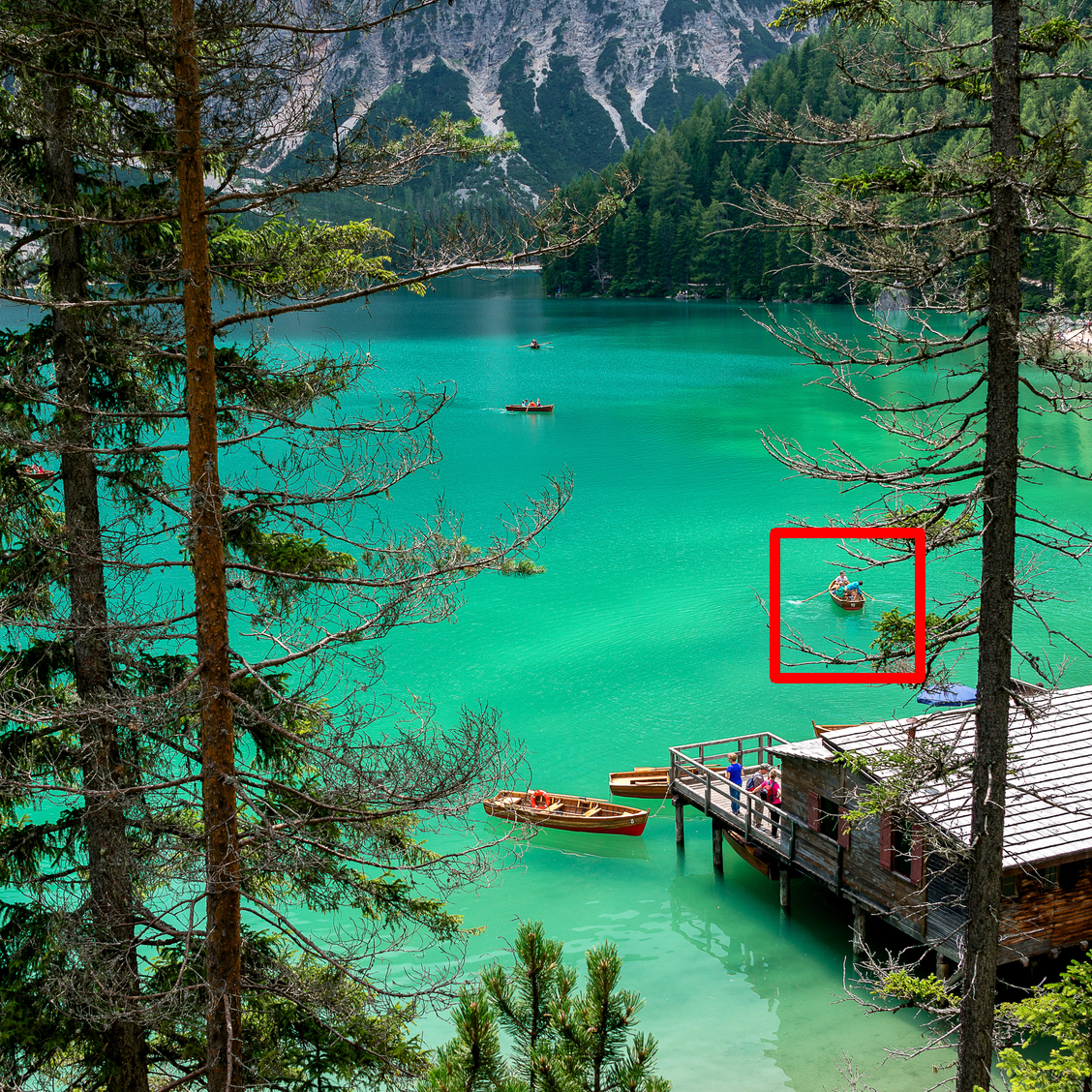}
       \text{}
    \end{subfigure}
    \begin{subfigure}[t]{0.19\linewidth}
       \centering
       \includegraphics[width=.95\linewidth]{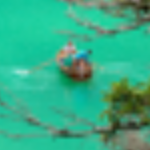}
       \text{}
    \end{subfigure}
    \begin{subfigure}[t]{0.19\linewidth}
       \centering
       \includegraphics[width=.95\linewidth]{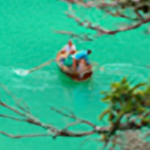}
       LPIPS: 0.3701
    \end{subfigure}
    \begin{subfigure}[t]{0.19\linewidth}
       \centering
       \includegraphics[width=.95\linewidth]{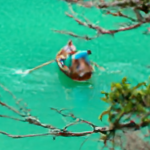}
       LPIPS: 0.2681
    \end{subfigure}
    \begin{subfigure}[t]{0.19\linewidth}
       \centering
       \includegraphics[width=.95\linewidth]{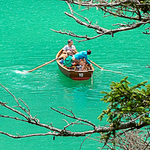}
       \text{}
    \end{subfigure}
    \begin{subfigure}[t]{0.19\linewidth}
       \centering
       \includegraphics[width=.95\linewidth]{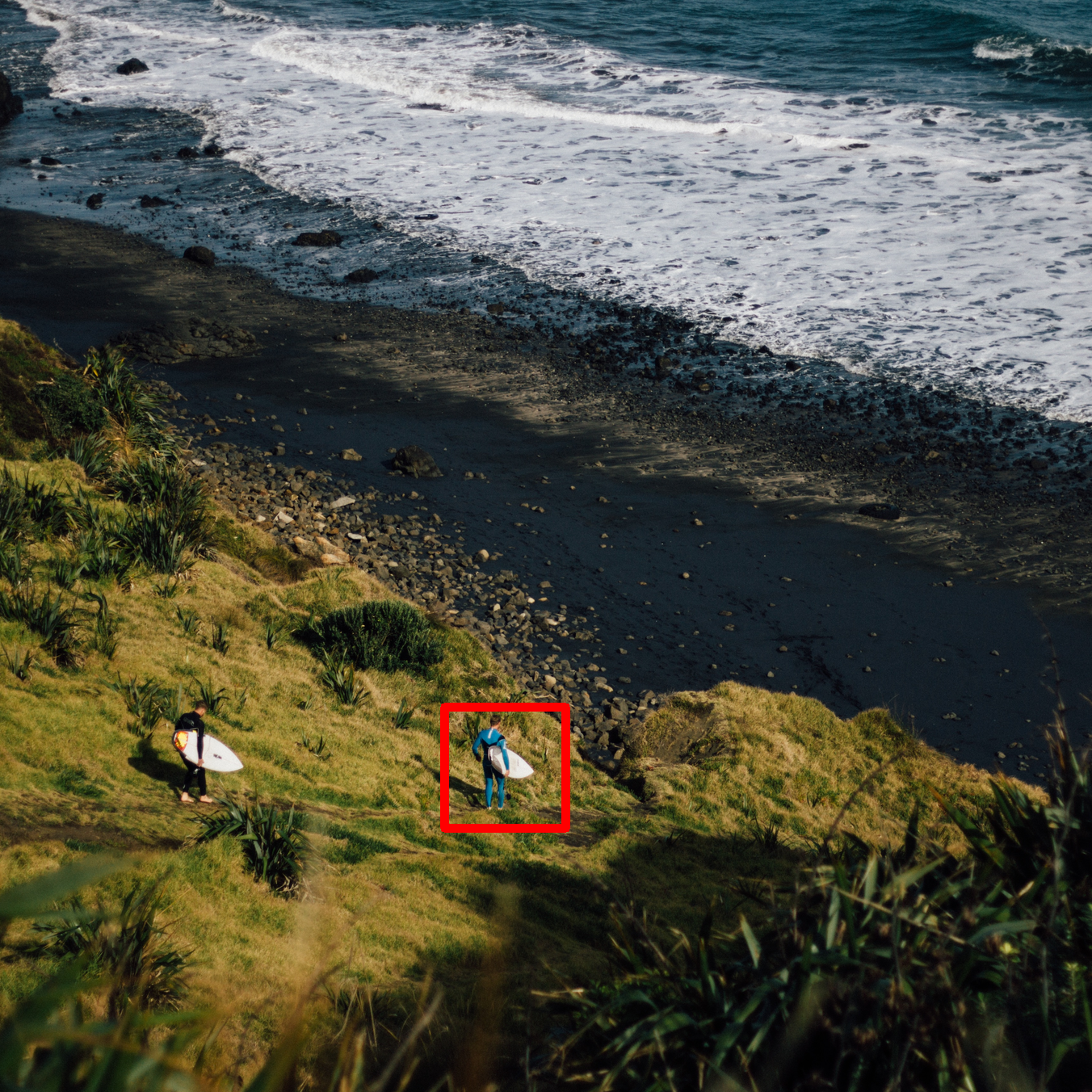}
       \text{}
    \end{subfigure}
    \begin{subfigure}[t]{0.19\linewidth}
       \centering
       \includegraphics[width=.95\linewidth]{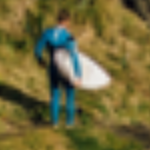}
       \text{}
    \end{subfigure}
    \begin{subfigure}[t]{0.19\linewidth}
       \centering
       \includegraphics[width=.95\linewidth]{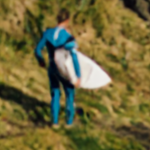}
       LPIPS: 0.2768
    \end{subfigure}
    \begin{subfigure}[t]{0.19\linewidth}
       \centering
       \includegraphics[width=.95\linewidth]{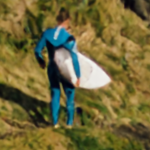}
       LPIPS: 0.2281
    \end{subfigure}
    \begin{subfigure}[t]{0.19\linewidth}
       \centering
       \includegraphics[width=.95\linewidth]{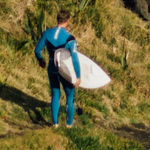}
       \text{}
    \end{subfigure}
    \begin{subfigure}[t]{0.19\linewidth}
       \centering
       \includegraphics[width=.95\linewidth]{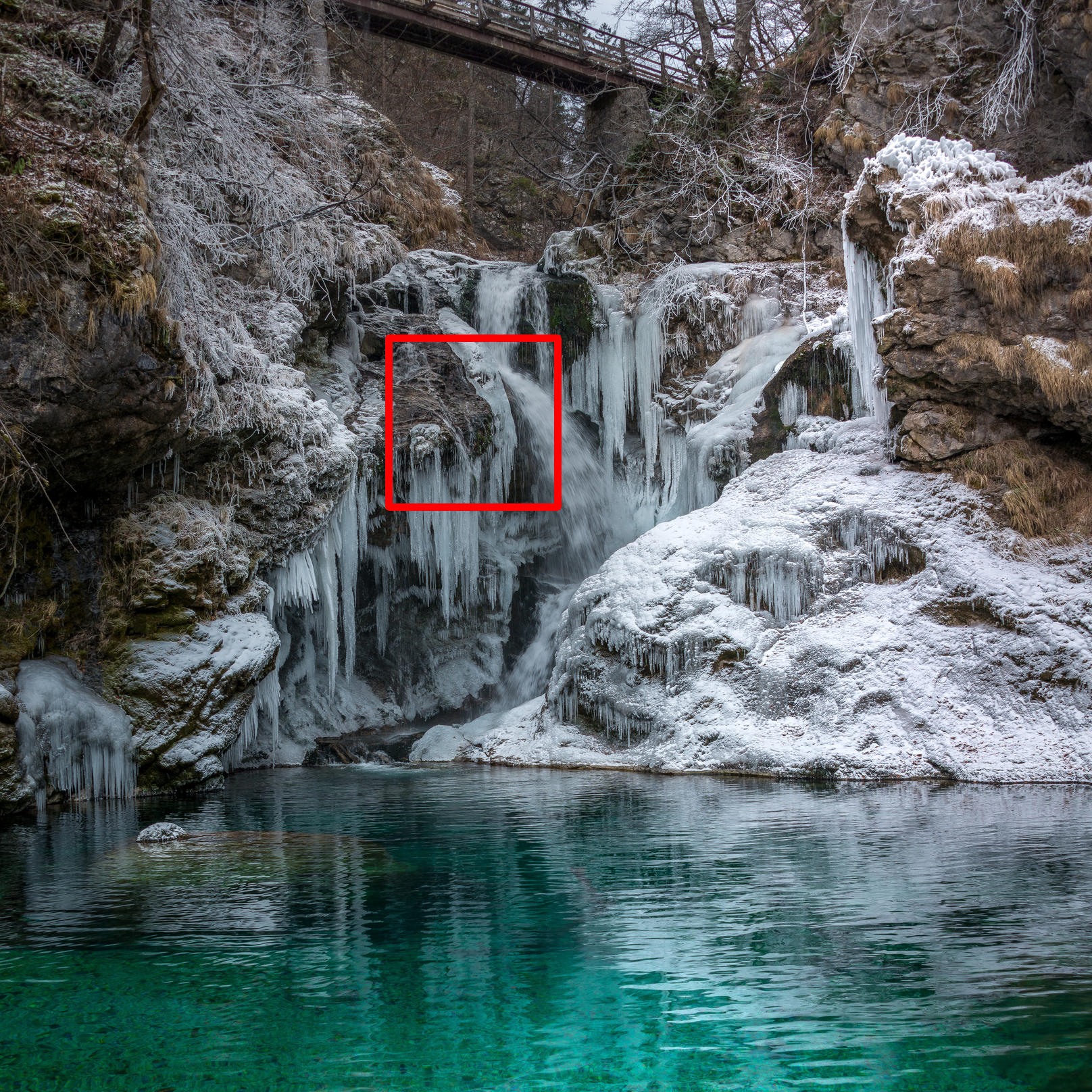}
       \text{}
    \end{subfigure}
    \begin{subfigure}[t]{0.19\linewidth}
       \centering
       \includegraphics[width=.95\linewidth]{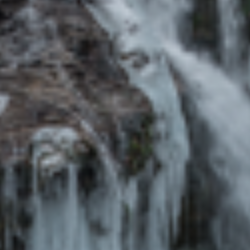}
       \text{}
    \end{subfigure}
    \begin{subfigure}[t]{0.19\linewidth}
       \centering
       \includegraphics[width=.95\linewidth]{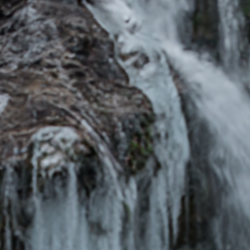}
       LPIPS: 0.4244
    \end{subfigure}
    \begin{subfigure}[t]{0.19\linewidth}
       \centering
       \includegraphics[width=.95\linewidth]{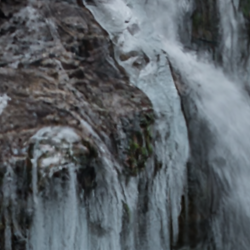}
       LPIPS: 0.2999
    \end{subfigure}
    \begin{subfigure}[t]{0.19\linewidth}
       \centering
       \includegraphics[width=.95\linewidth]{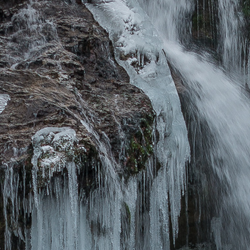}
       \text{}
    \end{subfigure}
    \begin{subfigure}[t]{0.19\linewidth}
       \centering
       \includegraphics[width=.95\linewidth]{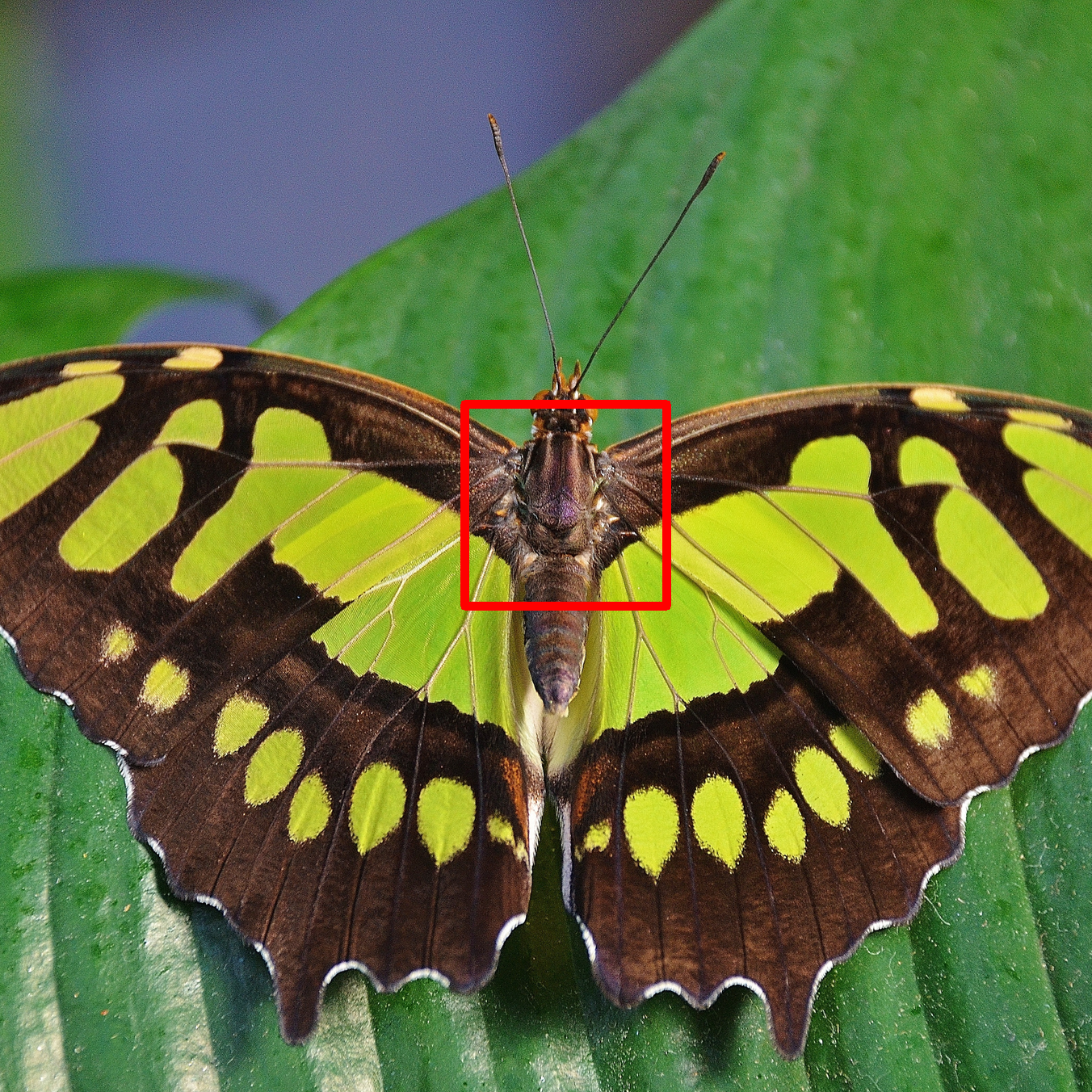}
       \text{}
    \end{subfigure}
    \hfill
    \begin{subfigure}[t]{0.19\linewidth}
       \centering
       \includegraphics[width=.95\linewidth]{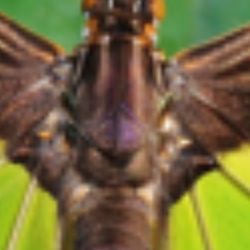}
       \text{}
    \end{subfigure}
    \hfill
    \begin{subfigure}[t]{0.19\linewidth}
       \centering
       \includegraphics[width=.95\linewidth]{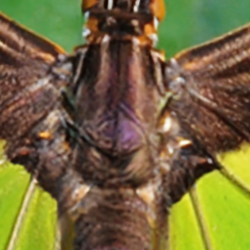}
       LPIPS: 0.4663 
    \end{subfigure}
    \hfill
    \begin{subfigure}[t]{0.19\linewidth}
       \centering
       \includegraphics[width=.95\linewidth]{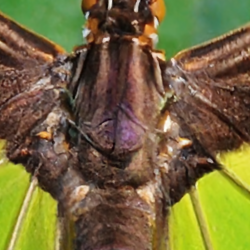}
       LPIPS: 0.3433
    \end{subfigure}
    \hfill
    \begin{subfigure}[t]{0.19\linewidth}
       \centering
       \includegraphics[width=.95\linewidth]{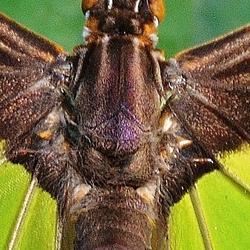}
       \text{}
    \end{subfigure}
   \caption{\label{fig:more_zoom_in}
   Zoomed-in comparison of LR, HR, SRDiff, and SRDiff+YODA on DIV2K.
   }
\end{figure*}

\begin{figure*}[!h] 

\captionsetup[subfigure]{position=b}
    \centering
    \begin{subfigure}[t]{.85\linewidth}
       \includegraphics[width=\linewidth]{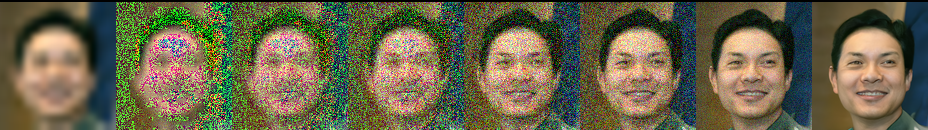}
    \end{subfigure}
    \begin{subfigure}[t]{.85\linewidth}
       \includegraphics[width=\linewidth]{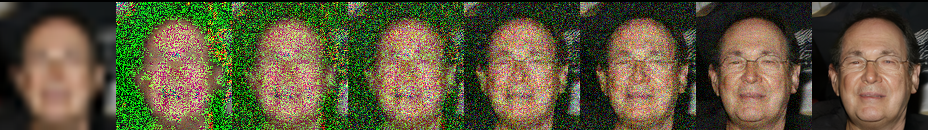}
    \end{subfigure}
    \begin{subfigure}[t]{.85\linewidth}
       \includegraphics[width=\linewidth]{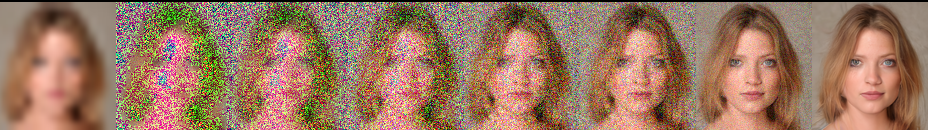}
    \end{subfigure}
    \begin{subfigure}[t]{.85\linewidth}
       \includegraphics[width=\linewidth]{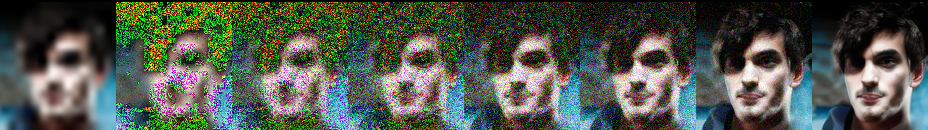}
    \end{subfigure}
    \begin{subfigure}[t]{.85\linewidth}
       \includegraphics[width=\linewidth]{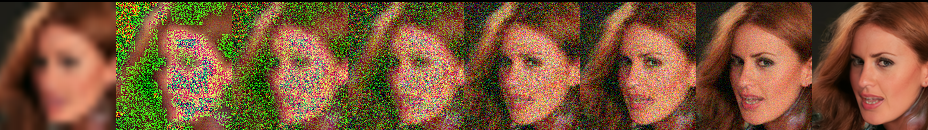}
    \end{subfigure}
    \begin{subfigure}[t]{.85\linewidth}
       \includegraphics[width=\linewidth]{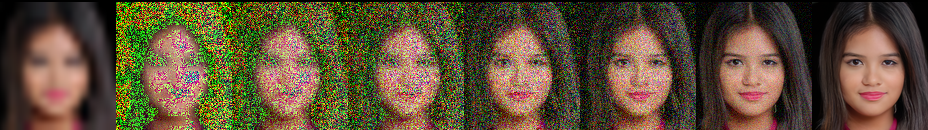}
    \end{subfigure}
    \begin{subfigure}[t]{.85\linewidth}
       \includegraphics[width=\linewidth]{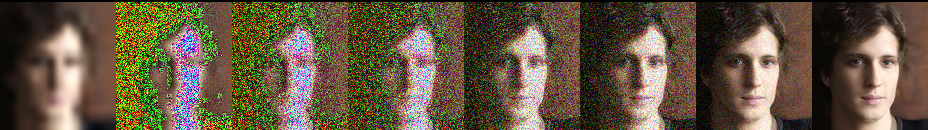}
    \end{subfigure}
    \begin{subfigure}[t]{.85\linewidth}
       \includegraphics[width=\linewidth]{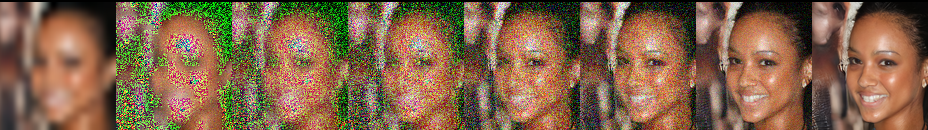}
    \end{subfigure}
    \begin{subfigure}[t]{.85\linewidth}
       \includegraphics[width=\linewidth]{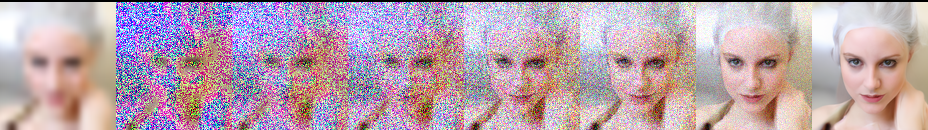}
    \end{subfigure}
    \begin{subfigure}[t]{.85\linewidth}
       \includegraphics[width=\linewidth]{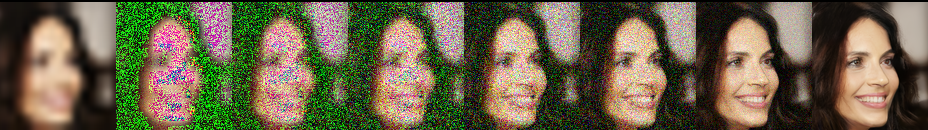}
    \end{subfigure}

   \caption{\label{fig:more_intermediate}
   Intermediate results of the YODA's guided diffusion process on CelebA-HQ
   (Time steps 189, 168, 147, 126, 105, 84, 63, 42, 21, 0 from left to right).
   }
\end{figure*}

\begin{figure*}[h!]
    \begin{center}
        \includegraphics[width=\linewidth]{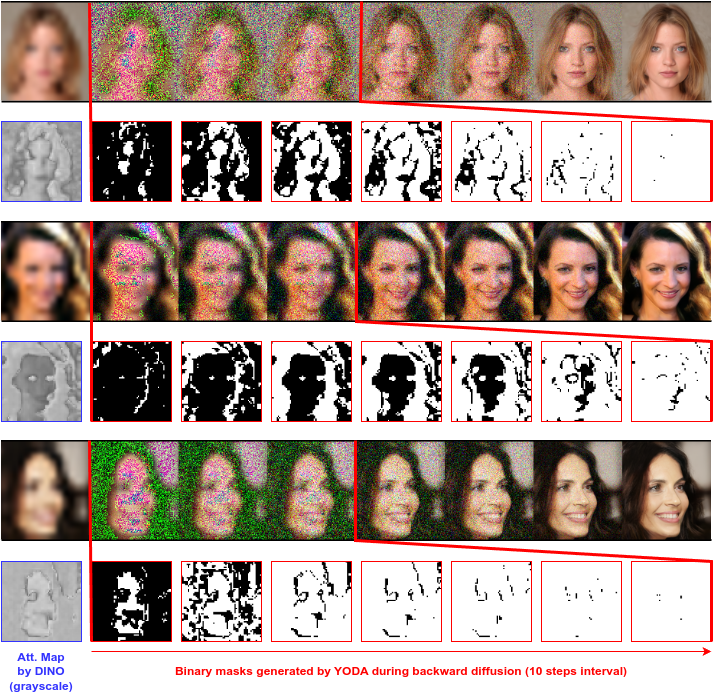}
        \caption{\label{fig:more_intermediate_binary}
        Intermediate binary masks of YODA's guided diffusion process on CelebA-HQ.
        }
    \end{center}
\end{figure*}

In the remaining part of the supplementary material, we provide additional visualizations, such as more on attention maps derived from different DINO heads (\autoref{fig:more_att_heads}), user study (\autoref{fig:userstudy}), an overview of the working pipeline of YODA (\autoref{fig:more_working_pipeline}), error maps (\autoref{fig:more_error maps}), comparisons on $16 \rightarrow 128$ (\autoref{fig:image16by16} and \autoref{fig:more_comparison_16}), zoomed-in comparisons (\autoref{fig:more_zoom_in}), intermediate results (\autoref{fig:more_intermediate}) as well as intermediate binary masking (\autoref{fig:more_intermediate_binary}).

\newpage

\newpage

\newpage

\newpage

\newpage

\end{document}